\PassOptionsToPackage{table}{xcolor}
\documentclass[letterpaper,twocolumn,10pt]{article}

\usepackage{hegroup}
\usepackage[table]{xcolor}
\usepackage[normalem]{ulem}
\useunder{\uline}{\ul}{}
\usepackage{tikz}
\usepackage{amsfonts}
\usepackage{xspace} 
\usepackage{amsmath}
\usepackage{mathrsfs}
\usepackage{amssymb}
\usepackage{enumitem}
\usepackage{amsmath}
\usepackage{amsthm}
\usepackage{multirow}
\usepackage{makecell}
\usepackage{caption}
\usepackage{subcaption}
\usepackage{float}
\usepackage{booktabs}
\usepackage{balance}
\usepackage{tcolorbox}
\usepackage{xurl}
\usepackage{hyperref}
\usepackage{tabularx}
\usepackage{cleveref}

\newcommand{\mypara}[1]{\noindent{\bf {#1}.}}
\newcommand{\Dataset}{$\mathsf{MGT\text{-}Academic}$\xspace}

\begin{document}
\pagestyle{plain}

\title{	
On the Generalization and Adaptation Ability of Machine-Generated Text Detectors in Academic Writing}
\date{}

\author{
Yule Liu\textsuperscript{1}\thanks{Equal contribution.}     \ \ \
Zhiyuan Zhong\textsuperscript{1,2}\footnotemark[1]   \ \ \ 
Yifan Liao\textsuperscript{3}  \ \ \ 
Zhen Sun\textsuperscript{1}  \ \ \ 
Jingyi Zheng\textsuperscript{1}  \ \ \ 
\\
\\
Jiaheng Wei\textsuperscript{1}  \ \ \ 
Qingyuan Gong\textsuperscript{4}  \ \ \ 
Fenghua Tong\textsuperscript{5} \ \ \
Yang Chen\textsuperscript{4} \ \ \
Yang Zhang\textsuperscript{6} \ \ \
Xinlei He\textsuperscript{1}\thanks{Corresponding author (\href{mailto:xinleihe@hkust-gz.edu.cn}{xinleihe@hkust-gz.edu.cn}).} \ \ \
\\
\\
\textsuperscript{1}\textit{The Hong Kong University of Science and Technology (Guangzhou)} \ \ \  \\
\textsuperscript{2}\textit{Southern University of Science and Technology } \ \ \ \\
\textsuperscript{3}\textit{National University of Singapore (Chongqing Research Institute)}\ \ \ \\
\textsuperscript{4}\textit{Fudan University} \ \ \
\textsuperscript{5}\textit{Qilu University of Technology} \ \ \
\textsuperscript{6}\textit{CISPA Helmholtz Center for Information Security} 
}

\maketitle
\begin{abstract}
The rising popularity of large language models (LLMs) has raised concerns about machine-generated text (MGT), particularly in academic settings, where issues like plagiarism and misinformation are prevalent. 
As a result, developing a highly generalizable and adaptable MGT detection system has become an urgent priority. 
Given that LLMs are most commonly misused in academic writing, this work investigates the generalization and adaptation capabilities of MGT detectors in three key aspects specific to academic writing:
First, we construct \Dataset, a large-scale dataset comprising over 336M tokens and 749K samples.
\Dataset focuses on academic writing, featuring human-written texts (HWTs) and MGTs across STEM, Humanities, and Social Sciences, paired with an extensible code framework for efficient benchmarking.
Second, we benchmark the performance of various detectors for binary classification and attribution tasks in both in-domain and cross-domain settings. 
This benchmark reveals the often-overlooked challenges of attribution tasks.
Third, we introduce a novel attribution task where models have to adapt to new classes over time without (or with very limited) access to prior training data in both few-shot and many-shot scenarios.
We implement eight different adapting techniques to improve the performance and highlight the inherent complexity of the task.
Our findings provide insights into the generalization and adaptation ability of MGT detectors across diverse scenarios and lay the foundation for building robust, adaptive detection systems.
The code framework is available at \url{https://github.com/Y-L-LIU/MGTBench-2.0}.
\end{abstract}

\begin{figure}
    \centering
    \includegraphics[width=1\linewidth]{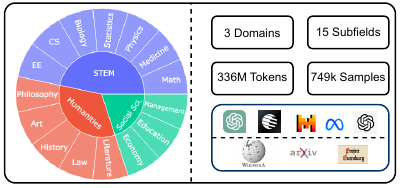}
    \caption{Overview of \Dataset. It shows the data source and split domains. }
    \label{fig:overview}
\end{figure}
\section{Introduction}

Recent advancements in large language models (LLMs) showcase their strong ability to tackle a wide range of natural language processing (NLP) tasks~\cite{TMSAABBBBBBBCCCEFFFFGGGHHHIKKKKKKLLLLLMMMMMNPRRSSSSSTTTWKXYZZFKNRSES23, OWJAWMZASRSHKMSAWCLL22,O23}. 
Its versatility and superiority across numerous domains unlock remarkable real-world applications, e.g., education, idea crafting, and context refinement~\cite{zhao2024surveylargelanguagemodels}.
However, the ease and convenience have opened the door to abuse, particularly in academic writing and creative industries, leading to severe ethical and practical challenges~\cite{YJPJAAMTGTo24}.
Recent efforts~\cite{MLKMF23, BZTYZ24} have focused on developing techniques to distinguish machine-generated text (MGT) from human-written text (HWT) and benchmarking their performance~\cite{HSCBZ23}, primarily in binary classification tasks, where the goal is to determine whether a given text is MGT or HWT.
Besides binary classification, the text attribution task, which aims to identify the specific source LLM that generated the text, presents additional challenges and remains under-explored.
However, the performance and generalization ability of detectors in this task lacks comprehensive discussion and evaluation.

Given that LLMs are most commonly misused in academic writing, we conduct an in-depth investigation of the generalization ability of MGT detectors in the following aspects specific to academic writing:
First, we construct a large-scale MGT dataset named \Dataset focused on academic writing, comprising over 336M tokens and 749K samples across three academic domains: STEM, Humanities, and Social Sciences.
Within each domain, we collect HWT data from Wikipedia and academic texts sourced, including Arxiv or Project Gutenberg, depending on the scenario.
Each HWT has corresponding MGTs generated by five popular LLMs.
Further, we build a publicly available, extendable, and user-friendly code framework for the community, which enables fast and effective benchmarking for existing methods in binary classification and text attribution tasks. 
It covers state-of-the-art methods, including seven metric-based detectors and five model-based detectors.

Second, leveraging \Dataset, we conduct a comprehensive investigation into the performance and transferability of detectors across various tasks.
First, we assess the performance of detectors in both binary classification and attribution tasks. 
As attribution presents a particularly challenging problem, we analyze the underlying reasons why metric-based methods fail in this context.
Next, we carry out extensive experiments to examine how detectors transfer to specific domains, such as STEM, Humanities, and Social Sciences, as well as leading LLMs like GPT-4omini and Llama-3.1-70b.
In the binary classification task, we study the generalization abilities of detectors across diverse data distributions.
In the attribution task, we study the generalization abilities of detectors in different domains.
Additionally, we investigate the effect of adding few-shot examples from the target domains in improving the transferability.

Third, since new LLMs are continuously released, each with different characteristics and unique stylistics, we consider a new attribution task, where a model would adapt to the new class introduced over time without (or with very limited) access to the original training data for earlier classes. 
This task is crucial for real-world applications where MGTs from new LLMs become available in stages and retraining the model from scratch is impractical due to computational or data storage constraints.
In this paper, we consider two practical settings for adaptation, i.e., few-shot and many-shot settings, depending on the number of accessible examples in the new class. 
To the best of our knowledge, we are the first to discuss how the detectors adapt to new MGTs in detection.
We benchmark the performance of detectors when adapting to new LLMs in different scenarios and equip the detector with eight different techniques to improve the performance.
In summary, our contributions can be listed as follows:

\begin{itemize}[leftmargin=*]
    \item We introduce \Dataset, a large-scale MGT dataset focused on academic writing, encompassing over 336M tokens and 749K samples across three academic domains: STEM, Humanities, and Social Sciences. 
    We additionally provide an extensible code framework, which will be made publicly available, to efficiently benchmark existing MGT detectors in different tasks.
    \item We conduct extensive experiments and benchmark the performance of various detectors for binary classification and attribution tasks in both in-domain and cross-domain settings. 
    The results show the inherent challenges of the attribution task.
    We further analyze the underlying reasons why metric-based methods fail in the attribution task and investigate the effect of adding few-shot examples from the target domains in improving transferability.
    \item We introduce a new attribution task where detectors adapt to new classes over time without (or with very limited) access to prior training data in few-shot and many-shot scenarios. 
    We benchmark the performance of model-based detectors with different adaptation techniques.
    Despite the improved performance, the remaining gap to the ideal performance (fine-tuning all data at once) highlights the complexity of this task, underscoring the need for further investigation in the future.
\end{itemize}

\section{Related Work}
\label{sec:related}
\mypara{MGT Detection}
Recent advancements in LLMs have empowered users to tackle a wide range of NLP tasks, demonstrating their versatility and superiority across numerous domains~\cite{TMSAABBBBBBBCCCEFFFFGGGHHHIKKKKKKLLLLLMMMMMNPRRSSSSSTTTWKXYZZFKNRSES23, OWJAWMZASRSHKMSAWCLL22, O23}. 
Exploiting LLMs is especially convenient in academic writing~\cite{zhao2024surveylargelanguagemodels, minaee2024largelanguagemodelssurvey}, such as generating ideas, drafting articles, or refining content.
However, the ease and convenience can be significantly abused, raising concerns about authenticity, as well as ethical questions regarding originality and over-dependence on AI-generated content~\cite{yi2024jailbreakattacksdefenseslarge, gan2024navigatingriskssurveysecurity}.
To prevent the misuse of MGT data, recent studies~\cite{BAA08, GSR19, GZWJNDYW23, MLKMF23, HSCBZ23} have developed a variety of MGT detectors, which can be broadly categorized into metric-based and model-based methods.
Metric-based methods~\cite{GSR19, MLKMF23, JYDP23} leverage proxy LLMs to extract features from processed text and train an additional classifier to model the relationship between features and labels. 
In contrast, model-based methods~\cite{IDCE20, GZWJNDYW23} typically integrate a classification head into a BERT model and fine-tune the augmented model on supervised datasets.
The detectors in this paper are listed in \Cref{sec-app:detecors}.

Several efforts have aimed to benchmark the performance of MGT detectors. 
For example, MGTBench~\cite{HSCBZ23} provided a comprehensive evaluation framework for these detectors, which utilizes existing HWT datasets, including Essay, WP, and Reuters.
M4GTBench~\cite{YJPJAAMTGTo24} extended this by benchmarking performance on multilingual and multi-source datasets.
CUDRT~\cite{tao2024reliabledetectionllmgeneratedtexts} supports reproducible experiments and enables in-depth analysis of how operational diversity influences detection performance.
Additionally, DetectRL~\cite{DBLP:journals/corr/abs-2410-23746} assessed detectors' robustness and generalization capabilities in the face of adversarial attacks.
While existing studies emphasize transferability across datasets and LLMs in binary classification, they pay less attention to the generalization ability of detectors in attribution tasks.

\begin{table*}[t]
\centering
\caption{\textbf{Experiment Result for In-distribution Binary Classification.} 
We train and test the detectors on the same data domain.
The results are reported using F1 score.
ST. represents STEM, Hu. represents Humanity, and So. represents Social Science. 
The larger values with blue colors indicate better performance and lower values with red colors indicate weaker performance.
For the abnormal results, we use ``-'' as the replace holder.}
\setlength{\tabcolsep}{3pt}
\renewcommand{\arraystretch}{1.1}
\label{tab:binary-benchmark}
\begin{tabular}{c|ccc|ccc|ccc|ccc|ccc}\toprule
                         & \multicolumn{3}{c|}{Llama-3.1-70b}                                                            & \multicolumn{3}{c|}{Mixtral-8$\times$7b}                                                             & \multicolumn{3}{c|}{MoonShot-8k}                                                              & \multicolumn{3}{c|}{GPT-4omini}                                                              & \multicolumn{3}{c}{GPT-3.5}                                                                   \\ \cmidrule{2-16} 
\multirow{-2}{*}{Method} & ST.                           & Hu.                           & So.                           & ST.                           & Hu.                           & So.                           & ST.                           & Hu.                           & So.                           & ST.                           & Hu.                           & So.                           & ST.                           & Hu.                           & So.                           \\ \midrule
LL                       & \cellcolor[HTML]{FEFEFE}0.714 & \cellcolor[HTML]{F3F9FB}0.794 & \cellcolor[HTML]{F2F8FB}0.803 & \cellcolor[HTML]{FEFAF7}0.662 & \cellcolor[HTML]{FBFDFE}0.749 & \cellcolor[HTML]{F1F7FB}0.809 & \cellcolor[HTML]{FEFEFD}0.711 & \cellcolor[HTML]{F9FCFD}0.760 & \cellcolor[HTML]{F1F7FB}0.806 & \cellcolor[HTML]{FEF8F4}0.638 & \cellcolor[HTML]{FEFCFB}0.689 & \cellcolor[HTML]{F8FBFD}0.765 & \cellcolor[HTML]{FDEBE0}0.481 & \cellcolor[HTML]{FEF5F0}0.606 & \cellcolor[HTML]{FEFEFD}0.709 \\
Entropy                  & \cellcolor[HTML]{F9FCFD}0.759 & \cellcolor[HTML]{FEFDFD}0.707 & \cellcolor[HTML]{EDF5FA}0.829 & \cellcolor[HTML]{FEFCFA}0.688 & \cellcolor[HTML]{FEFAF7}0.659 & \cellcolor[HTML]{F3F8FB}0.795 & \cellcolor[HTML]{FFFFFF}0.723 & \cellcolor[HTML]{FEFAF7}0.662 & \cellcolor[HTML]{F6FAFC}0.780 & \cellcolor[HTML]{FEFCFA}0.683 & \cellcolor[HTML]{FEFCFA}0.686 & \cellcolor[HTML]{F8FBFD}0.763 & \cellcolor[HTML]{FEFBF9}0.679 & \cellcolor[HTML]{FEF8F4}0.640 & \cellcolor[HTML]{FEFDFC}0.700 \\
Rank                     & \cellcolor[HTML]{FEF6F2}0.618 & \cellcolor[HTML]{FEFDFC}0.697 & \cellcolor[HTML]{FEFEFE}0.713 & \cellcolor[HTML]{FEF6F1}0.617 & \cellcolor[HTML]{FEFCFB}0.695 & \cellcolor[HTML]{FEFEFE}0.719 & \cellcolor[HTML]{FEFCFA}0.685 & \cellcolor[HTML]{FAFDFE}0.750 & \cellcolor[HTML]{EFF6FA}0.817 & \cellcolor[HTML]{FEF8F5}0.643 & \cellcolor[HTML]{FEF9F6}0.651 & \cellcolor[HTML]{FEF7F3}0.627 & \cellcolor[HTML]{FEFBF9}0.678 & -                             & \cellcolor[HTML]{FEF2EC}0.571 \\
Log-Rank                 & \cellcolor[HTML]{FDFEFF}0.736 & \cellcolor[HTML]{FEFEFD}0.709 & \cellcolor[HTML]{F3F8FB}0.795 & \cellcolor[HTML]{FEF9F6}0.655 & \cellcolor[HTML]{FEF4EF}0.596 & \cellcolor[HTML]{FDFEFF}0.732 & \cellcolor[HTML]{FEFCFA}0.688 & \cellcolor[HTML]{FEF9F6}0.650 & \cellcolor[HTML]{F7FAFD}0.773 & \cellcolor[HTML]{FEF8F4}0.639 & \cellcolor[HTML]{FEF6F1}0.615 & \cellcolor[HTML]{FEFDFD}0.704 & \cellcolor[HTML]{FEF9F5}0.648 & \cellcolor[HTML]{FEF4EE}0.591 & \cellcolor[HTML]{FEFEFD}0.708 \\
Rank-GLTR                & \cellcolor[HTML]{FFFFFF}0.720 & \cellcolor[HTML]{F9FCFD}0.759 & \cellcolor[HTML]{F2F8FB}0.802 & \cellcolor[HTML]{FEF9F6}0.655 & \cellcolor[HTML]{FEFDFC}0.701 & \cellcolor[HTML]{F1F7FB}0.808 & \cellcolor[HTML]{FEF5EF}0.600 & \cellcolor[HTML]{FDFEFF}0.734 & \cellcolor[HTML]{F3F8FB}0.795 & \cellcolor[HTML]{FEF9F7}0.658 & \cellcolor[HTML]{FEFCFB}0.693 & \cellcolor[HTML]{FEFEFE}0.713 & \cellcolor[HTML]{FEF6F2}0.620 & \cellcolor[HTML]{FEFBF9}0.679 & \cellcolor[HTML]{FEFCFB}0.694 \\
Fast-DetectGPT           & \cellcolor[HTML]{EFF6FA}0.817 & \cellcolor[HTML]{EFF6FA}0.817 & \cellcolor[HTML]{E4F0F6}0.887 & \cellcolor[HTML]{F9FCFD}0.760 & \cellcolor[HTML]{F9FCFD}0.759 & \cellcolor[HTML]{EBF4F9}0.842 & \cellcolor[HTML]{EBF4F9}0.842 & \cellcolor[HTML]{F2F8FB}0.801 & \cellcolor[HTML]{E2EFF6}0.899 & \cellcolor[HTML]{FEFCFA}0.688 & \cellcolor[HTML]{FEFEFE}0.718 & \cellcolor[HTML]{FAFCFE}0.752 & \cellcolor[HTML]{FEFBF9}0.677 & \cellcolor[HTML]{FEFEFE}0.713 & \cellcolor[HTML]{F9FCFE}0.756 \\
Binoculars               & \cellcolor[HTML]{E5F0F7}0.881 & \cellcolor[HTML]{E2EFF6}0.897 & \cellcolor[HTML]{E0EEF5}0.911 & \cellcolor[HTML]{EDF5F9}0.833 & \cellcolor[HTML]{EBF4F9}0.845 & \cellcolor[HTML]{E3EFF6}0.890 & \cellcolor[HTML]{DEECF4}0.923 & \cellcolor[HTML]{E7F2F8}0.867 & \cellcolor[HTML]{DFEDF5}0.916 & \cellcolor[HTML]{FEFEFD}0.710 & \cellcolor[HTML]{F7FBFD}0.772 & \cellcolor[HTML]{F2F8FB}0.800 & \cellcolor[HTML]{FEFBFA}0.680 & \cellcolor[HTML]{F2F8FB}0.803 & \cellcolor[HTML]{F4F9FC}0.792 \\ \midrule
RADAR                    & \cellcolor[HTML]{FDEADF}0.470 & \cellcolor[HTML]{FEEFE6}0.524 & \cellcolor[HTML]{FEF3EC}0.574 & \cellcolor[HTML]{FDE4D5}0.396 & \cellcolor[HTML]{FEEFE6}0.530 & \cellcolor[HTML]{FEF5F0}0.603 & \cellcolor[HTML]{FDE7D9}0.427 & \cellcolor[HTML]{FEF2EB}0.564 & \cellcolor[HTML]{FEF4EF}0.597 & \cellcolor[HTML]{FDE6D9}0.426 & \cellcolor[HTML]{FEF5F0}0.609 & \cellcolor[HTML]{FEFBF8}0.672 & \cellcolor[HTML]{FEF0E9}0.547 & \cellcolor[HTML]{F3F9FC}0.793 & \cellcolor[HTML]{FDFEFF}0.736 \\
ChatGPT-D                & \cellcolor[HTML]{FEF1EA}0.557 & \cellcolor[HTML]{FEF1E9}0.552 & \cellcolor[HTML]{FEFEFE}0.712 & \cellcolor[HTML]{FDE9DC}0.452 & \cellcolor[HTML]{FEEFE6}0.526 & \cellcolor[HTML]{FEF8F5}0.642 & \cellcolor[HTML]{FEEFE7}0.531 & \cellcolor[HTML]{FEF8F5}0.643 & \cellcolor[HTML]{FCFDFE}0.743 & \cellcolor[HTML]{FDDBC7}0.280 & \cellcolor[HTML]{FDDECC}0.320 & \cellcolor[HTML]{FDE9DD}0.454 & \cellcolor[HTML]{FDE9DD}0.458 & \cellcolor[HTML]{FEFBF9}0.679 & \cellcolor[HTML]{FEF7F2}0.625 \\ \midrule
DistillBert-F            & \cellcolor[HTML]{D3E6F1}0.987 & \cellcolor[HTML]{D4E7F1}0.983 & \cellcolor[HTML]{D6E8F2}0.971 & \cellcolor[HTML]{D5E7F2}0.977 & \cellcolor[HTML]{D4E7F1}0.983 & \cellcolor[HTML]{D5E8F2}0.976 & \cellcolor[HTML]{D3E6F1}0.988 & \cellcolor[HTML]{D2E6F1}0.991 & \cellcolor[HTML]{D3E6F1}0.990 & \cellcolor[HTML]{D4E7F1}0.983 & \cellcolor[HTML]{D3E6F1}0.988 & \cellcolor[HTML]{D4E7F1}0.982 & \cellcolor[HTML]{D4E7F1}0.983 & \cellcolor[HTML]{D4E7F1}0.979 & \cellcolor[HTML]{D7E8F2}0.966 \\
Roberta-F                & \cellcolor[HTML]{D3E6F1}0.987 & \cellcolor[HTML]{D2E6F1}0.994 & \cellcolor[HTML]{D2E6F1}0.994 & \cellcolor[HTML]{D2E6F1}0.992 & \cellcolor[HTML]{D1E5F0}0.997 & \cellcolor[HTML]{D2E6F1}0.995 & \cellcolor[HTML]{D2E6F1}0.993 & \cellcolor[HTML]{D2E6F1}0.992 & \cellcolor[HTML]{D2E6F1}0.997 & \cellcolor[HTML]{D3E6F1}0.987 & \cellcolor[HTML]{D2E6F1}0.993 & \cellcolor[HTML]{D2E6F1}0.994 & \cellcolor[HTML]{D3E7F1}0.986 & \cellcolor[HTML]{D3E7F1}0.986 & \cellcolor[HTML]{D4E7F1}0.981 \\\bottomrule\end{tabular}\end{table*}
\begin{table*}[]
\centering
\setlength{\tabcolsep}{3pt}
\renewcommand{\arraystretch}{1.1}
\caption{\textbf{Experiment Result for In-distribution Text Attribution.} 
We train and test the model on the same data domain.
The results are reported using the F1 score.
The larger values with blue colors indicate better performance and lower values with red colors indicate weaker performance.
}
\label{tab:attribution-benchmark}
\resizebox{0.9\textwidth}{!}{\begin{tabular}{cccccccccc}\toprule
               & LL                             & Entropy                        & Rank                           & LRR                            & Rank-GLTR                      & Fast-Detect                    & Binoculars                     & DistillBert-F                  & RoBERTa-F                      \\ \midrule
STEM           & \cellcolor[HTML]{FEF1EA}0.1661 & \cellcolor[HTML]{FFFFFF}0.1941 & \cellcolor[HTML]{FDE4D6}0.1392 & \cellcolor[HTML]{FEEDE4}0.1581 & \cellcolor[HTML]{FDE5D7}0.1412 & \cellcolor[HTML]{FEFFFF}0.2137 & \cellcolor[HTML]{FEFFFF}0.2188 & \cellcolor[HTML]{D4E7F1}0.8444 & \cellcolor[HTML]{D1E5F0}0.8881 \\
Humanities     & \cellcolor[HTML]{FEF8F5}0.1806 & \cellcolor[HTML]{FDE6D8}0.1422 & \cellcolor[HTML]{FDE9DD}0.1481 & \cellcolor[HTML]{FDDBC7}0.1182 & \cellcolor[HTML]{FDE2D2}0.1334 & \cellcolor[HTML]{FDFEFF}0.2251 & \cellcolor[HTML]{FDE9DD}0.1483 & \cellcolor[HTML]{D8E9F3}0.7835 & \cellcolor[HTML]{D6E8F2}0.8151 \\
Social Science & \cellcolor[HTML]{FEFFFF}0.2151 & \cellcolor[HTML]{FDFEFF}0.2282 & \cellcolor[HTML]{FEEFE7}0.1624 & \cellcolor[HTML]{FEFCFB}0.1894 & \cellcolor[HTML]{FEFDFC}0.1912 & \cellcolor[HTML]{F9FCFE}0.2849 & \cellcolor[HTML]{FEFFFF}0.2141 & \cellcolor[HTML]{D6E8F2}0.8174 & \cellcolor[HTML]{D6E8F2}0.8177 \\\bottomrule
\end{tabular}
}
\end{table*}

\mypara{Adapting to New Classes}
Adapting to new classes with few-shot settings is related to few-shot learning, aiming to improve the quick adaptation ability.
One way is to use the distance between the samples and the representatives of each class for classification~\cite{snell2017prototypicalnetworksfewshotlearning, chen2019closer}.
Another way is to train a neural network and learn the relationship between samples and the representatives~\cite{sung2018learning}.
Additionally, data augmentation is used to increase the number of training samples and train a classifier~\cite{yang2021free}
Adapting to new classes with many-shot settings is related to class incremental learning (CIL), where the key is to alleviate the forgetting of previous knowledge~\cite{masana2022class, zhou2023revisiting}.
One way to improve the performance is to incorporate a distillation loss or regularization term to transfer knowledge from the old model to the updated one, thus reducing forgetting~\cite{li2017learning, riemer2018learning}.
Another way is to store a small subset of past representative data to enable the model to rehearse earlier tasks~\cite{rebuffi2017icarl}.
Additionally, some other work explores calibrating the output layer of the classification head to improve the performance~\cite{wu2019large}.

Some efforts have discussed the CIL in classification tasks such as entailment or intent classification~\cite{DBLP:conf/naacl/XiaYFY21, paul-etal-2022-class}.
To the best of our knowledge, our work is the first to benchmark the adaptation ability of MGT detectors in both few-shot and many-shot settings.

\section{Construction of \Dataset}

\subsection{\Dataset Collection}

In this section, we describe the sources and principles for collecting both HWTs and MGTs.

\mypara{Human Data}
Since exploiting LLMs brings great convenience to academic usage, e.g., generating ideas, drafting articles, or refining content, we collect data in three academic domains, i.e., STEM, Social Science, and Humanity, where each domain contains different fine-grained fields.
To collect data in each field, we leverage the existing classification structure in the source datasets. 
We use the Arxiv API to download papers from specific categories.
For Gutendex, we query specific bookshelves related to topics like History. 
For Wikipedia, we query the first-level content of categories, such as ART/*, to ensure proper domain-specific collection. 
Each field consists of Wikipedia and academic texts, with the content collected from Arxiv (pre-print papers) or Project Gutenberg (published e-books), depending on the scenario.
This method allows us to systematically map the datasets to the STEM, humanities, and social sciences sub-domains. 
The details are shown in \Cref{tab:data-mgt}.

\mypara{Machine Data}
We select five widely used LLMs, including Llama-3.1-70b-Instruct~\cite{TMSAABBBBBBBCCCEFFFFGGGHHHIKKKKKKLLLLLMMMMMNPRRSSSSSTTTWKXYZZFKNRSES23}, Mixtral-8$\times$7b-Instruct~\cite{Mistral}, KimiChat~\cite{Moonshot}, ChatGPT, GPT-4omini~\cite{chatgpt} to generate the MGTs.
Llama-3.1-70b-Instruct and Mixtral-8$\times$7b-Instruct are two commonly used open-source LLMs that exploit dense and MoE architecture respectively.
Moonshot, ChatGPT, and GPT-4omini are popular proprietary models, with Moonshot known for its long-context understanding and the GPT family recognized for its comprehensive capabilities.
We prompt the LLM to be a wiki/paper/book editor and polish the given human text, which is consistent with the previous dataset~\cite{MLKMF23,YJPJAAMTGTo24}.
The prompts for generating MGT data are listed in \Cref{sec-app:prompt}.
Other details of dataset construction are listed in \Cref{sec-app:moderate}.
Additionally, We incorporate linguistic metrics for a more in-depth analysis of the data, including readability, diversity, and syntactic complexity.
The metrics are shown in~\Cref{sec-app:data}.

\subsection{\Dataset Analysis}

We conduct further analysis of text length, embeddings, and key-words on MGT-Academic to provide more insights into understanding the MGT detection task, which is shown in Appendix C. We find that data from different domains and LLMs exhibit distinct distributions across all these aspects. These differences highlight the importance of studying the generalization ability of MGT detectors. Besides, our evaluation shows that MGT texts are less readable than HWT, but have greater vocabulary diversity and simpler sentence structures. These differences highlight the need for MGT detection models to adapt to the varying characteristics of machine-generated and human-written texts for better generalization.

\subsection{Code Framework}

Our framework follows the factory design pattern and implements \textit{AutoDetector} and \textit{AutoExperiment} for abstraction, which is aligned with the approach used in Huggingface Transformers~\cite{WDSCDMCRLFB19}, the most widely used library in NLP community.
It provides an easy-to-use and extendable code framework for the community and is publicly available. 
More design details are in \Cref{sec-app:code}.

\section{Benchmarking MGT Detectors}

We benchmark the performance of detectors in binary classification tasks, which predicts binary labels (human or machine) for the setting where training and testing data are in the same domain.
Then, we benchmark the performance in text attribution tasks, which aims to tell whether the input text is human-written or generated by a specific LLM.
Compared to the binary task, the text attribution task appears to be a more challenging multi-class classification task.

\subsection{In-distribution Performance}
\label{sec:in-dis}

\mypara{Experiment Settings}
For each domain, we first sample the same number of HWTs and MGTs from the corresponding domains, then randomly split them into the train/test dataset with an 80\%/20\% ratio.
For the evaluation metric, we select the F1 score, which balances precision and recall and is robust against class imbalance.
The experimental details are shown in \Cref{sec-app:hyperparameter}.
Regarding the detectors, we benchmark various detectors on \Dataset, covering both metric-based and model-based approaches.

For metric-based detectors, we evaluate Log-Likelihood, Entropy, Rank, Rank-GLTR \cite{GSR19}, LRR \cite{JYDP23}, Fast-DetectGPT \cite{BZTYZ24}, and Binoculars \cite{binoculars}.
We use Llama-2-7B as the default white-box model, determining an optimal threshold to maximize the F1 score for binary classification or training a logistic regression classifier for text attribution.
For model-based detectors, we include RADAR \cite{hu2023radar}, ChatGPT-D \cite{GZWJNDYW23}, DistillBERT \cite{distilbert}, and RoBERTa \cite{roberta}.
RADAR and ChatGPT-D use their officially released model weights, while DistillBERT and RoBERTa are fine-tuned with classification heads on \Dataset.
Further details are provided in \Cref{sec-app:detecors}.

\begin{table*}[]
\centering
\caption{\textbf{Experiment Result for Transferring Across Different Domains in Text Attribution.} 
We train the model on data in one domain and test the model on another domain.
The results are reported using F1 score.
The larger values with blue colors indicate better performance and lower values with red colors indicate weaker performance.}
\label{tab:transfer-attribution}
\setlength{\tabcolsep}{3pt}
\renewcommand{\arraystretch}{1.1}

\resizebox{\textwidth}{!}{
\begin{tabular}{c|c|ccc|c|c|ccc|c|c|ccc} \toprule
\multicolumn{1}{l|}{}       & \multicolumn{1}{l|}{} & Humanity                           & STEM                           & Social Science                           & \multicolumn{1}{l|}{}           & \multicolumn{1}{l|}{} & Humanity                           & STEM                           & Social Science                           & \multicolumn{1}{l|}{}           & \multicolumn{1}{l|}{} & Humanity                           & STEM                           & Social Science                           \\ \midrule
                            & Humanity                   & \cellcolor[HTML]{FEF0E8}0.181 & \cellcolor[HTML]{FDE8DC}0.160 & \cellcolor[HTML]{FEFAF7}0.204 &                                 & Humanity                   & \cellcolor[HTML]{FDE4D5}0.148 & \cellcolor[HTML]{FDE6D9}0.155 & \cellcolor[HTML]{FEF1E9}0.182 &                                 & Humanity                   & \cellcolor[HTML]{FFFFFF}0.225 & \cellcolor[HTML]{FFFFFF}0.221 & \cellcolor[HTML]{FDFEFF}0.250 \\
                            & STEM                   & \cellcolor[HTML]{FDE8DB}0.158 & \cellcolor[HTML]{FDEBE0}0.166 & \cellcolor[HTML]{FEF3EC}0.186 &                                 & STEM                   & \cellcolor[HTML]{FEF6F1}0.195 & \cellcolor[HTML]{FFFFFF}0.219 & \cellcolor[HTML]{FEFEFF}0.244 &                                 & STEM                   & \cellcolor[HTML]{FEFDFC}0.213 & \cellcolor[HTML]{FEFDFD}0.214 & \cellcolor[HTML]{FDFEFF}0.247 \\
\multirow{-3}{*}{LL}        & Social Science                   & \cellcolor[HTML]{FEEEE5}0.176 & \cellcolor[HTML]{FEF2EB}0.185 & \cellcolor[HTML]{FEFEFD}0.215 & \multirow{-3}{*}{Binoculars}    & Social Science                   & \cellcolor[HTML]{FEEFE7}0.178 & \cellcolor[HTML]{FEF4EF}0.191 & \cellcolor[HTML]{FEFDFD}0.214 & \multirow{-3}{*}{FastDetectGPT} & Social Science                   & \cellcolor[HTML]{FDFEFF}0.251 & \cellcolor[HTML]{FEFFFF}0.241 & \cellcolor[HTML]{FBFDFE}0.285 \\ \midrule
                            & Humanity                   & \cellcolor[HTML]{FDDECC}0.133 & \cellcolor[HTML]{FDDCC9}0.128 & \cellcolor[HTML]{FDEBE0}0.167 &                                 & Humanity                   & \cellcolor[HTML]{D8E9F3}0.784 & \cellcolor[HTML]{E6F1F7}0.592 & \cellcolor[HTML]{DDECF4}0.715 &                                 & Humanity                   & \cellcolor[HTML]{D6E8F2}0.815 & \cellcolor[HTML]{E0EEF5}0.674 & \cellcolor[HTML]{DAEAF3}0.762 \\
                            & STEM                   & \cellcolor[HTML]{FDDBC7}0.125 & \cellcolor[HTML]{FDE1D0}0.141 & \cellcolor[HTML]{FDE7DA}0.158 &                                 & STEM                   & \cellcolor[HTML]{E2EFF6}0.651 & \cellcolor[HTML]{D4E7F1}0.844 & \cellcolor[HTML]{D9E9F3}0.781 &                                 & STEM                   & \cellcolor[HTML]{DCECF4}0.727 & \cellcolor[HTML]{D1E5F0}0.883 & \cellcolor[HTML]{D6E8F2}0.814 \\
\multirow{-3}{*}{Rank GLTR} & Social Science                   & \cellcolor[HTML]{FDE6D8}0.153 & \cellcolor[HTML]{FDEBE0}0.167 & \cellcolor[HTML]{FEF4EF}0.191 & \multirow{-3}{*}{DistillBert-F} & Social Science                   & \cellcolor[HTML]{DDECF4}0.718 & \cellcolor[HTML]{D5E8F2}0.828 & \cellcolor[HTML]{D6E8F2}0.817 & \multirow{-3}{*}{Roberta-F}     & Social Science                   & \cellcolor[HTML]{DCECF4}0.727 & \cellcolor[HTML]{D5E8F2}0.827 & \cellcolor[HTML]{D6E8F2}0.818\\\bottomrule
\end{tabular}}
\end{table*}

\mypara{Binary Classification Task}
The performance of the binary classification task is shown in \Cref{tab:binary-benchmark}.
First, we observe that supervised model-based detectors consistently outperform other methods, achieving F1 scores above 0.98. 
This advantage is largely due to the availability of extensive supervised training data, enabling these detectors to learn highly effective classification boundaries.
Second, model-based detectors using released weights show comparatively weaker performance, as they were trained on smaller and less diverse datasets. 
Previous study~\cite{HSCBZ23} has demonstrated that these detectors can achieve comparable performance to supervised models after fine-tuning with a dedicated training dataset. 
This highlights the inherent challenges in the generalization ability of model-based detectors.
Third, metric-based detectors show notable improvement over time, with state-of-the-art (SOTA) methods like Fast-DetectGPT and Binoculars approaching the performance of supervised model-based detectors. 
Despite their advancements, these state-of-the-art metric-based detectors show only moderate performance on outputs from GPT-4omini and GPT-3.5.
This may stem from the inability of the metric generators to effectively extract distinguishing features from these datasets.

\mypara{Attribution Task}
Since RADAR and ChatGPT-D are not designed for the attribution task, they are excluded from this evaluation. 
The remaining results for the attribution task are presented in \Cref{tab:attribution-benchmark}.
As the attribution task is more challenging, the performance is overall lower than that of binary classification. 
The metric-based detectors show almost no capability for text attribution, performing at a level close to random guessing.
The performance of model-based detectors is better but there is still space to improve.
These findings highlight that text attribution is a critical yet underexplored task.

We conduct additional investigations for the failure of metric-based detectors in attribution tasks.
Since the dimension of the feature in metric-based detectors is low (generally one), it without doubt performs poorly for multi-classification problems.
To support our analysis, we concatenate five 1-D features from metric-based detectors (ll, logrank, entropy, FastDetectGPT and Binoculars) and run XGBboost for classification.
The combined results are shown in \Cref{tab:metric-study}.
The performance improved greatly compared to the best baseline in metric-based detectors.
It partially explains the failure of metric-based detectors in attribution tasks and implies the challenge in attribution tasks for metric-based detectors.
\begin{table}[h]
\centering
\caption{Experiment result for combining the metrics in attribution task. The results are reported using F1 score.}
\label{tab:metric-study}
\begin{tabular}{cccc}\toprule
Fast-Detect & 0.2137 & {\ul{ 0.2251}} & {\ul{ 0.2849}} \\
Binoculars & {\ul{0.2188}} & 0.1483 & 0.2141 \\ \midrule
Combined & \textbf{0.3528} & \textbf{0.3560} & \textbf{0.3624} \\
Improve & 61\% & 88\% & 36\% \\\bottomrule
\end{tabular}
\end{table}

\begin{figure}[htbp]
    \centering
    \includegraphics[width=\linewidth]{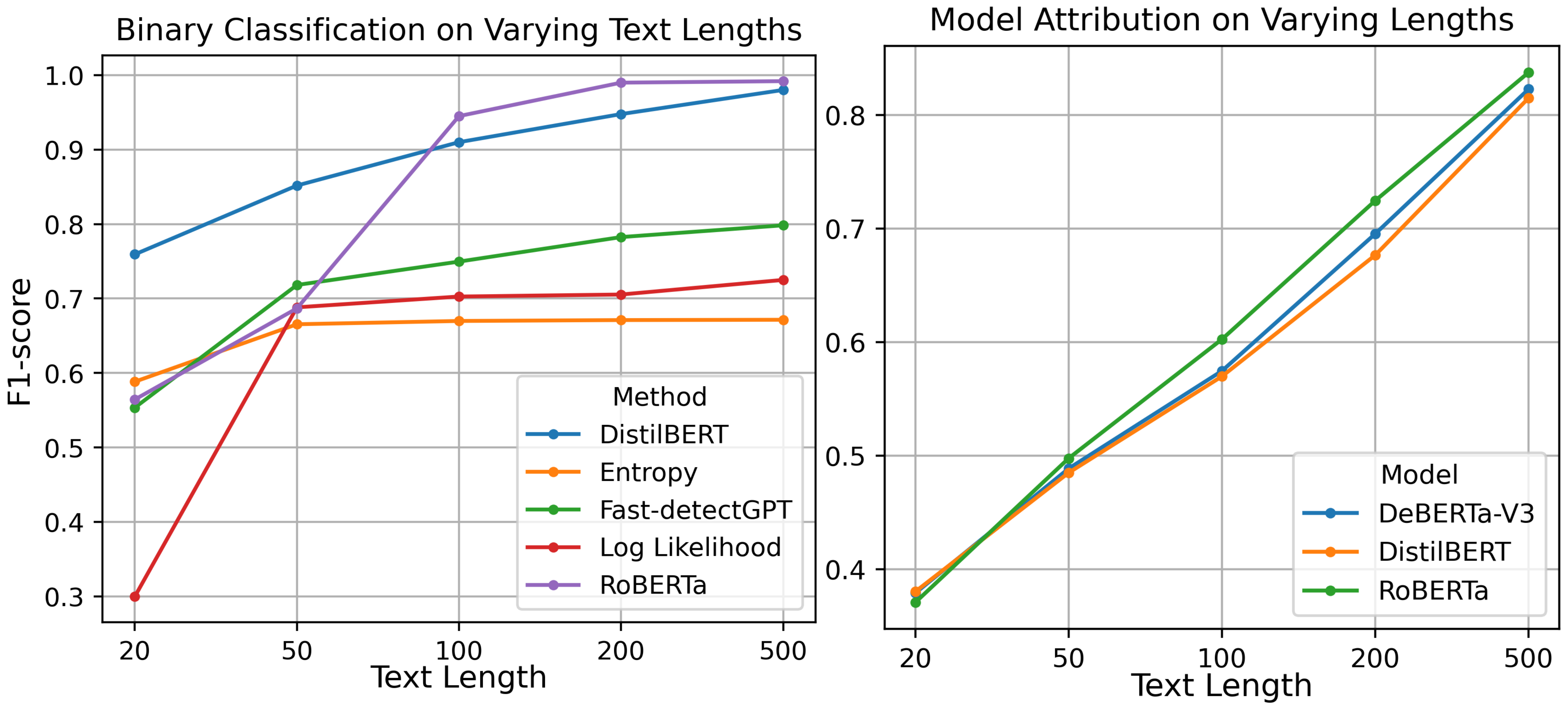}
    \caption{\textbf{The F1-score of different detectors under texts of varying lengths.}}
    \label{fig:perf_length}
\end{figure}

Additionally, we conduct an ablation study for two important factors in metric-based detectors, i.e., the LLM metric generator and the classifier, and the results are shown in \Cref{sec-app:ablate_bench}.
We investigate the effect of text length on detectors' performance in \Cref{fig:perf_length}. 
The binary classification requires roughly 100 words to have decent performance while the attribution tasks require more words, which implies the challenges in attribution task as well.

\mypara{Takeaways}
We benchmark the performance of detectors in binary classification and text attribution tasks. 
For the binary classification task, model-based detectors consistently perform better than metric-based detectors with the same distributions in train and test data.
For the attribution task, the model-based detectors show competitive performance while the metric-based detectors perform poorly (near random guessing), which is caused by the low dimension of the features.
Future works are encouraged to develop metric-based detectors suitable for attribution tasks.

\subsection{On the Generalization Ability in Domain Transferring}

We conduct comprehensive experiments under domain-transferring settings to show the generalization ability of different kinds of detectors.
First, for the binary classification task, we consider two cases where the domain or LLM during training and testing data are different.
Second, for the attribution tasks, we only consider that the data domains during training and testing are different.
Third, we implement a few-shot domain adaptation technique to mitigate the performance drop.

\mypara{Experiment Settings}
For each domain, we first select the same number of HWTs and MGTs from the corresponding domains or LLMs, then randomly split them into the train/test dataset with an 80\%/20\% ratio.
During training, the detectors get an optimal threshold on the data of the source domain/LLM, which will be used in the testing stage to predict the data of the target domain/LLM.

Specifically, we only choose the representative detectors of different kinds, i.e., LL, RankGLTR, FastDetectGPT, Binoculars in metric-based detectors, and DistillBert, RoBERTa in supervised model-based detectors.
We report the results using F1 score and the training hyperparameters are shown in \Cref{sec-app:hyperparameter}.
\begin{figure*}[tbp]
    \centering
    \includegraphics[width=\textwidth]{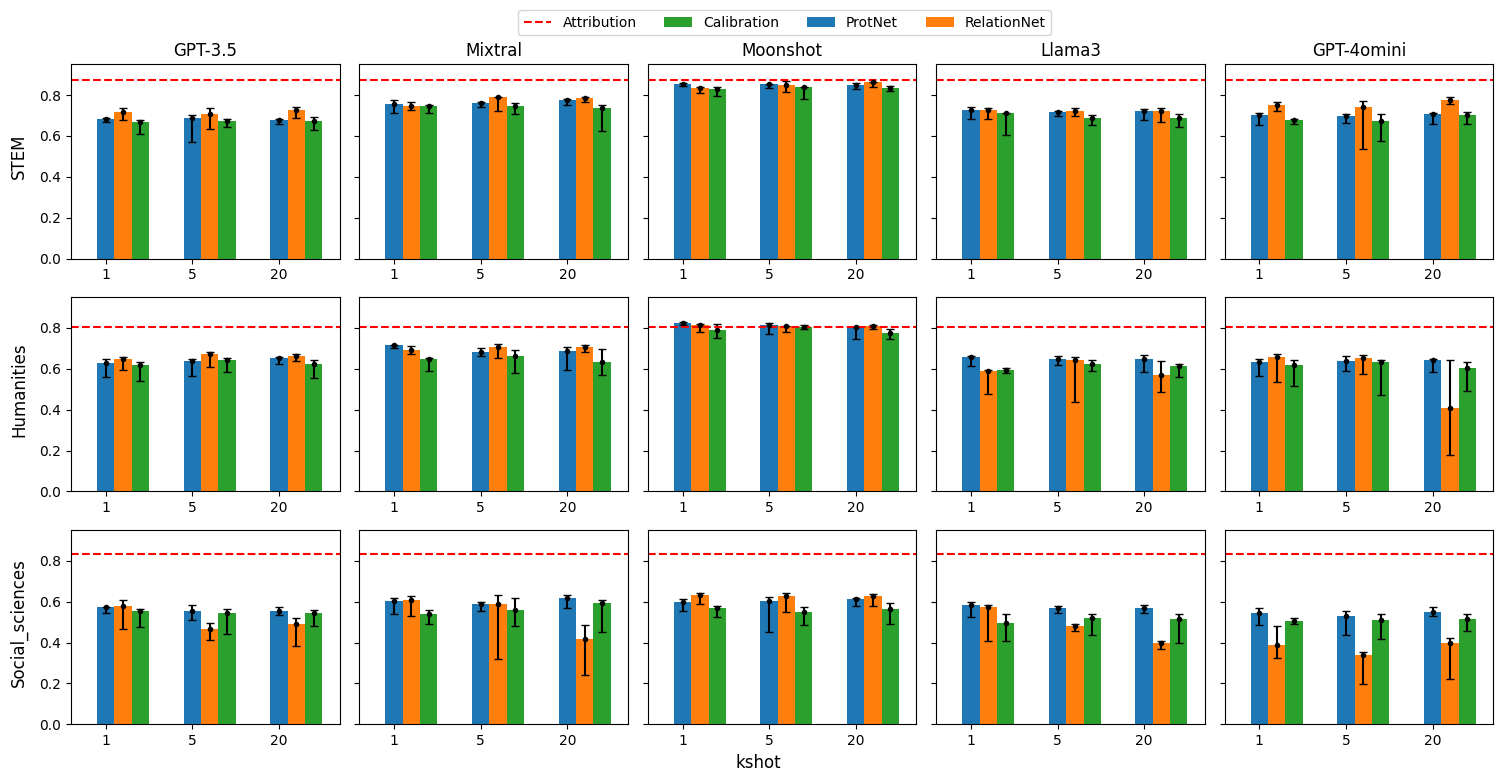}
    \caption{Experiment results for adapting to new LLM in the few-shot setting, where only one class is introduced.
    Kshot means the number of examples in the newly introduced class and previous classes.
    The title of each column is the class that the newly introduced LLM.
    The detector is built on the DeBerta-v3-base.
    }
    \label{fig:few-deberta-51}
\end{figure*}

\mypara{Transferring in Binary Classification Task}
The domain transferring performance in binary tasks across different domains and LLMs is summarized in \Cref{tab:binary-transfer-domain} and \Cref{tab:binary-transfer-LLM}, respectively. 
We observe that, regarding domain transferring, zero-shot metric-based detectors exhibit limited robustness in domain adaptation. 
The threshold values for distinguishing between MGT and HWT vary significantly across domains, making consistent performance difficult to achieve.
Regarding LLM transferring, fine-grained results for each domain are detailed in \Cref{sec-app:transfer}, and the average performance is reported in \Cref{tab:binary-transfer-LLM}.
Zero-shot metric-based detectors struggle notably when predicting results for texts generated by the GPT family of LLMs, highlighting a specific area for improvement.
While supervised model-based detectors continue to lead overall, domain adaptation across different LLMs is a more challenging task.
Similarly, we conduct the ablation study for two important factors in metric-based detectors, and the results are shown in \Cref{sec-app:transfer}.

\mypara{Transferring in Attribution Task}
The domain transferring performance in attribution tasks across different domains is summarized in \Cref{tab:transfer-attribution}.
We have several observations.
First, the metric-based detectors still perform poorly, while the model-based detectors show acceptable performance.
Second, the supervised model-based detectors show worse transferability than binary classification tasks.
Since attribution tasks require the representations to capture more sophisticated features, it is harder for detectors to generate embeddings for transferring.

\begin{figure}[h!]
    \centering
    \includegraphics[width=\linewidth]{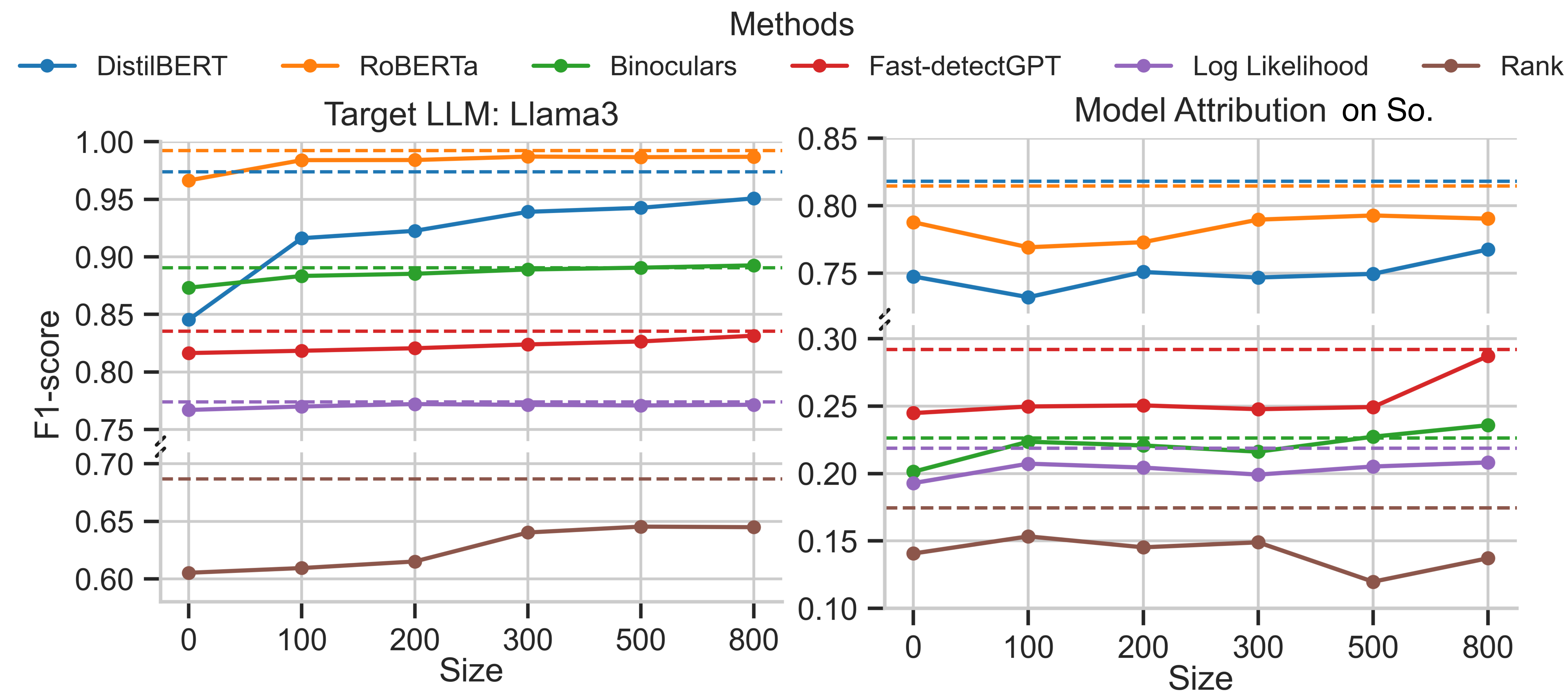}
    \caption{\textbf{Mitigation Result for Domain  Transferring.} 
    The left figure shows the result of transferring to Llama-3.1's outputs in binary classification.
    The right figure shows the result of transferring to the Social Science domain in the attribution task.}
    \label{fig:transfer_mitigate}
\end{figure}

\mypara{Mitigating Techniques}
To enhance detector transferability, We incorporate limited labeled examples from the target domain.
\Cref{fig:transfer_mitigate} reports the average performance for binary classification when transferring to Llama-3.1's outputs (left) and model attribution when transferring to the Social Science domain (right).
Full results are provided in \Cref{sec-app:transfer}.
The results demonstrate that adding labeled samples from the target domain improves the performance of supervised model-based detectors.
For the binary classification (attribution), the performance improves by 13.2\% (3.7\%) on average.
Interestingly, zero-shot metric-based detectors also show enhanced performance, suggesting that the derived threshold may be domain-specific.

\mypara{Takeaways}
We evaluate the transferability of detectors in binary classification and attribution tasks.
The results indicate transferring across different LLMs in binary and transferring across different domains in attribution tasks may suffer from severe performance degradation.
Adding data from the target domain can improve the performance in binary classification settings while showing limited effect on transferring in attribution.
Future works are encouraged to develop robust detectors for domain adaptation. 
\begin{table*}[ht]
\centering
\caption{\textbf{Experiment result for adapting to new LLM in many shot setting.} 
We split the data into two parts: one part consists of five classes of data to train the original model and the other consists of one new class of data for adaptation, which is indicated by \textbf{Last Model}.
The results are reported using the F1 score.
The larger values with blue colors indicate better performance and lower values with red colors indicate weaker performance.}
\label{tab:cil-main}
\resizebox{\textwidth}{!}{\begin{tabular}{cc|cccccc|cccccc} \toprule
\multicolumn{1}{c|}{} &  & \multicolumn{6}{c|}{DistilBert} & \multicolumn{6}{c}{RoBERTa} \\ \cmidrule{3-14} 
\multicolumn{1}{c|}{\multirow{-2}{*}{Domain}} & \multirow{-2}{*}{\textbf{Last Model}} & Normal & LwF & iCaRL & BiC & \multicolumn{1}{c|}{Combine} & Attribution & Normal & LwF & iCaRL & BiC & \multicolumn{1}{c|}{Combine} & Attribution \\ \midrule
\multicolumn{1}{c|}{} & GPT-3.5 & \cellcolor[HTML]{FDE5D7}0.5459 & \cellcolor[HTML]{FDE5D7}0.545 & \cellcolor[HTML]{FEF7F2}0.6013 & \cellcolor[HTML]{FEF4EE}0.5923 & \multicolumn{1}{c|}{\cellcolor[HTML]{FEF4EE}0.5923} &  & \cellcolor[HTML]{FDE3D3}0.5380 & \cellcolor[HTML]{FDDBC7}0.5114 & \cellcolor[HTML]{FEF6F1}0.5985 & \cellcolor[HTML]{FEF5F0}0.5960 & \multicolumn{1}{c|}{\cellcolor[HTML]{FEF5F0}0.5966} &  \\
\multicolumn{1}{c|}{} & Mixtral & \cellcolor[HTML]{FEF9F7}0.6099 & \cellcolor[HTML]{FEF9F7}0.6099 & \cellcolor[HTML]{FEFEFE}0.6253 & \cellcolor[HTML]{FDFEFF}0.6341 & \multicolumn{1}{c|}{\cellcolor[HTML]{FDFEFF}0.6341} &  & \cellcolor[HTML]{FEFCFA}0.6168 & \cellcolor[HTML]{FEFBFA}0.6164 & \cellcolor[HTML]{F6FAFD}0.6604 & \cellcolor[HTML]{F7FBFD}0.6588 & \multicolumn{1}{c|}{\cellcolor[HTML]{F7FBFD}0.6566} &  \\
\multicolumn{1}{c|}{} & Moonshot & \cellcolor[HTML]{FDE3D4}0.5401 & \cellcolor[HTML]{FDE3D4}0.5401 & \cellcolor[HTML]{FEFDFC}0.6214 & \cellcolor[HTML]{FEFDFC}0.6215 & \multicolumn{1}{c|}{\cellcolor[HTML]{FEFDFC}0.6215} &  & \cellcolor[HTML]{FEF1EA}0.5844 & \cellcolor[HTML]{FEF2EA}0.5849 & \cellcolor[HTML]{F7FBFD}0.6580 & \cellcolor[HTML]{FCFEFE}0.6383 & \multicolumn{1}{c|}{\cellcolor[HTML]{FCFEFE}0.6383} &  \\
\multicolumn{1}{c|}{} & Llama3 & \cellcolor[HTML]{F7FAFD}0.6595 & \cellcolor[HTML]{F7FBFD}0.6591 & \cellcolor[HTML]{F9FCFD}0.6500 & \cellcolor[HTML]{F7FBFD}0.6585 & \multicolumn{1}{c|}{\cellcolor[HTML]{F7FBFD}0.6581} &  & \cellcolor[HTML]{FEFAF7}0.6108 & \cellcolor[HTML]{FEFAF7}0.6105 & \cellcolor[HTML]{F5FAFC}0.6638 & \cellcolor[HTML]{F4F9FC}0.6686 & \multicolumn{1}{c|}{\cellcolor[HTML]{F4F9FC}0.6695} &  \\
\multicolumn{1}{c|}{} & GPT-4omini & \cellcolor[HTML]{FEF0E8}0.5793 & \cellcolor[HTML]{FEF0E8}0.5798 & \cellcolor[HTML]{FEF3EC}0.5881 & \cellcolor[HTML]{FEF3ED}0.5906 & \multicolumn{1}{c|}{\cellcolor[HTML]{FEF3ED}0.5906} &  & \cellcolor[HTML]{FEF5EF}0.5951 & \cellcolor[HTML]{FEF5EF}0.5946 & \cellcolor[HTML]{FEF7F2}0.6014 & \cellcolor[HTML]{FEFAF8}0.6121 & \multicolumn{1}{c|}{\cellcolor[HTML]{FEFAF7}0.6117} &  \\ \cmidrule{2-7} \cmidrule{9-13}
\multicolumn{1}{c|}{\multirow{-6}{*}{Social Science}} & Average & \cellcolor[HTML]{FEF2EB}0.5869 & \cellcolor[HTML]{FEF2EB}0.5868 & \cellcolor[HTML]{FEFCFA}0.6172 & \cellcolor[HTML]{FEFCFB}0.6194 & \multicolumn{1}{c|}{\cellcolor[HTML]{FEFCFB}0.6193} & \multirow{-6}{*}{0.8174} & \cellcolor[HTML]{FEF3EC}0.5890 & \cellcolor[HTML]{FEF1EA}0.5835 & \cellcolor[HTML]{FDFEFF}0.6364 & \cellcolor[HTML]{FDFEFF}0.6348 & \multicolumn{1}{c|}{\cellcolor[HTML]{FDFEFF}0.6345} & \multirow{-6}{*}{0.8177} \\ \midrule
\multicolumn{1}{c|}{} & GPT-3.5 & \cellcolor[HTML]{FEF9F5}0.6077 & \cellcolor[HTML]{FEF6F2}0.5999 & \cellcolor[HTML]{FEFFFF}0.6319 & \cellcolor[HTML]{FDFEFF}0.6355 & \multicolumn{1}{c|}{\cellcolor[HTML]{FDFEFF}0.6348} &  & \cellcolor[HTML]{FEFBF9}0.6151 & \cellcolor[HTML]{F7FAFD}0.6595 & \cellcolor[HTML]{F8FBFD}0.6555 & \cellcolor[HTML]{F7FBFD}0.6579 & \multicolumn{1}{c|}{\cellcolor[HTML]{F7FBFD}0.6585} &  \\
\multicolumn{1}{c|}{} & Mixtral & \cellcolor[HTML]{FEFEFE}0.6258 & \cellcolor[HTML]{FEFEFE}0.6262 & \cellcolor[HTML]{F3F8FB}0.6742 & \cellcolor[HTML]{F5FAFC}0.6651 & \multicolumn{1}{c|}{\cellcolor[HTML]{F5F9FC}0.6668} &  & \cellcolor[HTML]{FFFFFF}0.6290 & \cellcolor[HTML]{FFFFFF}0.6286 & \cellcolor[HTML]{EFF6FA}0.6862 & \cellcolor[HTML]{EEF6FA}0.6896 & \multicolumn{1}{c|}{\cellcolor[HTML]{EFF6FA}0.6894} &  \\
\multicolumn{1}{c|}{} & Moonshot & \cellcolor[HTML]{FAFDFE}0.6459 & \cellcolor[HTML]{FAFDFE}0.6455 & \cellcolor[HTML]{DCECF4}0.7569 & \cellcolor[HTML]{E5F0F7}0.7255 & \multicolumn{1}{c|}{\cellcolor[HTML]{E3F0F6}0.7309} &  & \cellcolor[HTML]{DEECF4}0.7526 & \cellcolor[HTML]{DDECF4}0.753 & \cellcolor[HTML]{D1E5F0}0.7975 & \cellcolor[HTML]{D4E7F1}0.7898 & \multicolumn{1}{c|}{\cellcolor[HTML]{D3E7F1}0.7905} &  \\
\multicolumn{1}{c|}{} & Llama3 & \cellcolor[HTML]{FEF8F4}0.6055 & \cellcolor[HTML]{FEF8F4}0.6055 & \cellcolor[HTML]{F3F9FC}0.6710 & \cellcolor[HTML]{F5F9FC}0.6666 & \multicolumn{1}{c|}{\cellcolor[HTML]{F5F9FC}0.6667} &  & \cellcolor[HTML]{FDDECC}0.5225 & \cellcolor[HTML]{FDDFCD}0.5246 & \cellcolor[HTML]{EEF6FA}0.6920 & \cellcolor[HTML]{EDF5FA}0.6934 & \multicolumn{1}{c|}{\cellcolor[HTML]{EDF5FA}0.6935} &  \\
\multicolumn{1}{c|}{} & GPT-4omini & \cellcolor[HTML]{FEFDFC}0.6219 & \cellcolor[HTML]{FEFDFC}0.6221 & \cellcolor[HTML]{FBFDFE}0.6447 & \cellcolor[HTML]{F9FCFD}0.6498 & \multicolumn{1}{c|}{\cellcolor[HTML]{F9FCFD}0.6503} &  & \cellcolor[HTML]{FEFAF8}0.6124 & \cellcolor[HTML]{FEF9F7}0.6099 & \cellcolor[HTML]{F8FBFD}0.6529 & \cellcolor[HTML]{F4F9FC}0.6692 & \multicolumn{1}{c|}{\cellcolor[HTML]{F4F9FC}0.6699} &  \\ \cmidrule{2-7} \cmidrule{9-13}
\multicolumn{1}{c|}{\multirow{-6}{*}{STEM}} & Average & \cellcolor[HTML]{FEFDFC}0.6214 & \cellcolor[HTML]{FEFCFB}0.6198 & \cellcolor[HTML]{F2F8FB}0.6757 & \cellcolor[HTML]{F4F9FC}0.6685 & \multicolumn{1}{c|}{\cellcolor[HTML]{F4F9FC}0.6699} & \multirow{-6}{*}{0.8444} & \cellcolor[HTML]{FFFFFF}0.6263 & \cellcolor[HTML]{FDFEFF}0.6351 & \cellcolor[HTML]{EDF5F9}0.6968 & \cellcolor[HTML]{ECF4F9}0.7000 & \multicolumn{1}{c|}{\cellcolor[HTML]{ECF4F9}0.7003} & \multirow{-6}{*}{0.8881} \\ \midrule
\multicolumn{1}{c|}{} & GPT-3.5 & \cellcolor[HTML]{FEEEE5}0.5742 & \cellcolor[HTML]{FEEEE5}0.5742 & \cellcolor[HTML]{FEEDE3}0.5707 & \cellcolor[HTML]{FEF0E8}0.5805 & \multicolumn{1}{c|}{\cellcolor[HTML]{FEF0E8}0.5809} &  & \cellcolor[HTML]{FDE2D2}0.5355 & \cellcolor[HTML]{FDE2D2}0.5351 & \cellcolor[HTML]{FEF4ED}0.5913 & \cellcolor[HTML]{FEF1EA}0.5843 & \multicolumn{1}{c|}{\cellcolor[HTML]{FEF2EB}0.5860} &  \\
\multicolumn{1}{c|}{} & Mixtral & \cellcolor[HTML]{FEEEE5}0.5738 & \cellcolor[HTML]{FEEEE5}0.5744 & \cellcolor[HTML]{F8FBFD}0.6558 & \cellcolor[HTML]{F7FBFD}0.6583 & \multicolumn{1}{c|}{\cellcolor[HTML]{F7FBFD}0.6583} &  & \cellcolor[HTML]{FDE2D3}0.5367 & \cellcolor[HTML]{FDE2D3}0.5369 & \cellcolor[HTML]{F3F8FB}0.6733 & \cellcolor[HTML]{F8FBFD}0.6542 & \multicolumn{1}{c|}{\cellcolor[HTML]{F7FBFD}0.6561} &  \\
\multicolumn{1}{c|}{} & Moonshot & \cellcolor[HTML]{FEF5EF}0.5946 & \cellcolor[HTML]{FEF5EF}0.5945 & \cellcolor[HTML]{E3EFF6}0.7328 & \cellcolor[HTML]{E4F0F6}0.7303 & \multicolumn{1}{c|}{\cellcolor[HTML]{E4F0F6}0.7291} &  & \cellcolor[HTML]{F6FAFC}0.6634 & \cellcolor[HTML]{F6FAFC}0.6609 & \cellcolor[HTML]{E0EEF5}0.7431 & \cellcolor[HTML]{DCEBF4}0.7586 & \multicolumn{1}{c|}{\cellcolor[HTML]{DCEBF4}0.7586} &  \\
\multicolumn{1}{c|}{} & Llama3 & \cellcolor[HTML]{FDE2D2}0.5343 & \cellcolor[HTML]{FDE2D2}0.5346 & \cellcolor[HTML]{FBFDFE}0.6437 & \cellcolor[HTML]{FDFEFF}0.6347 & \multicolumn{1}{c|}{\cellcolor[HTML]{FDFEFF}0.6352} &  & \cellcolor[HTML]{FEF2EB}0.5852 & \cellcolor[HTML]{FEF1EA}0.5842 & \cellcolor[HTML]{F6FAFD}0.6602 & \cellcolor[HTML]{F4F9FC}0.6695 & \multicolumn{1}{c|}{\cellcolor[HTML]{F4F9FC}0.6683} &  \\
\multicolumn{1}{c|}{} & GPT-4omini & \cellcolor[HTML]{FEEDE4}0.5713 & \cellcolor[HTML]{FEEDE4}0.5714 & \cellcolor[HTML]{FEF6F2}0.6004 & \cellcolor[HTML]{FEF8F4}0.6042 & \multicolumn{1}{c|}{\cellcolor[HTML]{FEF8F4}0.6042} &  & \cellcolor[HTML]{FEF1EA}0.5839 & \cellcolor[HTML]{FEF1EA}0.5837 & \cellcolor[HTML]{FEFAF8}0.6132 & \cellcolor[HTML]{FEF4EF}0.5935 & \multicolumn{1}{c|}{\cellcolor[HTML]{FEF4EE}0.5932} &  \\ \cmidrule{2-7} \cmidrule{9-13}
\multicolumn{1}{c|}{\multirow{-6}{*}{Humanity}} & Average & \cellcolor[HTML]{FEEDE3}0.5697 & \cellcolor[HTML]{FEEDE3}0.5698 & \cellcolor[HTML]{FCFDFE}0.6407 & \cellcolor[HTML]{FBFDFE}0.6416 & \multicolumn{1}{c|}{\cellcolor[HTML]{FBFDFE}0.6415} & \multirow{-6}{*}{0.7835} & \cellcolor[HTML]{FEF0E8}0.5809 & \cellcolor[HTML]{FEF0E8}0.5802 & \cellcolor[HTML]{F7FBFD}0.6562 & \cellcolor[HTML]{F9FCFD}0.6520 & \multicolumn{1}{c|}{\cellcolor[HTML]{F8FCFD}0.6524} & \multirow{-6}{*}{0.8151} \\ \midrule
\multicolumn{2}{c|}{Overall Average} & 0.5927 & 0.5921 & 0.6445 & 0.6432 & \multicolumn{1}{c|}{0.6436} & 0.8151 & 0.5988 & 0.5996 & 0.6632 & 0.6623 & \multicolumn{1}{c|}{0.6624} & 0.8403 \\\bottomrule
\end{tabular}}
\end{table*}

\subsection{On the Adaptation to New LLMs}

Since new LLMs with different characteristics are continuously released, we study how the pre-trained detector (trained on attribution task already) would adapt to the new class introduced over time with very limited (or without) access to the original training data for earlier classes.
This scenario is common in real-world applications, where models often need to generalize to unseen distributions without retraining on past data.
Since the metric-based detectors generally show poor performance close to random guess in attribution tasks, our study mainly focuses on the supervised model-based detectors, i.e., DistilBert, RoBERTa, and DeBERTa, for this more challenging setting.
For simplicity, we mainly focus on the scenario where only one new LLM is introduced to the original detector (trained with HWTs and four types of MGTs).
Moreover, We also consider the case where the detectors must adapt to two classes (see \Cref{sec-app:cil} for more details).
To the best of our knowledge, we are the first to investigate the adaptation ability of MGT detectors.

\mypara{Experiment Settings}
During the pre-training stage, the objective is to train a five-class classifier.
We adopt similar training arguments as previous attribution evaluations.
The learning rate is set to 1e-6, and the training epoch is set to 2.
During the adaptation stage, the training data consists only of MGTs from one or two new LLMs, optionally supplemented with a limited number of samples from previous HWTs and MGTs. 
In few-shot settings, we use a few previous samples (e.g. 1, 5, or 20), while in many-shot settings, the budget remains fewer than 100 samples.
In the many-shot setting, the budget remains fewer than 100 samples. 
Meanwhile, the testing data includes a balanced set of samples from all six classes.
The new learning objective is to extend the previous five-class classifier into a six-class classifier.

For few-shot settings, the newly introduced classes only consist of a small number of examples, e.g., 1, 5, or 20 shots.
Few-shot adaptation methods do not require fine-tuning and we use the pre-trained detector as the feature extractor and evaluate three representative methods.
ProtNet~\cite{snell2017prototypicalnetworksfewshotlearning} uses the distance between the samples and the representatives of each class for classification.
RelationNet~\cite{sung2018learning} trains a neural network and learns the relationship between samples and the representatives.
Distribution Calibration~\cite{yang2021free} leverages the statistical information of base classes with sufficient data to calibrate the feature distribution of newly introduced classes, enhancing the performance of few-shot learning.

For many-shot settings, the newly introduced classes consist of the same number of examples as the previous classes.
Many-shot adaptation needs to be fine-tuned on the dataset, and we lower the learning rate to 1/4 of that in attribution in pre-training and initialize a new classification head with one extra dimension to adapt to the newly introduced class. 
This approach is widely used by previous work~\cite{li2017learning} and ensures the detector incrementally adapts to new classes.
We evaluate the performance of detectors using several widely-used adapting techniques, i.e., LwF~\cite{li2017learning}, iCaRL~\cite{rebuffi2017icarl}, BiC~\cite{wu2019large}, and combination.
The training details and techniques are provided in \Cref{sec-app:hyperparameter}.

\mypara{Few-shot Results}
The results of DeBerta-v3-base are shown in \Cref{fig:few-deberta-51} and results for DistilBert and RoBerta are shown in \Cref{sec-app:few}.
First, we evaluate the performance of few-shot techniques when there is only one class introduced.
The performance of DeBerta-V3 is more stable and consistently better than the other baselines.
Second, we find that increasing the number of shots in each class has limited improvement in performance.
The performance of Social Science is generally poor, which may reflect the inherent similarity of text generated by different LLMs.
Third, we analyze the performance of different techniques.
ProtNet generally gives the most stable performance with the smallest variance.
Since RelationNet utilizes a neural network for classification and the dimension of extracted features is high (768), the training of the neural network can be unstable and lead to high variance.
Distribution calibration first transforms the features into normal distribution and sample data points from the distribution to augment the classifier.
However, the high dimension of extracted features may lead to improper augmentation and relatively poor performance.

\begin{figure}
    \centering
    \includegraphics[width=1\linewidth]{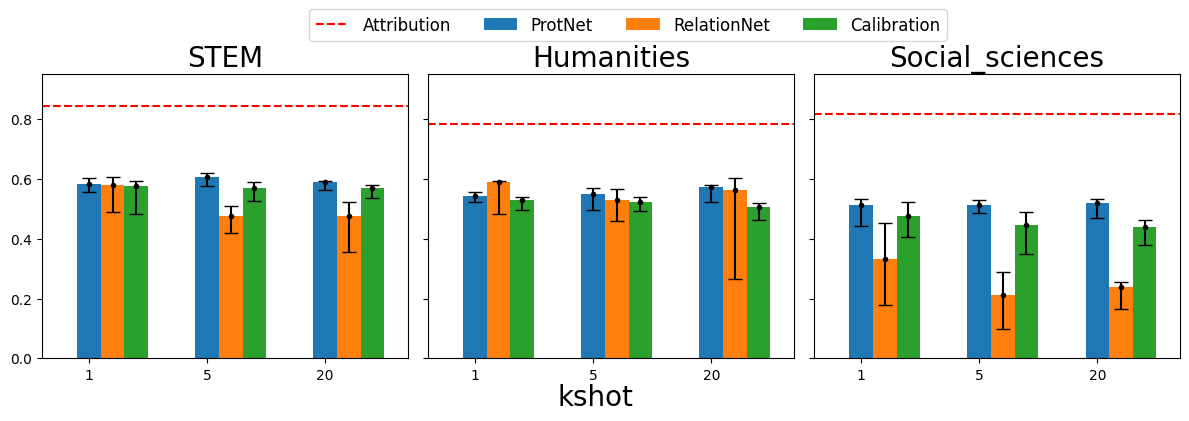}
    \caption{Experiment results for adapting to new LLM in the few-shot setting, where two new classes are introduced.
    Kshot means the number of examples in the newly introduced class and previous classes.
    The title of each column is the class that the newly introduced LLM.
    The detector is built on DeBerta-v3-base.}
    \label{fig:few-deberta-42}
\end{figure}

Furthermore, we study the scenario where two new LLMs are introduced.
The results of DeBerta-v3-based are shown in \Cref{fig:few-deberta-42}, and the remaining results are shown in \Cref{sec-app:few}.
We only consider a practical case where the two LLMs are the latest released models, i.e., Llama-3 and GPT-4omini.
The performance drops rapidly and highlights the perplexity of the task.

\mypara{Many-shot Results}
The results are shown in \Cref{tab:cil-main} and \Cref{sec-app:cil}.
First, we evaluate the performance of two supervised models equipped with five different techniques during the update stage.
The results reveal a performance drop compared to the standard six-class attribution task, primarily due to catastrophic forgetting~\cite{li2017learning} as we mentioned in \Cref{sec:related}. 
Among the models, RoBERTa consistently outperforms the other, reflecting its stronger context-understanding ability.
Second, employing different techniques can effectively mitigate performance degradation. 
For instance, the performance of RoBERTa improves by 1\%, 10.75\%, 10.59\%, and 10.63\% respectively, using the four shown techniques.
ICaRL and BiC techniques demonstrate competitive results, with each excelling in specific domains or LLMs.
Despite these improvements, the results remain below the upper-bound performance observed in standard attribution tasks, underscoring the challenges of adapting to new LLMs.

Furthermore, we extend the evaluation to the scenario where two new LLMs are introduced. 
Specifically, since exhausting all possible cases is resource-consuming and brings limited insights, we only consider a practical case where the two LLMs are the latest released models.
The results presented in \Cref{tab:cil-42} indicate that adding more LLMs rapidly increases the complexity. 
These findings highlight the inherent challenges in many-shot settings and emphasize the need for the exploration of scalable techniques.

\mypara{Takeaways}
We introduce a new attribution task setting that adapts a detector to new LLMs with very limited access to the previous data with few-shot and many-shot settings.
Both few-shot and many-shot techniques show competitive performance in adapting to new LLMs.
However, there is still a performance gap between the current methods and the ideal values, which train the detectors on all data at once.
Future works are encouraged to explore and develop more techniques to build continuously evolved MGT detectors.

\section{Limitation}

\mypara{Datasets and Methods}
Despite the collected data being from multiple resources, some specific subject such as Education or Chemistry only covers Wiki data.
Additionally, the MGTs are generated using one prompt and lack complicated processes, e.g., machine-written machine-humanized~\cite{abassy2024llmdetectaivetoolfinegrainedmachinegenerated}, or human-edited~\cite{artemova2025beemobenchmarkexperteditedmachinegenerated}.
Since our study focuses on the performance of detectors in attribution tasks, we leave these limitations for the future continuous development of our open-source code framework and dataset.

\mypara{Metric-based Detectors for Text Attribution}
Our results show that the metric-based detectors for text attribution only show performance slightly above random guesses but do not provide a solution to poor performance.
Although we find that simply concatenating different metrics may improve performance, there is still a huge gap between the model-based detectors.
This is an important yet underexplored topic in developing robust and comprehensive MGT detectors.
We leave it as our future work for further exploration.

\mypara{Adapting Detectors to New LLMs}
We are the first to introduce this setting in the MGT detection task and evaluate the result on detectors equipped with several adapting techniques.
Due to the limitation of research resources, we only collect data from 5 different LLMs, as a result, the number of classes in the pre-training stage and adapting stage is small.
We will continuously expand the data resources and add more advancing LLMs.
Additionally, our evaluation shows that there is still a large performance gap between the current method and the performance upper bound (attribution task with full training data for all classes).
We leave it as our future work for further exploration.
\section{Conclusion}
In this work, we introduce \Dataset, a large-scale academic writing dataset containing over 336M tokens and 749K samples from STEM, Humanities, and Social Sciences, including HWTs and MGTs, generated by five different LLMs.
We also build up an extensible code framework to benchmark MGT detectors efficiently in various tasks.
Based on \Dataset and our code framework, we investigate the performance and generalization ability of MGT detectors and have several observations.
First, Our findings show that model-based detectors consistently outperform metric-based ones in binary classification, while metric-based detectors struggle in attribution tasks due to the low dimensionality of their features. 
This underscores the need for further development of metric-based approaches suitable for attribution tasks.
Second, our evaluation of detector transferability reveals that transferring across different LLMs in binary classification and across domains in attribution tasks leads to significant performance degradation. 
Third, we introduce a novel attribution task setting where detectors must adapt to new LLMs with minimal access to prior data. 
While techniques for both few-shot and many-shot settings show competitive performance, a gap remains between current methods and the ideal case of training detectors on all available data. 
Future research should focus on developing more effective techniques for building continuously evolving MGT detectors capable of handling the challenges posed by new LLMs and domains.



\bibliographystyle{plain}
\bibliography{normal}

\begin{thebibliography}{10}

\bibitem{abassy2024llmdetectaivetoolfinegrainedmachinegenerated}
Mervat Abassy, Kareem Elozeiri, Alexander Aziz, Minh~Ngoc Ta, Raj~Vardhan Tomar, Bimarsha Adhikari, Saad El~Dine Ahmed, Yuxia Wang, Osama~Mohammed Afzal, Zhuohan Xie, Jonibek Mansurov, Ekaterina Artemova, Vladislav Mikhailov, Rui Xing, Jiahui Geng, Hasan Iqbal, Zain~Muhammad Mujahid, Tarek Mahmoud, Akim Tsvigun, Alham~Fikri Aji, Artem Shelmanov, Nizar Habash, Iryna Gurevych, and Preslav Nakov.
\newblock Llm-detectaive: a tool for fine-grained machine-generated text detection, 2024.

\bibitem{artemova2025beemobenchmarkexperteditedmachinegenerated}
Ekaterina Artemova, Jason Lucas, Saranya Venkatraman, Jooyoung Lee, Sergei Tilga, Adaku Uchendu, and Vladislav Mikhailov.
\newblock Beemo: Benchmark of expert-edited machine-generated outputs, 2025.

\bibitem{BAA08}
Sameer Badaskar, Sachin Agarwal, and Shilpa Arora.
\newblock {Identifying Real or Fake Articles: Towards better Language Modeling}.
\newblock In {\em {International Joint Conference on Natural Language Processing (IJCNLP)}}, pages 817--822. ACL, 2008.

\bibitem{BZTYZ24}
Guangsheng Bao, Yanbin Zhao, Zhiyang Teng, Linyi Yang, and Yue Zhang.
\newblock Fast-detectgpt: Efficient zero-shot detection of machine-generated text via conditional probability curvature.
\newblock In {\em The Twelfth International Conference on Learning Representations, {ICLR} 2024, Vienna, Austria, May 7-11, 2024}, 2024.

\bibitem{chen2019closer}
Wei-Yu Chen, Yen-Cheng Liu, Zsolt Kira, Yu-Chiang~Frank Wang, and Jia-Bin Huang.
\newblock A closer look at few-shot classification.
\newblock {\em arXiv preprint arXiv:1904.04232}, 2019.

\bibitem{gan2024navigatingriskssurveysecurity}
Yuyou Gan, Yong Yang, Zhe Ma, Ping He, Rui Zeng, Yiming Wang, Qingming Li, Chunyi Zhou, Songze Li, Ting Wang, Yunjun Gao, Yingcai Wu, and Shouling Ji.
\newblock Navigating the risks: A survey of security, privacy, and ethics threats in llm-based agents, 2024.

\bibitem{GSR19}
Sebastian Gehrmann, Hendrik Strobelt, and Alexander~M. Rush.
\newblock {{GLTR:} Statistical Detection and Visualization of Generated Text}.
\newblock In {\em {Annual Meeting of the Association for Computational Linguistics (ACL)}}, pages 111--116. ACL, 2019.

\bibitem{geng2024impact}
Mingmeng Geng, Caixi Chen, Yanru Wu, Dongping Chen, Yao Wan, and Pan Zhou.
\newblock The impact of large language models in academia: from writing to speaking.
\newblock {\em arXiv preprint arXiv:2409.13686}, 2024.

\bibitem{GZWJNDYW23}
Biyang Guo, Xin Zhang, Ziyuan Wang, Minqi Jiang, Jinran Nie, Yuxuan Ding, Jianwei Yue, and Yupeng Wu.
\newblock {How Close is ChatGPT to Human Experts? Comparison Corpus, Evaluation, and Detection}.
\newblock {\em {CoRR abs/2301.07597}}, 2023.

\bibitem{binoculars}
Abhimanyu Hans, Avi Schwarzschild, Valeriia Cherepanova, Hamid Kazemi, Aniruddha Saha, Micah Goldblum, Jonas Geiping, and Tom Goldstein.
\newblock Spotting {LLM}s with binoculars: Zero-shot detection of machine-generated text.
\newblock In {\em Proceedings of the 41st International Conference on Machine Learning}. PMLR, 2024.

\bibitem{HSCBZ23}
Xinlei He, Xinyue Shen, Zeyuan Chen, Michael Backes, and Yang Zhang.
\newblock {MGTBench: Benchmarking Machine-Generated Text Detection}.
\newblock {\em {CoRR abs/2303.14822}}, 2023.

\bibitem{hu2023radar}
Xiaomeng Hu, Pin-Yu Chen, and Tsung-Yi Ho.
\newblock {RADAR}: Robust {AI}-text detection via adversarial learning.
\newblock In {\em Thirty-seventh Conference on Neural Information Processing Systems}, 2023.

\bibitem{IDCE20}
Daphne Ippolito, Daniel Duckworth, Chris Callison{-}Burch, and Douglas Eck.
\newblock {Automatic Detection of Generated Text is Easiest when Humans are Fooled}.
\newblock In {\em {Annual Meeting of the Association for Computational Linguistics (ACL)}}, pages 1808--1822. ACL, 2020.

\bibitem{li2017learning}
Zhizhong Li and Derek Hoiem.
\newblock Learning without forgetting.
\newblock {\em IEEE transactions on pattern analysis and machine intelligence}, 40(12):2935--2947, 2017.

\bibitem{roberta}
Yinhan Liu, Myle Ott, Naman Goyal, Jingfei Du, Mandar Joshi, Danqi Chen, Omer Levy, Mike Lewis, Luke Zettlemoyer, and Veselin Stoyanov.
\newblock Roberta: {A} robustly optimized {BERT} pretraining approach.
\newblock {\em CoRR}, abs/1907.11692, 2019.

\bibitem{masana2022class}
Marc Masana, Xialei Liu, Bart{\l}omiej Twardowski, Mikel Menta, Andrew~D Bagdanov, and Joost Van De~Weijer.
\newblock Class-incremental learning: survey and performance evaluation on image classification.
\newblock {\em IEEE Transactions on Pattern Analysis and Machine Intelligence}, 45(5):5513--5533, 2022.

\bibitem{minaee2024largelanguagemodelssurvey}
Shervin Minaee, Tomas Mikolov, Narjes Nikzad, Meysam Chenaghlu, Richard Socher, Xavier Amatriain, and Jianfeng Gao.
\newblock Large language models: A survey, 2024.

\bibitem{Mistral}
MistralAI.
\newblock \url{https://mistral.ai/}.

\bibitem{MLKMF23}
Eric Mitchell, Yoonho Lee, Alexander Khazatsky, Christopher~D. Manning, and Chelsea Finn.
\newblock {DetectGPT: Zero-Shot Machine-Generated Text Detection using Probability Curvature}.
\newblock {\em {CoRR abs/2301.11305}}, 2023.

\bibitem{Moonshot}
Moonshot.
\newblock \url{https://www.moonshot.cn}.

\bibitem{chatgpt}
OpenAI.
\newblock \url{https://chat.openai.com/chat}.

\bibitem{O23}
OpenAI.
\newblock {{GPT-4} Technical Report}.
\newblock {\em {CoRR abs/2303.08774}}, 2023.

\bibitem{OWJAWMZASRSHKMSAWCLL22}
Long Ouyang, Jeffrey Wu, Xu~Jiang, Diogo Almeida, Carroll~L. Wainwright, Pamela Mishkin, Chong Zhang, Sandhini Agarwal, Katarina Slama, Alex Ray, John Schulman, Jacob Hilton, Fraser Kelton, Luke Miller, Maddie Simens, Amanda Askell, Peter Welinder, Paul~F. Christiano, Jan Leike, and Ryan Lowe.
\newblock {Training language models to follow instructions with human feedback}.
\newblock In {\em {Annual Conference on Neural Information Processing Systems (NeurIPS)}}. NeurIPS, 2022.

\bibitem{paul-etal-2022-class}
Debjit Paul, Daniil Sorokin, and Judith Gaspers.
\newblock Class incremental learning for intent classification with limited or no old data.
\newblock In Francesco Barbieri, Jose Camacho-Collados, Bhuwan Dhingra, Luis Espinosa-Anke, Elena Gribovskaya, Angeliki Lazaridou, Daniel Loureiro, and Leonardo Neves, editors, {\em Proceedings of the First Workshop on Ever Evolving NLP (EvoNLP)}, pages 16--25, Abu Dhabi, United Arab Emirates (Hybrid), December 2022. Association for Computational Linguistics.

\bibitem{RWCLAS19}
Alec Radford, Jeffrey Wu, Rewon Child, David Luan, Dario Amodei, and Ilya Sutskever.
\newblock {Language Models are Unsupervised Multitask Learners}.
\newblock {\em {OpenAI blog}}, 2019.

\bibitem{rebuffi2017icarl}
Sylvestre-Alvise Rebuffi, Alexander Kolesnikov, Georg Sperl, and Christoph~H Lampert.
\newblock icarl: Incremental classifier and representation learning.
\newblock In {\em Proceedings of the IEEE conference on Computer Vision and Pattern Recognition}, pages 2001--2010, 2017.

\bibitem{reimers2019sentence}
N~Reimers.
\newblock Sentence-bert: Sentence embeddings using siamese bert-networks.
\newblock {\em arXiv preprint arXiv:1908.10084}, 2019.

\bibitem{riemer2018learning}
Matthew Riemer, Ignacio Cases, Robert Ajemian, Miao Liu, Irina Rish, Yuhai Tu, and Gerald Tesauro.
\newblock Learning to learn without forgetting by maximizing transfer and minimizing interference.
\newblock {\em arXiv preprint arXiv:1810.11910}, 2018.

\bibitem{distilbert}
Victor Sanh, Lysandre Debut, Julien Chaumond, and Thomas Wolf.
\newblock Distilbert, a distilled version of {BERT:} smaller, faster, cheaper and lighter.
\newblock {\em CoRR}, abs/1910.01108, 2019.

\bibitem{snell2017prototypicalnetworksfewshotlearning}
Jake Snell, Kevin Swersky, and Richard~S. Zemel.
\newblock Prototypical networks for few-shot learning, 2017.

\bibitem{JYDP23}
Jinyan Su, Terry~Yue Zhuo, Di~Wang, and Preslav Nakov.
\newblock {Detectllm: Leveraging log rank information for zero-shot detection of machine-generated text}.
\newblock {\em {CoRR abs/2306.05540}}, 2023.

\bibitem{sung2018learning}
Flood Sung, Yongxin Yang, Li~Zhang, Tao Xiang, Philip~HS Torr, and Timothy~M Hospedales.
\newblock Learning to compare: Relation network for few-shot learning.
\newblock In {\em Proceedings of the IEEE conference on computer vision and pattern recognition}, pages 1199--1208, 2018.

\bibitem{tao2024reliabledetectionllmgeneratedtexts}
Zhen Tao, Yanfang Chen, Dinghao Xi, Zhiyu Li, and Wei Xu.
\newblock Towards reliable detection of llm-generated texts: A comprehensive evaluation framework with cudrt, 2024.

\bibitem{TMSAABBBBBBBCCCEFFFFGGGHHHIKKKKKKLLLLLMMMMMNPRRSSSSSTTTWKXYZZFKNRSES23}
Hugo Touvron, Louis Martin, Kevin Stone, Peter Albert, Amjad Almahairi, Yasmine Babaei, Nikolay Bashlykov, Soumya Batra, Prajjwal Bhargava, Shruti Bhosale, Dan Bikel, Lukas Blecher, Cristian Canton{-}Ferrer, Moya Chen, Guillem Cucurull, David Esiobu, Jude Fernandes, Jeremy Fu, Wenyin Fu, Brian Fuller, Cynthia Gao, Vedanuj Goswami, Naman Goyal, Anthony Hartshorn, Saghar Hosseini, Rui Hou, Hakan Inan, Marcin Kardas, Viktor Kerkez, Madian Khabsa, Isabel Kloumann, Artem Korenev, Punit~Singh Koura, Marie{-}Anne Lachaux, Thibaut Lavril, Jenya Lee, Diana Liskovich, Yinghai Lu, Yuning Mao, Xavier Martinet, Todor Mihaylov, Pushkar Mishra, Igor Molybog, Yixin Nie, Andrew Poulton, Jeremy Reizenstein, Rashi Rungta, Kalyan Saladi, Alan Schelten, Ruan Silva, Eric~Michael Smith, Ranjan Subramanian, Xiaoqing~Ellen Tan, Binh Tang, Ross Taylor, Adina Williams, Jian~Xiang Kuan, Puxin Xu, Zheng Yan, Iliyan Zarov, Yuchen Zhang, Angela Fan, Melanie Kambadur, Sharan Narang, Aur{\'{e}}lien Rodriguez, Robert Stojnic, Sergey Edunov,
  and Thomas Scialom.
\newblock {Llama 2: Open Foundation and Fine-Tuned Chat Models}.
\newblock {\em {CoRR abs/2307.09288}}, 2023.

\bibitem{van2008visualizing}
Laurens Van~der Maaten and Geoffrey Hinton.
\newblock Visualizing data using t-sne.
\newblock {\em Journal of machine learning research}, 9(11), 2008.

\bibitem{YJPJAAMTGTo24}
Yuxia Wang, Jonibek Mansurov, Petar Ivanov, Jinyan Su, Artem Shelmanov, Akim Tsvigun, Osama~Mohanned Afzal, Tarek Mahmoud, Giovanni Puccetti, Thomas Arnold, et~al.
\newblock {M4GT-Bench: Evaluation Benchmark for Black-Box Machine-Generated Text Detection}.
\newblock In {\em {Annual Meeting of the Association for Computational Linguistics (ACL)}}, 2024.

\bibitem{WDSCDMCRLFB19}
Thomas Wolf, Lysandre Debut, Victor Sanh, Julien Chaumond, Clement Delangue, Anthony Moi, Pierric Cistac, Tim Rault, R{\'{e}}mi Louf, Morgan Funtowicz, and Jamie Brew.
\newblock {HuggingFace's Transformers: State-of-the-art Natural Language Processing}.
\newblock {\em {CoRR abs/1910.03771}}, 2019.

\bibitem{DBLP:journals/corr/abs-2410-23746}
Junchao Wu, Runzhe Zhan, Derek~F. Wong, Shu Yang, Xinyi Yang, Yulin Yuan, and Lidia~S. Chao.
\newblock Detectrl: Benchmarking llm-generated text detection in real-world scenarios.
\newblock {\em CoRR}, abs/2410.23746, 2024.

\bibitem{wu2019large}
Yue Wu, Yinpeng Chen, Lijuan Wang, Yuancheng Ye, Zicheng Liu, Yandong Guo, and Yun Fu.
\newblock Large scale incremental learning.
\newblock In {\em Proceedings of the IEEE/CVF conference on computer vision and pattern recognition}, pages 374--382, 2019.

\bibitem{DBLP:conf/naacl/XiaYFY21}
Congying Xia, Wenpeng Yin, Yihao Feng, and Philip~S. Yu.
\newblock Incremental few-shot text classification with multi-round new classes: Formulation, dataset and system.
\newblock In Kristina Toutanova, Anna Rumshisky, Luke Zettlemoyer, Dilek Hakkani{-}T{\"{u}}r, Iz~Beltagy, Steven Bethard, Ryan Cotterell, Tanmoy Chakraborty, and Yichao Zhou, editors, {\em Proceedings of the 2021 Conference of the North American Chapter of the Association for Computational Linguistics: Human Language Technologies, {NAACL-HLT} 2021, Online, June 6-11, 2021}, pages 1351--1360. Association for Computational Linguistics, 2021.

\bibitem{yang2021free}
Shuo Yang, Lu~Liu, and Min Xu.
\newblock Free lunch for few-shot learning: Distribution calibration.
\newblock {\em arXiv preprint arXiv:2101.06395}, 2021.

\bibitem{yi2024jailbreakattacksdefenseslarge}
Sibo Yi, Yule Liu, Zhen Sun, Tianshuo Cong, Xinlei He, Jiaxing Song, Ke~Xu, and Qi~Li.
\newblock Jailbreak attacks and defenses against large language models: A survey, 2024.

\bibitem{zhao2024surveylargelanguagemodels}
Wayne~Xin Zhao, Kun Zhou, Junyi Li, Tianyi Tang, Xiaolei Wang, Yupeng Hou, Yingqian Min, Beichen Zhang, Junjie Zhang, Zican Dong, Yifan Du, Chen Yang, Yushuo Chen, Zhipeng Chen, Jinhao Jiang, Ruiyang Ren, Yifan Li, Xinyu Tang, Zikang Liu, Peiyu Liu, Jian-Yun Nie, and Ji-Rong Wen.
\newblock A survey of large language models, 2024.

\bibitem{zhou2023revisiting}
Da-Wei Zhou, Zi-Wen Cai, Han-Jia Ye, De-Chuan Zhan, and Ziwei Liu.
\newblock Revisiting class-incremental learning with pre-trained models: Generalizability and adaptivity are all you need.
\newblock {\em arXiv preprint arXiv:2303.07338}, 2023.

\end{thebibliography}
\newpage
\renewcommand{\thefigure}{A\arabic{figure}}
\renewcommand{\thetable}{A\arabic{table}}
\setcounter{figure}{0}
\setcounter{table}{0}
\appendix

\section{Generation Prompts}\label{sec-app:prompt}

\begin{tcolorbox}[colback=orange!10,
                  colframe=orange!70,
                  width=\columnwidth,
                  fonttitle=\bfseries\centering,
                  coltitle=black, 
                  arc=3mm, auto outer arc,
                  before=\vspace{3pt},  
                  after=\vspace{3pt},
                  boxsep=1pt,
                  left=2pt,
                  right=2pt,
                  title=Prompt Template for Arxiv Text
                 ]
\small

\begin{tabularx}{\linewidth}{X}
    <\underline{Background}>: \\
    Please act as an expert paper editor and revise a section of the paper to make it more fluent and elegant. 
Please only include the revised section in your answer. Here are the specific
requirements: 

1. Enable readers to grasp the main points or
essence of the paper quickly. 

2. Allow readers to understand
the important information, analysis, and arguments throughout
the entire paper. 

3. Help readers remember the key points of
the paper. 

4. Please clearly state the innovative aspects of your
research in the section, emphasizing your contributions. 

5. Use
concise and clear language to describe your findings and results,
making it easier for reviewers to understand the paper. Here is
the original section of the paper: \\
    <\underline{text}>: //to-be-polished text \\

\end{tabularx}
\end{tcolorbox}

\begin{tcolorbox}[colback=orange!10,
                  colframe=orange!70,
                  width=\columnwidth,
                  fonttitle=\bfseries\centering, 
                  coltitle=black, 
                  arc=3mm, auto outer arc,
                  before=\vspace{3pt},  
                  after=\vspace{3pt},
                  boxsep=1pt,
                  left=2pt,
                  right=2pt,
                  title=Prompt Template for Arxiv Text
                 ]
\small
\begin{tabularx}{\linewidth}{X}
    
    <\underline{Background}>: \\
    Please act as an expert paper editor and revise a section of the paper to make it more fluent and elegant. 
Please only include the revised section in your answer. Here are the specific
requirements: 

1. Enable readers to grasp the main points or
essence of the paper quickly. 

2. Allow readers to understand
the important information, analysis, and arguments throughout
the entire paper. 

3. Help readers remember the key points of
the paper. 

4. Please clearly state the innovative aspects of your
research in the section, emphasizing your contributions. 

5. Use
concise and clear language to describe your findings and results,
making it easier for reviewers to understand the paper. Here is
the original section of the paper: \\
    <\underline{text}>: //to-be-polished text \\

\end{tabularx}
\end{tcolorbox}

\begin{tcolorbox}[colback=orange!10,
                  colframe=orange!70,
                  width=\columnwidth,
                  fonttitle=\bfseries\centering, 
                  coltitle=black, 
                  arc=3mm, auto outer arc,
                  before=\vspace{3pt},  
                  after=\vspace{3pt},
                  boxsep=1pt,
                  left=2pt,
                  right=2pt,
                  title=Prompt Template for Gutendex Text
                 ]
\small
\begin{tabularx}{\linewidth}{X}
    <\underline{Background}>: \\
    Please act as an expert book editor and revise the
book content from the perspective of a book editor
to make it fluent and elegant.

1. Clarity: Ensure that your writing is clear and easy to understand. 
Avoid jargon and complex language that may confuse the reader. 

2. Relevance: Make sure that the content you are writing is relevant
 to the topic at hand. Do not deviate from the main subject.
 
3. Accuracy: Ensure that all the information you provide is 
accurate and up-to-date. This includes statistics, facts, and theories.

4. Brevity: Keep your writing concise. Avoid unnecessary words or 
phrases that do not add value to the content.
Here is the original book content: \\
    <\underline{text}>: //to-be-polished text \\

\end{tabularx}
\end{tcolorbox}

\begin{table*}[]
\centering
\caption{\textbf{Sources of Data for Machine Generated Text:}
This table lists the primary paper sources (such as Arxiv and Project Gutenberg) and supplementary resources (Wiki) available for different STEM, Social Science, and Humanity subfields, along with notes on the availability of paper sources where applicable.}
\label{tab:data-mgt}
\setlength{\tabcolsep}{3pt}
\renewcommand{\arraystretch}{0.9}
\resizebox{\textwidth}{!}{\begin{tabular}{ccccccccc} \toprule
Domain                                               & Subfield                        & Source                                                  & Human          & GPT-4omini    & GPT-3.5       & Mixtral-8$\times$7b   & Llama-3.1-70b  & Moonshot-8k   \\ \midrule
\multirow{8}{*}{STEM}                                & Physics                         & \multirow{6}{*}{Arxiv \& Wiki}                         & 10.8K / 11,926.6K & 9.1K / 5,485.9K  & 8.4K / 2,106.2K & 8.7K / 3,291.4K  & 3.5K / 1,059.1K  & 2.5K / 891.9K   \\
                                                     & Math                            &                                                        & 14.1K / 12,338.5K & 12.0K / 5,717.0K & 13.6K / 3,444.1K & 10.3K / 3,425.9K  & 6.4K / 1,836.8K  & 2.4K / 808.8K   \\
                                                     & CS               &                                                        & 14.7K / 12,275.5K & 11.9K / 4,989.4K & 14.2K / 3,415.4K & 9.4K / 3,277.9K   & 3.5K / 934.2K    & 2.9K / 983.6K   \\
                                                     & Biology                         &                                                        & 15.7K / 11,485.3K & 13.6K / 5,704.5K & 14.9K / 3,508.6K & 10.9K / 3,716.7K  & 3.5K / 924.9K    & 3.5K / 1,159.3K \\
                                                     & EE
                                                     &                                                        & 19.7K / 13,346.2K & 16.9K / 6,129.3K & 18.6K / 4,378.4K & 13.0K / 4,384.9K  & 4.2K / 1,130.5K  & 4.1K / 1,350.1K \\
                                                     & Statistics                      &                                                        & 9.7K / 11,491.9K  & 7.6K / 4,339.8K  & 9.5K / 2,545.2K  & 6.8K / 2,610.7K   & 2.9K / 832.4K    & 2.0K / 694.1K   \\ \cmidrule{2-9} 
                                                     & Chemistry                       & \multirow{2}{*}{Wiki}                                  & 2.4K / 415.4K     & 2.2K / 425.2K    & 2.8K / 536.8K    & 1.6K / 438.6K     & 1.0K / 148.6K    & 0.5K / 141.2K   \\
                                                     & Medicine                        &                                                        & 8.7K / 1,668.3K   & 8.1K / 1,662.6K  & 7.8K / 1,470.6K  & 5.1K / 1,419.3K   & 2.0K / 454.8K    & 1.8K / 528.9K   \\ \midrule
\multirow{3}{*}{Social Science}                      & Education                       & Gutenberg \& Wiki                                       & 14.2K / 3,831.0K  & 13.0K / 2,833.9K & 12.5K / 2,377.2K & 9.2K / 2,704.2K   & 3.1K / 925.6K    & 2.8K / 905.9K   \\
                                                     & Economy                         & Arxiv \& Wiki                                           & 12.6K / 6,807.5K  & 11.3K / 3,663.5K & 11.7K / 2,531.4K & 6.1K / 2,025.1K   & 2.4K / 655.6K    & 1.9K / 621.5K   \\
                                                     & Management                      & Wiki                                                   & 3.5K / 648.5K     & 3.3K / 618.0K    & 3.3K / 739.4K    & 2.1K / 589.0K     & 0.6K / 132.5K    & 0.7K / 190.9K   \\ \midrule
\multirow{5}{*}{Humanities}                          & Literature                      & \multirow{5}{*}{Gutenberg \& Wiki}                     & 18.7K / 13,276.7K & 13.6K / 5,306.9K & 11.4K / 2,077.9K & 13.5K / 5,306.9K  & 9.3K / 4,439.6K  & 3.9K / 1,823.6K \\
                                                     & Law                             &                                                        & 7.5K / 2,695.6K    & 6.5K / 1,639.1K  & 6.1K / 1,106.6K  & 5.2K / 1,727.3K   & 2.3K / 780.2K    & 1.5K / 522.0K   \\
                                                     & Art                             &                                                        & 8.4K / 5,899.5K    & 6.4K / 2,576.2K  & 5.7K / 1,110.1K  & 6.0K / 2,411.3K   & 4.2K / 2,040.1K  & 1.8K / 816.4K   \\
                                                     & History                         &                                                        & 30.0K / 33,517.6K  & 18.4K / 12,848.3K& 14.5K / 3,505.5K & 23.6K / 11,572.2K & 18.4K / 11,019.4K& 6.1K / 3,644.4K \\
                                                     & Philosophy                      &                                                        & 3.5K / 1,998.4K    & 2.7K / 776.5K    & 2.3K / 407.8K    & 2.6K / 937.0K     & 1.5K / 598.6K    & 0.7K / 280.6K   \\ \bottomrule
\end{tabular}}
\end{table*}

\section{Data Moderation Policy}\label{sec-app:moderate}
In this section, we will introduce the data moderation policy for both \textit{Human} and \textit{Machine} splits to remove the data of poor quality and obvious identifiers.

\mypara{Human Split}  
To ensure data quality, we apply a rigorous cleaning process to remove noise and irrelevant content.
We discard texts with fewer than 50 tokens and ensure that all entries start and end with complete sentences, preserving their coherence and clarity.
For wiki data, entries containing terms like ``ISBN,'' ``PMID,'' ``doi,'' ``vol.,'' ``p.,'' ``References,'' and ``External links'' are excluded, as these could act as identifiers (\Cref{tab:keywords}). 
For book data, texts containing Project Gutenberg license information are removed to avoid duplication.
For arXiv data, we filtered out entries with excessive formatting symbols, specifically those with more than 500 instances of ``\$\$'', 150 ``\&'', or 1000 ``\textbackslash'' (\Cref{tab:keywords}), to maintain readability.

\mypara{Machine Split}
To moderate the machine-generated data, we focus on removing the obvious identifiers for text detection.
First, we remove the text of short length below 50 words (splited by space), which is usually produced by failed or incomplete API queries.
Second, since every text is truncated to 2048 tokens, we start back-tracing from the last token until a period appears and then drop all content after the period to ensure the completeness of the text and avoid easy detection.
Third, we customize different filtering rules for the keywords (\Cref{tab:keywords}) featuring machine generation.
For generation identifiers (``The revised content is:''), we find the closest colon and remove all content before the colon.
For special keywords, we drop the entire text if any of the listed keywords appear after removing the generation identifiers.
For format symbols, we drop the entire text if there are more than 50 tabs (\&) or equation (\$). 
The remaining format symbols, such as `**' and `\#\#', are markdown tags, so we drop the entire text if any of them appear.

\section{Full \Dataset Analysis}\label{sec-app:data}
\begin{figure}[h!]
    \centering
    \includegraphics[scale=0.21]{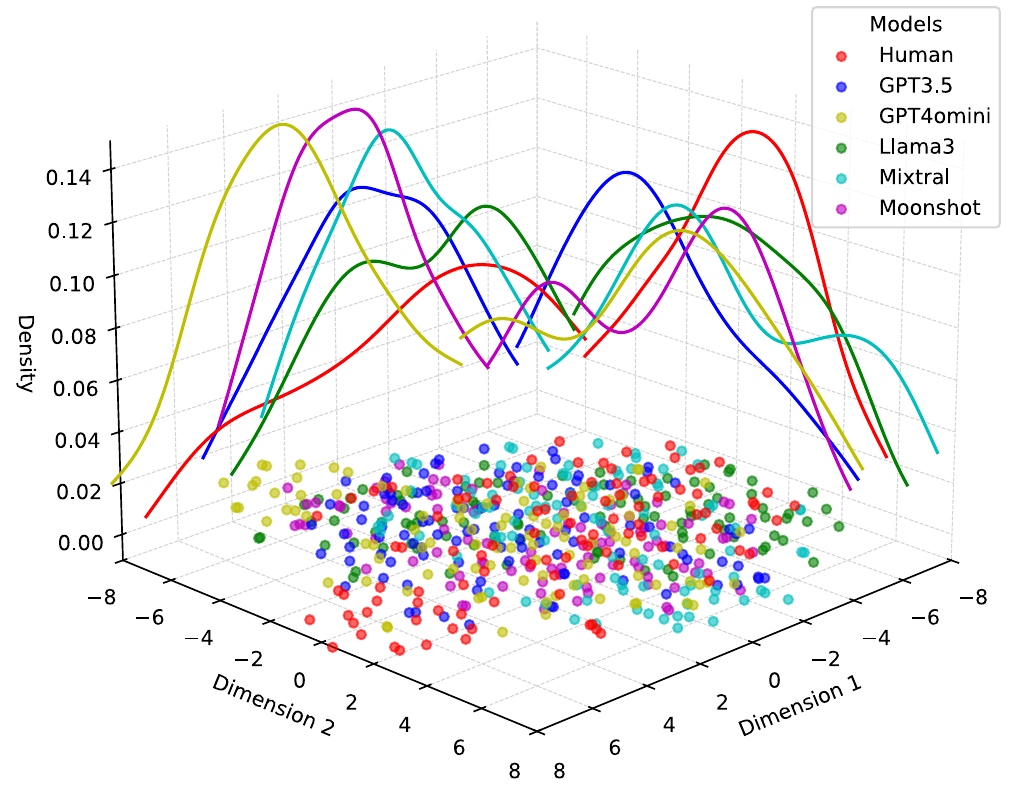}
    \includegraphics[scale=0.18]{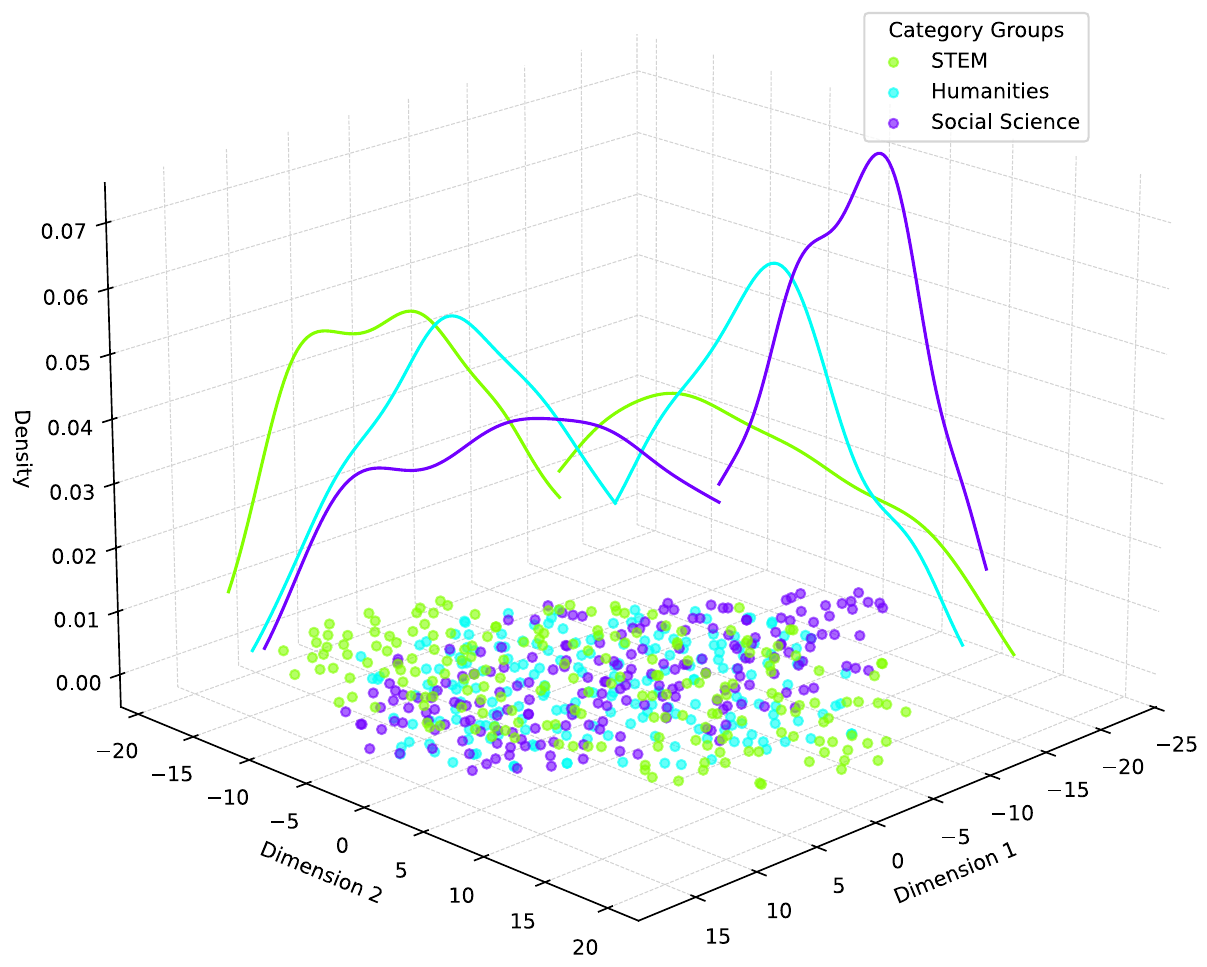}

    \caption{2D Projection of Art Embeddings Across Models}
    \label{fig:Art Embeddings Across Models}
\end{figure}

\mypara{Embedding Distribution}
In this part, we investigate the data distribution of \Dataset from the domain and LLM aspects.
We utilize SentenceBert \cite{reimers2019sentence} to get the embeddings of each data and further adopt TSNE \cite{van2008visualizing} to project the high-dimensional representations into two dimensions.
Regarding the domain aspect, we fix the source LLM and project the data from different domains into \Cref{fig:Art Embeddings Across Models}.
As the projection result shows, we find that the distribution of data from different domains and LLMs varies a lot, which shows the necessity of studying the generalization ability of the MGT detectors. 

\mypara{Data Source and Length}
In this section, we present the length distribution of data across different domains and LLMs in \Cref{tab:data-mgt}. Generally, we collect 749,625 samples with 336,714,335 tokens.
We observe that the token length of HWTs is generally longer than that of MGTs.
This disparity arises because some LLMs are less proficient at generating extended content.
Additionally, within the STEM domain, certain subfields exhibit longer token lengths. 
This can be attributed to our data collection methodology, which focuses primarily on the main sections of research papers. 
These sections often include formulas and technical content that contribute to the increased token count.

\begin{figure}[t]
    \centering
    \includegraphics[width=0.5\textwidth]{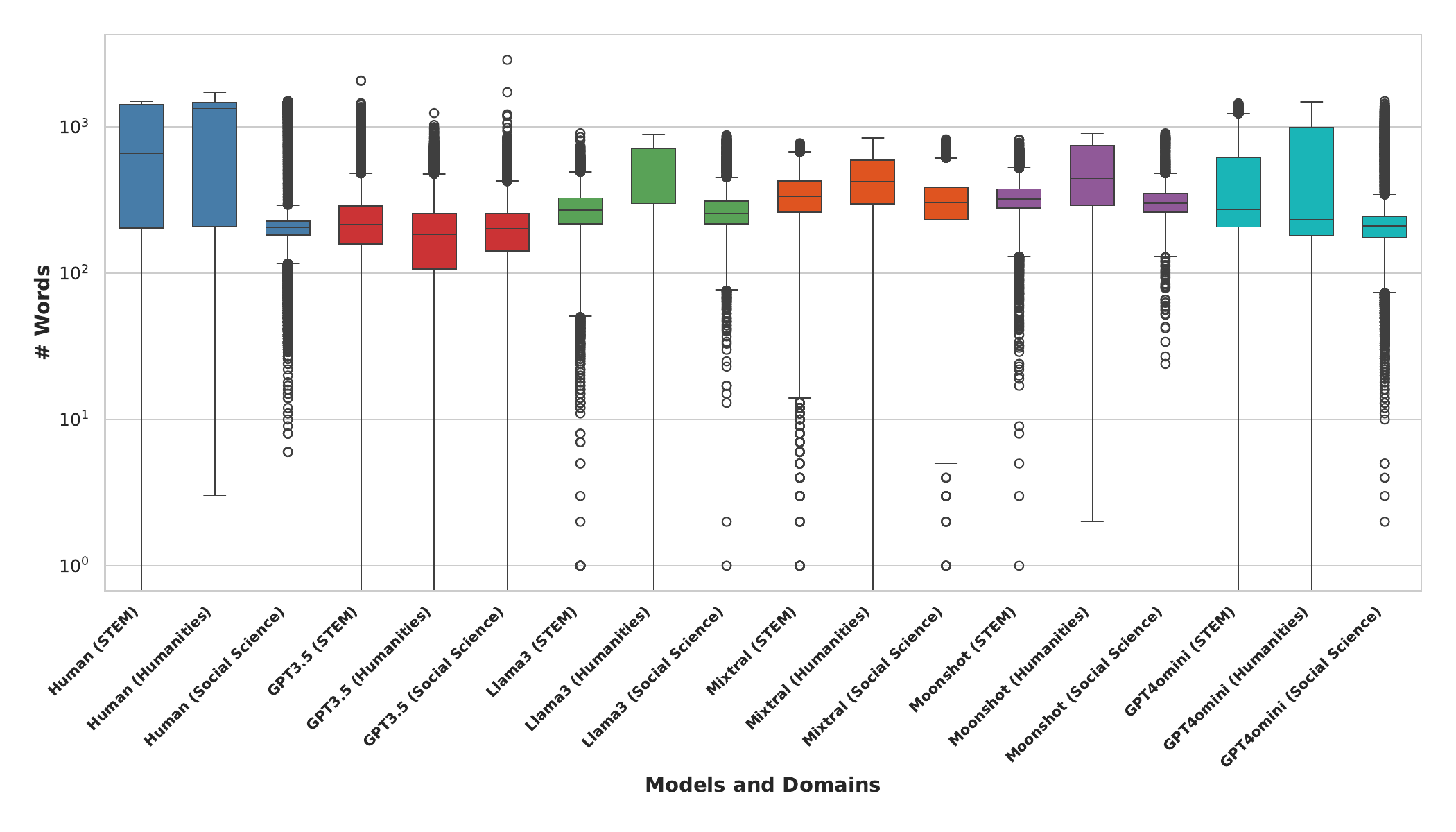}
    \caption{Text Length Distribution Across Models and Domains}
    \label{fig:Text Length Distribution}
\end{figure}

\begin{figure*}[ht]
    \centering
    \begin{subfigure}[b]{0.32\textwidth}
        \includegraphics[width=\textwidth]{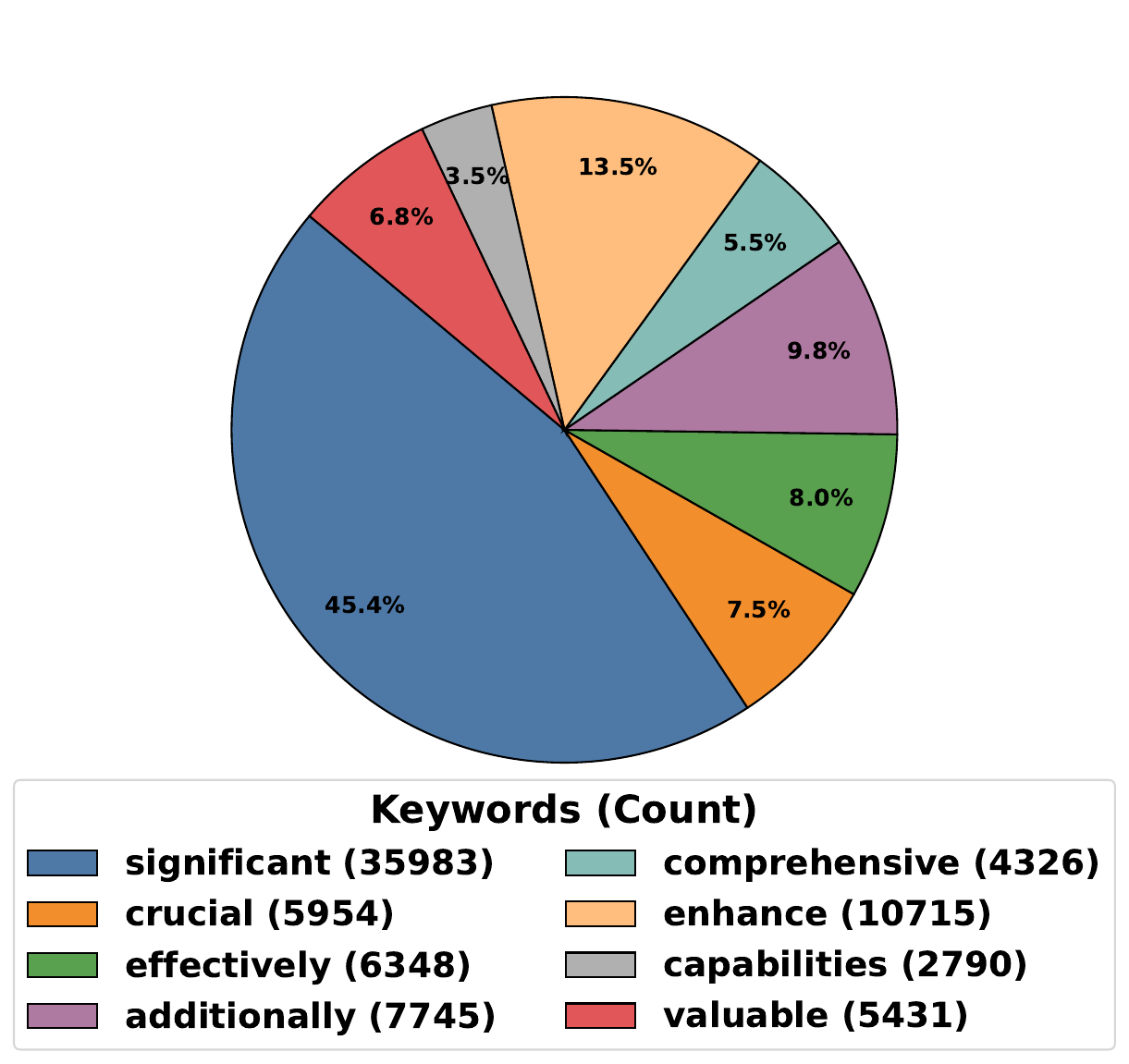}
        \caption{Human}
    \end{subfigure}
    \hfill
    \begin{subfigure}[b]{0.32\textwidth}
        \includegraphics[width=\textwidth]{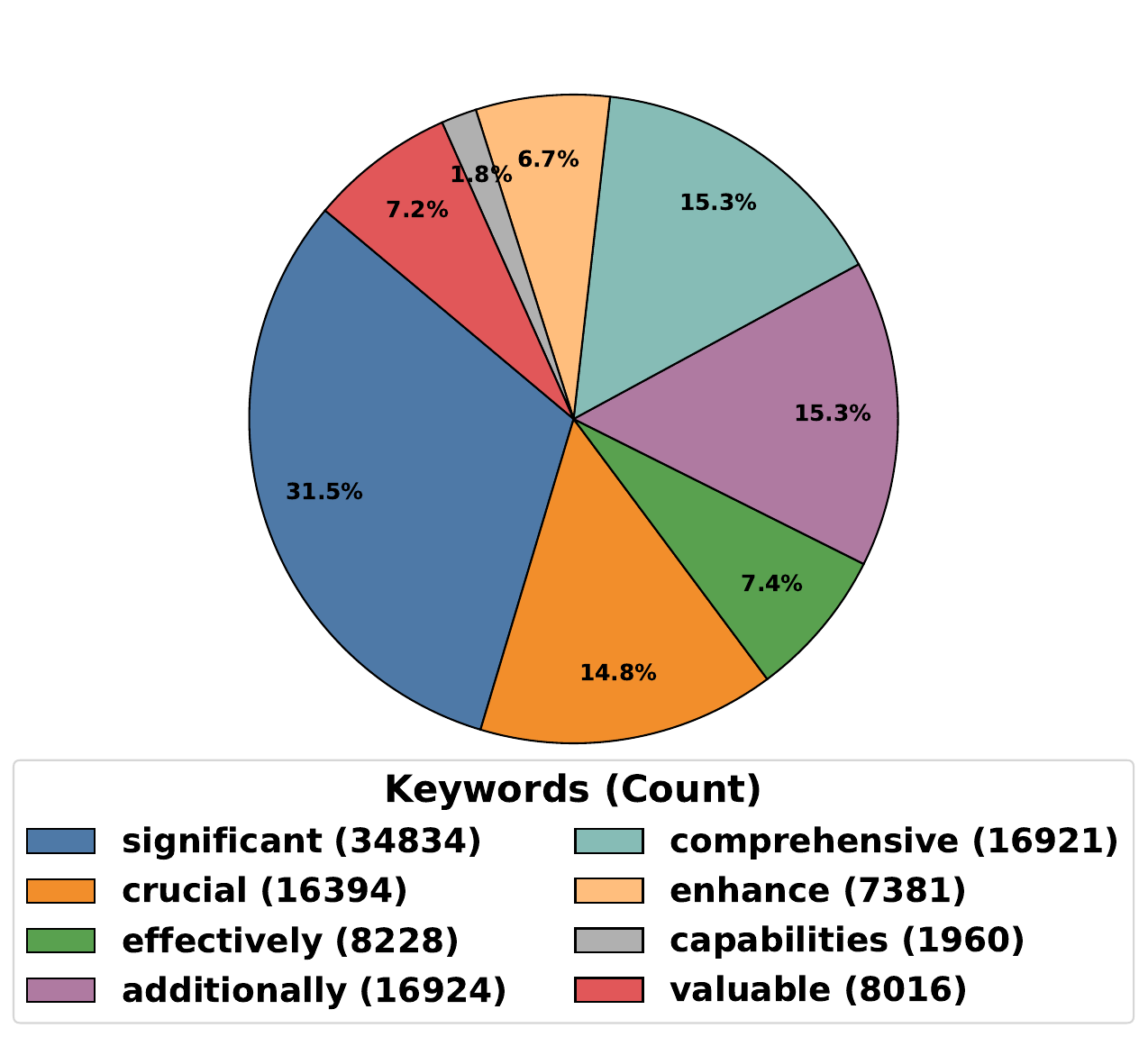}
        \caption{GPT-3.5}
    \end{subfigure}
    \hfill
    \begin{subfigure}[b]{0.32\textwidth}
        \includegraphics[width=\textwidth]{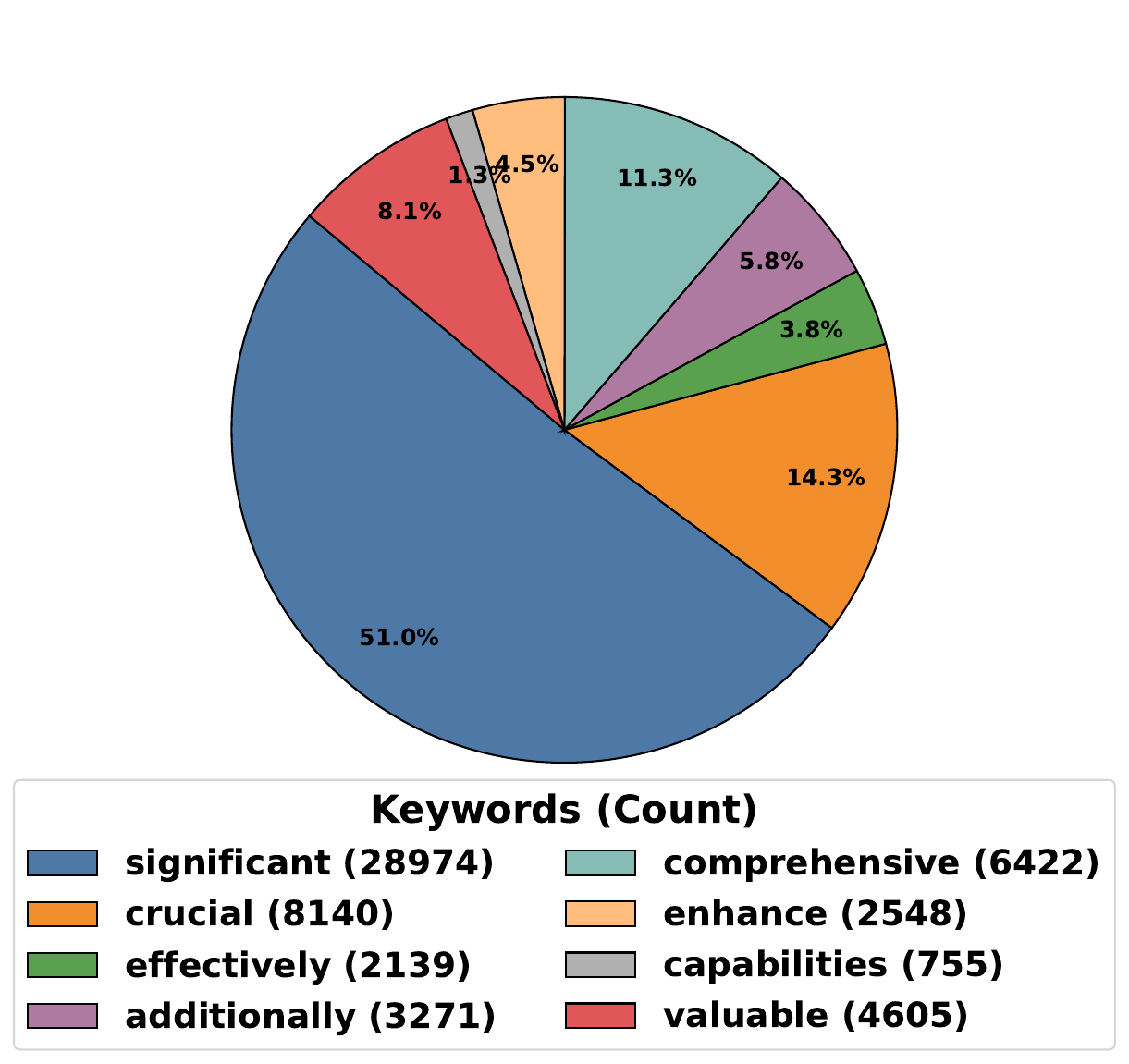}
        \caption{Llama-3}
    \end{subfigure}
    \\
    \begin{subfigure}[b]{0.32\textwidth}
        \includegraphics[width=\textwidth]{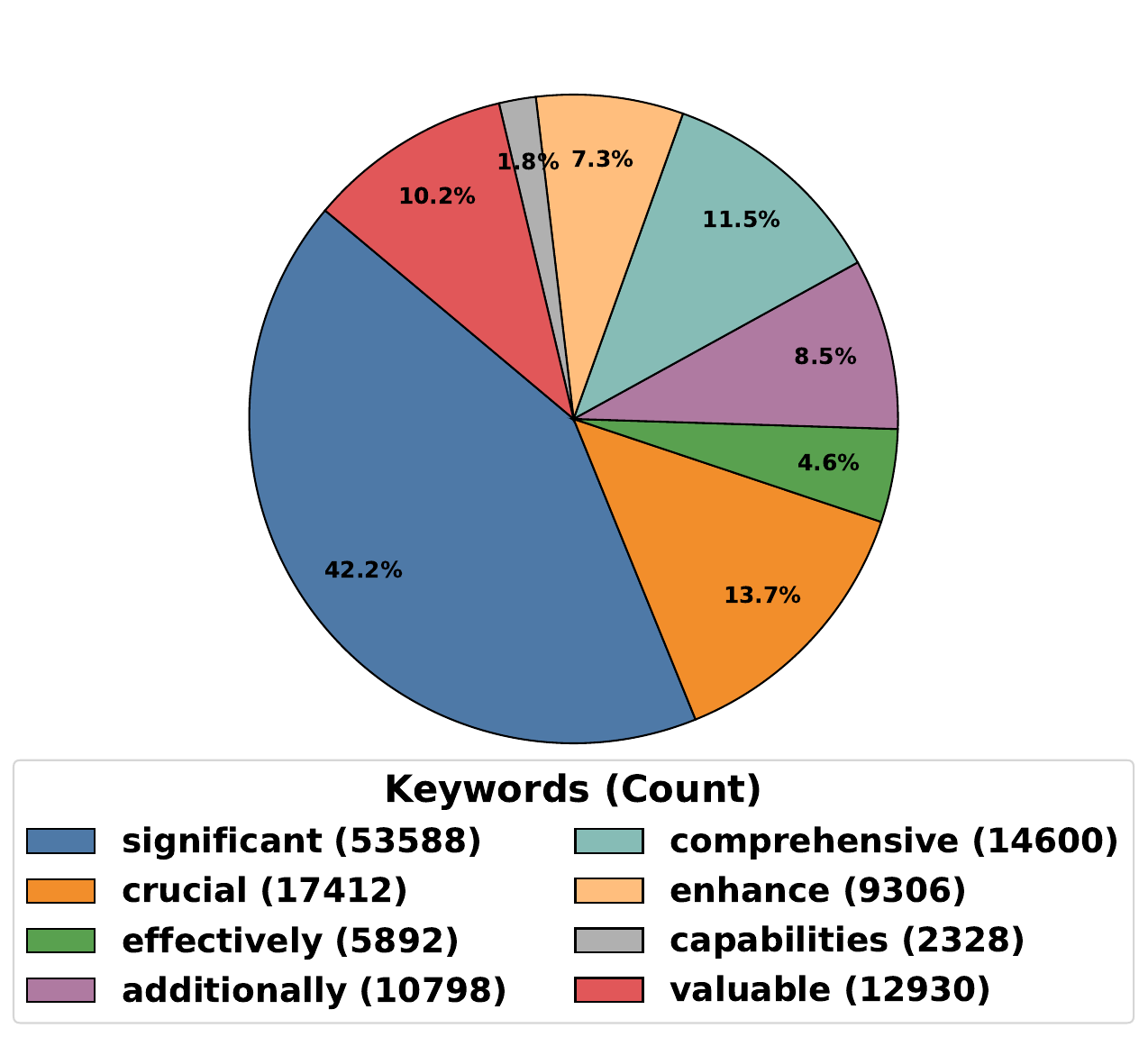}
        \caption{Mixtral}
    \end{subfigure}
    \hfill
    \begin{subfigure}[b]{0.32\textwidth}
        \includegraphics[width=\textwidth]{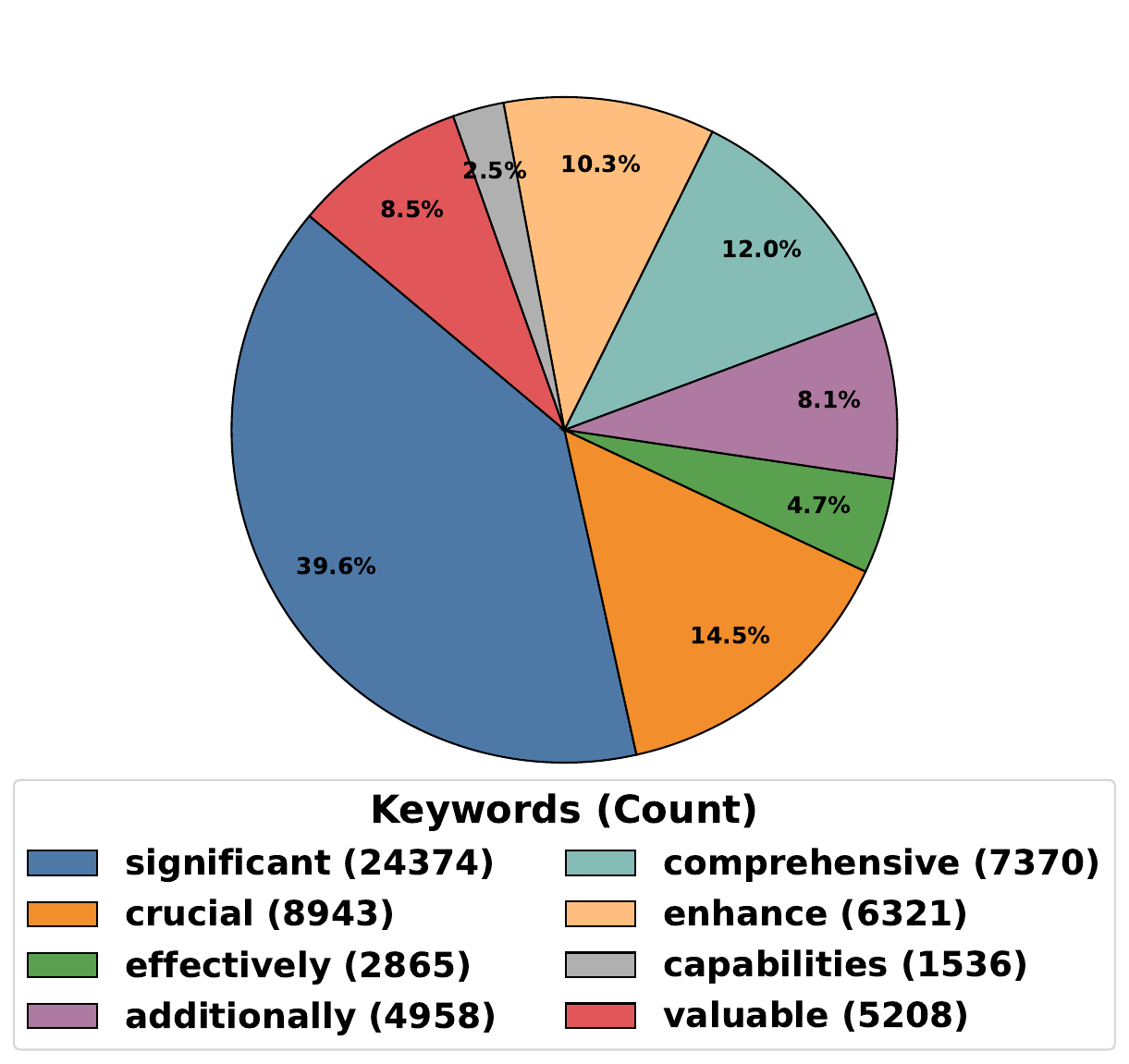}
        \caption{Moonshot}
    \end{subfigure}
    \hfill
    \begin{subfigure}[b]{0.32\textwidth}
        \includegraphics[width=\textwidth]{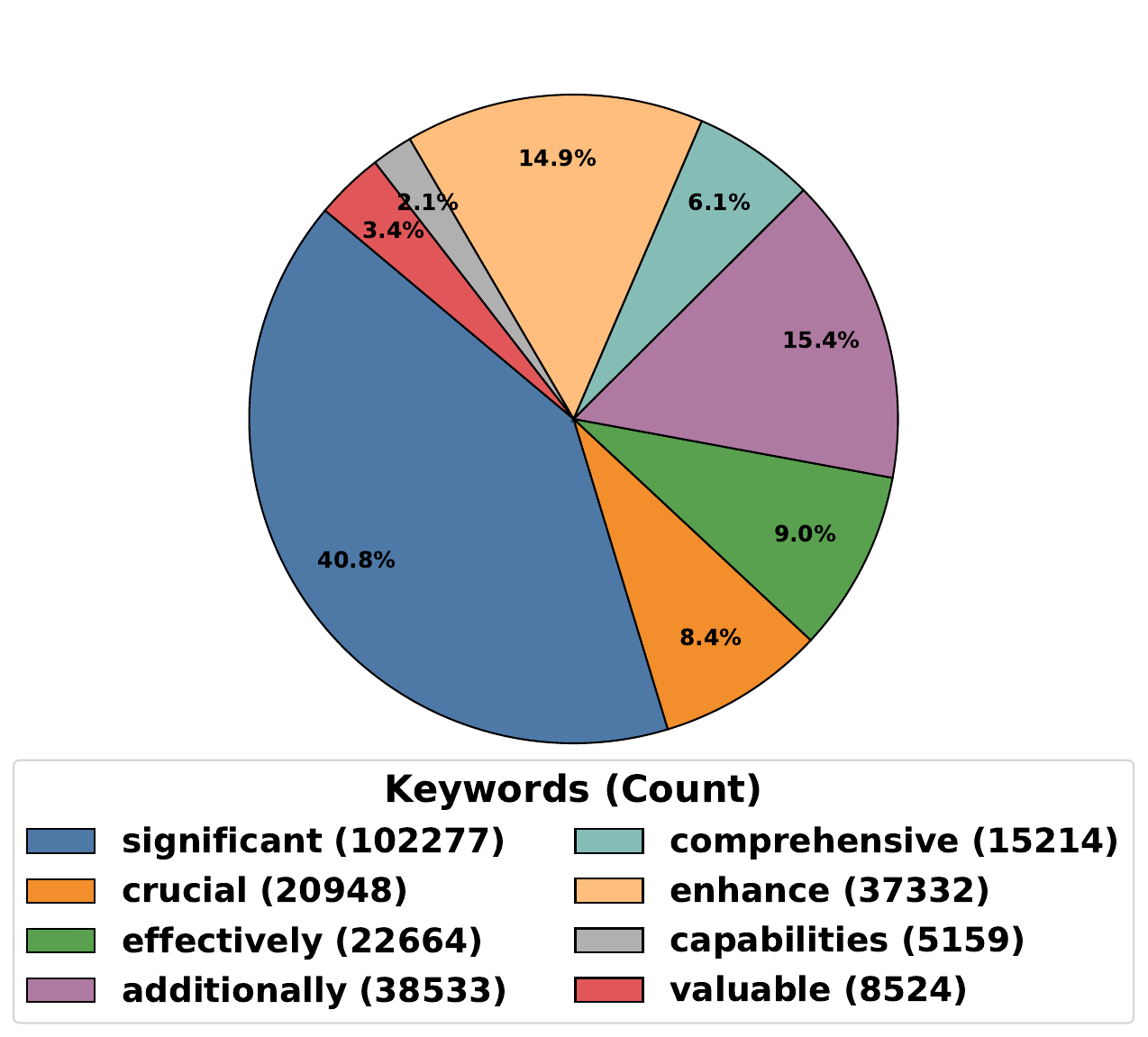}
        \caption{GPT4omini}
    \end{subfigure}
    \caption{Distribution of keyword weights across human-written text and various LLM outputs. Each pie chart shows the proportion of critical keywords in different sources under the given task settings.}
    \label{fig:pie_charts}
\end{figure*}

\mypara{Keyword Analysis}
Based on the findings of ~\cite{geng2024impact}, this study highlighted several critical keywords, such as \textit{significant}, \textit{comprehensive}, and \textit{enhance}, which exhibit notable patterns in their usage across different sources. 
Following the approach, we organized these keywords and analyzed their weights under our specific task settings, comparing human-written text with outputs generated by models such as GPT-3.5, Llama-3, Mixtral, and Moonshot. 
The resulting pie charts, as shown in \Cref{fig:pie_charts}, illustrate the distribution of these keywords for each source. We observe that human-written text heavily emphasizes impactful terms like \textit{significant} and \textit{enhance}, reflecting a formal academic style. GPT-3.5 and Llama-3 prioritize similar terms but exhibit a more balanced keyword usage. While Mixtral and Moonshot are closer to the human pattern in certain aspects, they have a relatively high proportion of descriptive terms such as \textit{effectively}, which emphasises their practical tone.

\mypara{Text Quality} 
To assess the quality of MGT compared to HWT, we conducted additional experiments using linguistic metrics. First, we measured readability using the Flesch Reading Ease (FSE) metric, which is calculated as:
\begin{align*}
    FSE &= 206.835 - 1.015 \times \left(\frac{\text{total words}}{\text{total sentences}}\right) \\
    &\quad - 84.6 \times \left(\frac{\text{total syllables}}{\text{total words}}\right)
\end{align*}

The FSE score ranges from 0 to 100, with higher scores indicating easier readability. 
The results show that MGT texts are generally harder to read compared to HWT, likely due to the varied expressions used by LLMs. However, the difference in readability is moderate and acceptable, with MGT texts scoring between 31.46 and 38.39, and HWT texts scoring between 39.68 and 39.79 (see Table~\ref{table:fse}).
    
    \vspace{-2mm}
\begin{table}[h]
    \centering
        \resizebox{\linewidth}{!}{\begin{tabular}{|c|c|c|c|c|c|}
    \hline
    Model & Moonshot & GPT-3.5 & Mixtral & Llama3 & GPT-4omini \\
    \hline
    MGT & 34.412 & 36.556 & 38.387 & 36.143 & 31.460 \\
    HWT & 39.682 & 39.778 & 39.831 & 39.733 & 39.789 \\
    \hline
    \end{tabular}}
    \caption{FSE scores for MGT and HWT.}
    \label{table:fse}
    \end{table}
    
\mypara{Text Characteristics} 
For text diversity, we use the Text-Token Ratio (TTR), defined as the ratio of unique tokens to the total number of tokens in the text. TTR values range from 0 to 1, with higher values indicating greater lexical diversity. The results showed that MGT texts generally have a higher TTR than HWT, indicating that machine-generated texts tend to use a wider range of vocabulary. MGT texts exhibited TTR values ranging from 0.522 to 0.596, while the HWT texts were more consistent, with TTR values around 0.517 (see Table~\ref{table:ttr}).

    \begin{table}[h]
    \centering
    \resizebox{\linewidth}{!}{\begin{tabular}{|c|c|c|c|c|c|}
    \hline
    Model & Moonshot & GPT-3.5 & Mixtral & Llama3 & GPT-4omini \\
    \hline
    MGT & 0.530 & 0.596 & 0.522 & 0.533 & 0.582 \\
    HWT & 0.518 & 0.517 & 0.517 & 0.517 & 0.517 \\
    \hline
    \end{tabular}}
    \caption{Text-Token Ratio (TTR) for MGT and HWT.}
    \label{table:ttr}
    \end{table}
    
    Lastly, for syntactic complexity, we measured the Average Sentence Length (ASL), which is the average number of words per sentence. MGT texts generally have shorter sentence lengths compared to HWT, suggesting simpler sentence structures. MGT sentences ranged from 19.39 to 22.16 words per sentence, whereas HWT sentences ranged from 23.30 to 23.35 words (see Table~\ref{table:asl}).

    \begin{table}[h]
    \centering
    \resizebox{\linewidth}{!}{\begin{tabular}{|c|c|c|c|c|c|}
    \hline
    Model & Moonshot & GPT-3.5 & Mixtral & Llama3 & GPT-4omini \\
    \hline
    MGT & 22.163 & 21.726 & 19.394 & 21.258 & 20.760 \\
    HWT & 23.347 & 23.296 & 23.294 & 23.314 & 23.296 \\
    \hline
    \end{tabular}}
    \caption{Average Sentence Length (ASL) for MGT and HWT.}
    \label{table:asl}
    \end{table}
    These results show that while MGT is generally more diverse, it tends to be simpler in terms of readability and syntactic complexity compared to HWT. However, these differences are not drastic and MGT does not show signs of overly simplistic or low-quality text.

\section{Code Framework}
\label{sec-app:code}

Despite the increasing number of benchmarks for MGT detection, there is no extensible standard library that offers out-of-the-box functionality.
Most existing benchmark code framework require additional configuration and have inefficient architectures, which increase the cost of further development and limit broader adoption. 
To address these issues, we introduce a newly refactored code framework based on MGTBench~\cite{HSCBZ23}.
Our framework follows the factory design pattern and implements \textit{AutoDetector} and \textit{AutoExperiment} for abstraction. 
This design is aligned with the approach used in Huggingface Transformers~\cite{WDSCDMCRLFB19}, which is the most widely used library in the NLP community.
Note that during the implementation, we leverage ChatGPT to assist in the development of the framework and manually check the generated code to ensure its correctness.

\mypara{Detector}
All the detectors follow the same hierarchy of abstraction that the base class implements the detecting process.
The base class follows the detectors' original implementation, enabling the flexibility to inspect the inner workflow of each detector easily.
Furthermore, we implement a unified API, i.e., \textit{AutoDetector}, to automatically instantiate the user-specified detector and enable convenient detector loading and switching.

\mypara{Experiment}
The experiment is designed to reflect the standard MGT detection pipeline: data processing, feature extraction (if any), and prediction making.
We adopt a similar abstraction strategy as the detector in which the base class implements the data processing and prediction analysis, leaving the specific predictions to different sub-classes.
The unified API, \textit{AutoExperiment}, is provided as well.

\begin{table*}[htbp]
\centering
\caption{Detailed Data Moderation Policy}
\label{tab:keywords}
\setlength{\tabcolsep}{8pt}
\renewcommand{\arraystretch}{1.6} 
\resizebox{\textwidth}{!}{\begin{tabular}{clcllllllll}
\toprule
\multirow{3}{*}{\textbf{Machine Split}} &
  Generation Identifier &
  \multicolumn{9}{c}{\begin{tabular}[c]{@{}c@{}}`revised book',`revised content',`revised version',`title',\\ `after editing',`revised section'\end{tabular}} \\ \cmidrule{2-11} 
 &
  \multicolumn{1}{c}{Special Keywords} &
  \multicolumn{9}{c}{\begin{tabular}[c]{@{}c@{}}`book editor', `clarity', `revisions', 'I apologize',`I am sorry',  \\  `Unfortunately', `complex language', `revised content', `revised version',   \\ `language model', `revised content', `revised version', `accuracy of',  \\  `project gutenberg',  `reliable information', `ISBN', `PMID', `doi:',\\ `Sure,',   `Retrieved from',`Category', `http', `As an editor', `As an expert'\end{tabular}} \\ \cmidrule{2-11} 
 &
  \multicolumn{1}{c}{Format Symbols} &
  \multicolumn{9}{c}{`\&', `\$', `====', `---',`**', `\#\#', } \\ \midrule
\multirow{3}{*}{\textbf{Human Split}} &
   Special Keywords &
  \multicolumn{9}{c}{`ISBN', `PMID', `doi', `vol.', `p.', `https:', `http:', `References External links'} \\ \cmidrule{2-11} 
 &
  \multicolumn{1}{c}{Format Symbols} &
 \multicolumn{9}{c}{`\textbackslash n---', `\textbackslash n===', `\texttt{**}', `\#\#', `\$' ($>500$), `\&' ($>150$), `\texttt{\textbackslash}' ($>1000$)} \\ 
  \bottomrule
  
\end{tabular}
}\end{table*}

\section{Detectors}\label{sec-app:detecors}
For zero-shot detectors, metrics were obtained from the white-box model Llama2-7B-Instruct~\cite{TMSAABBBBBBBCCCEFFFFGGGHHHIKKKKKKLLLLLMMMMMNPRRSSSSSTTTWKXYZZFKNRSES23}, unless stated otherwise.
Fast-DetectGPT and Binoculars were evaluated using the optimal settings specified in their respective papers.
For model-based detectors, DistilBERT and RoBERTa were fine-tuned with a learning rate of 5e-6, batch size of 64, 3 epochs, and a random seed of 3407. 
RADAR and ChatGPT-D used their officially released weights without additional fine-tuning.
Details of detectors in binary classification and text attribution are provided in~\Cref{sec-app:hyperparameter}.

\mypara{Log-Likehood~\cite{GSR19}}
A zero-shot method uses a language model to compute the log probability of each token in a text. A higher average log-likelihood suggests the text is more likely generated by an LLM.

\mypara{Rank~\cite{GSR19}}
A zero-shot method calculates the absolute rank of each token in a text based on its previous context and determines the text's score by averaging these rank values. A smaller average rank score indicates a higher likelihood that the text is machine-generated.

\mypara{Rank GLTR~\cite{GSR19}}
GLTR is designed to assist in labeling machine-generated text. We uses Test-2 features as suggested by Guo et al.~\cite{GZWJNDYW23}, evaluating the fraction of words ranked within 10, 100, 1,000, and beyond.

\mypara{LRR~\cite{JYDP23}} 
The Log-Likelihood Log-Rank Ratio (LRR) combines Log-Likelihood and Log-Rank, with a higher LRR indicating a greater likelihood of text being machine-generated.

\mypara{Entropy~\cite{GSR19}}
A zero-shot method uses entropy to measure text randomness, with lower entropy indicating a higher likelihood of being LLM-generated, as human-written text shows greater unpredictability.

\mypara{Fast-DetectGPT~\cite{BZTYZ24}}
An optimized zero-shot detector improves DetectGPT~\cite{MLKMF23} by replacing perturbation with efficient sampling. We followed the authors' optimal settings, using GPT-Neo-2.7b as the scoring model and GPT-J-6b as the reference model.

\mypara{Binoculars~\cite{binoculars}}
A zero-shot detection method uses two LLMs to compute the perplexity-to-cross-perplexity ratio. A lower score indicates the text is likely machine-generated. Following the authors' optimal settings, we used Falcon-7B-Instruct for PPL and Falcon-7B with Falcon-7B-Instruct for X-PPL.

\mypara{RADAR~\cite{hu2023radar}}
RADAR uses adversarial training between a paraphraser and a detector. We used the pre-trained RoBERTa detector from Hugging Face without additional training.

\mypara{ChatGPT-D~\cite{GZWJNDYW23}}
ChatGPT Detector distinguishes human-written from ChatGPT-generated texts by fine-tuning a RoBERTa model on the HC3~\cite{GZWJNDYW23} dataset.

\mypara{DistilBERT~\cite{distilbert}}
The detector is built by fine-tuning a pre-trained DistilBERT model with an additional classification layer.

\mypara{RoBERTa~\cite{roberta}}
The detector is built by fine-tuning a pre-trained RoBERTa model with an additional classification layer.

\section{Experimental Settings}\label{sec-app:hyperparameter}

\mypara{In-distribution Experiment}
For zero-shot detectors, we randomly selected 1,000 training samples to predict the metrics. 
The classification threshold was set to maximize the F1 score on the training set, and a classifier was trained using the same data. 
Note that GLTR produces vectors of four rank values, and thus, threshold-based classification is not applicable. 
For model-based detectors, we fine-tuned the model using at most 10,000 training samples. 
We used 2000 randomly selected data points for the testing set.
Zero-shot detectors employ threshold-based classification and logistic regression for binary human-machine classification tasks. 

Model attribution tasks use SVM and logistic regression classifiers with default sklearn implementations and a linear kernel for SVM.
Model-based detectors have their classification heads adjusted to match the number of classes. Fine-tuning was done with a learning rate of 5e-6, batch size of 64, 3 epochs, and a random seed of 3407.

\mypara{Domains and LLMs Transfer Experiment}
For zero-shot detectors, we applied the threshold and classifier from the source domain directly to the target domain. 
For model-based detectors, we used models fine-tuned on the source domain and evaluated them on the target domain test data.

\mypara{Class Incremental Experiment}
To train the original model, we use a setting similar to that in the model attribution task.
We train the model for 2 epochs in this stage and set the learning rate to 5e-6 and the batch size to 64.
To train the updated model, training data has the same number of data as each class had in the previous stage.
We train the model for 1 epoch in this stage and set the learning rate to 2.5e-7 (1/4 of the original lr) and the batch size to 64.
For the LwF technique, the regularization parameter is set to 0.2.
To maintain the example in iCaRL, we set the cache size for each class to 100.
The validation set in BiC is constructed by combining the data in the example together.
Specifically, since a small amount of old data is introduced in the training process of iCaRL and BiC, we adopt weighted cross entropy to avoid the side effects of data imbalance.
\section{Ablation Study of Zero-shot Detectors in In-distribution Experiment}\label{sec-app:ablate_bench}
We use GPT-2-medium \cite{RWCLAS19} and Llama-2-7B-Instruct \cite{TMSAABBBBBBBCCCEFFFFGGGHHHIKKKKKKLLLLLMMMMMNPRRSSSSSTTTWKXYZZFKNRSES23} as the metric-generator model for zero-shot detectors. Results are shown in Table \ref{tab:ablation-in-d}. Full results of zero-shot detectors in model attribution are provided in Table \ref{tab:zeroshot-attribution}.

\begin{table*}[]
\centering
\caption{In-Distribution Performance of Zero-shot Detectors with GPT-2 and LLama-2}
\label{tab:ablation-in-d}
\resizebox{\textwidth}{!}{
\begin{tabular}{c|ccc|ccc|ccc|ccc|ccc}
\toprule
{\color[HTML]{333333} }                              & \multicolumn{3}{c|}{{\color[HTML]{333333} Llama-3.1-70b}}                                                                                                          & \multicolumn{3}{c|}{{\color[HTML]{333333} Mixtral-8$\times$7b}}                                                                                                           & \multicolumn{3}{c|}{{\color[HTML]{333333} MoonShot-8k}}                                                                                                            & \multicolumn{3}{c|}{{\color[HTML]{333333} GPT-4omini}}                                                                                                            & \multicolumn{3}{c}{{\color[HTML]{333333} GPT-3.5}}                                                                                                                 \\ \cmidrule{2-16} 
\multirow{-2}{*}{{\color[HTML]{333333} }}            & {\color[HTML]{333333} ST.}                           & {\color[HTML]{333333} Hu.}                           & {\color[HTML]{333333} So.}                           & {\color[HTML]{333333} ST.}                           & {\color[HTML]{333333} Hu.}                           & {\color[HTML]{333333} So.}                           & {\color[HTML]{333333} ST.}                           & {\color[HTML]{333333} Hu.}                           & {\color[HTML]{333333} So.}                           & {\color[HTML]{333333} ST.}                           & {\color[HTML]{333333} Hu.}                           & {\color[HTML]{333333} So.}                           & {\color[HTML]{333333} ST.}                           & {\color[HTML]{333333} Hu.}                           & {\color[HTML]{333333} So.}                           \\
\bottomrule
\end{tabular}}
\end{table*}

\section{Results of Domains and LLMs Transfer}\label{sec-app:transfer}

\mypara{Domain Transfer}
Table \ref{tab:complete-domain-transfer} presents the full results of detector generalization across different domains.
\begin{table*}[]
\centering
\caption{\textbf{Experiment Result for Transferring Across Different Domains in Binary Classification.} 
We train the model on one domain and test the model on the other domain.
ST. represents STEM, Hu. represents Humanity, and So. represents Social Science. 
The results are reported using F1 score.
The larger values with blue colors indicate better performance and lower values with red colors indicate weaker performance.}
\setlength{\tabcolsep}{3pt}
\renewcommand{\arraystretch}{1.1}
\small
\label{tab:binary-transfer-domain}
\resizebox{\textwidth}{!}{
\begin{tabular}{c|c|ccc|ccc|ccc|ccc|ccc}\toprule
                                 & \multicolumn{1}{l|}{}                                                                                                      & \multicolumn{3}{c|}{LLama3}                                                                   & \multicolumn{3}{c|}{Mixtral}                                                                  & \multicolumn{3}{c|}{Moonshot}                                                                 & \multicolumn{3}{c|}{GPT-4omini}                                                               & \multicolumn{3}{c}{GPT-3.5}                                                                   \\ \cmidrule{3-17} 
\multirow{-2}{*}{Method}         & \multicolumn{1}{l|}{\multirow{-2}{*}{\begin{tabular}[c]{@{}l@{}} Target Topic $\rightarrow$ \\ Source Topic $\downarrow$\end{tabular}}} & Hu.                           & ST.                           & So.                           & Hu.                           & ST.                           & So.                           & Hu.                           & ST.                           & So.                           & Hu.                           & ST.                           & So.                           & Hu.                           & ST.                           & So.                           \\ \midrule
                                 & Hu.                                                                                                                        & \cellcolor[HTML]{FEFCFB}0.794 & \cellcolor[HTML]{FEF6F1}0.736 & \cellcolor[HTML]{FEFDFD}0.805 & \cellcolor[HTML]{FEF7F4}0.749 & \cellcolor[HTML]{FEF1E9}0.688 & \cellcolor[HTML]{FFFFFF}0.815 & \cellcolor[HTML]{FEF9F5}0.760 & \cellcolor[HTML]{FEF5EF}0.723 & \cellcolor[HTML]{FFFFFF}0.816 & \cellcolor[HTML]{FEF1E9}0.689 & \cellcolor[HTML]{FEF0E8}0.683 & \cellcolor[HTML]{FEF8F5}0.759 & \cellcolor[HTML]{FDE8DB}0.606 & \cellcolor[HTML]{FDE0CF}0.535 & \cellcolor[HTML]{FEEDE4}0.656 \\
                                 & ST.                                                                                                                        & \cellcolor[HTML]{FEF0E8}0.682 & \cellcolor[HTML]{FEF4EE}0.714 & \cellcolor[HTML]{FEFCFA}0.790 & \cellcolor[HTML]{FEEFE6}0.671 & \cellcolor[HTML]{FEEEE5}0.662 & \cellcolor[HTML]{FEFDFD}0.805 & \cellcolor[HTML]{FEF6F2}0.740 & \cellcolor[HTML]{FEF3ED}0.711 & \cellcolor[HTML]{FFFFFF}0.818 & \cellcolor[HTML]{FDE4D5}0.569 & \cellcolor[HTML]{FDEBE1}0.638 & \cellcolor[HTML]{FEF2EB}0.696 & \cellcolor[HTML]{FDE2D1}0.546 & \cellcolor[HTML]{FDDBC7}0.481 & \cellcolor[HTML]{FDE7D9}0.592 \\
\multirow{-3}{*}{log-likelihood} & So.                                                                                                                        & \cellcolor[HTML]{FEF7F2}0.743 & \cellcolor[HTML]{FEF6F1}0.732 & \cellcolor[HTML]{FEFDFD}0.803 & \cellcolor[HTML]{FEF0E8}0.682 & \cellcolor[HTML]{FEEEE6}0.666 & \cellcolor[HTML]{FEFEFE}0.809 & \cellcolor[HTML]{FEF2EC}0.703 & \cellcolor[HTML]{FEF0E8}0.683 & \cellcolor[HTML]{FEFEFD}0.806 & \cellcolor[HTML]{FEF3ED}0.709 & \cellcolor[HTML]{FEF0E8}0.682 & \cellcolor[HTML]{FEF9F6}0.765 & \cellcolor[HTML]{FEF1EA}0.691 & \cellcolor[HTML]{FDE7DA}0.598 & \cellcolor[HTML]{FEF3ED}0.709 \\ \midrule
                                 & Hu.                                                                                                                        & \cellcolor[HTML]{FEF8F5}0.759 & \cellcolor[HTML]{FEF7F3}0.749 & \cellcolor[HTML]{FEFEFE}0.810 & \cellcolor[HTML]{FEF2EB}0.701 & \cellcolor[HTML]{FEF2EB}0.701 & \cellcolor[HTML]{FFFFFF}0.815 & \cellcolor[HTML]{FEF6F1}0.734 & \cellcolor[HTML]{FEF5F0}0.726 & \cellcolor[HTML]{FEFDFC}0.799 & \cellcolor[HTML]{FEF1EA}0.693 & \cellcolor[HTML]{FEEFE6}0.671 & \cellcolor[HTML]{FEF7F3}0.747 & \cellcolor[HTML]{FEF0E8}0.679 & \cellcolor[HTML]{FDE9DE}0.619 & \cellcolor[HTML]{FEF4EE}0.719 \\
                                 & ST.                                                                                                                        & \cellcolor[HTML]{FDE9DD}0.617 & \cellcolor[HTML]{FEF4EF}0.720 & \cellcolor[HTML]{FEFBF9}0.784 & \cellcolor[HTML]{FDE9DC}0.611 & \cellcolor[HTML]{FEEDE4}0.655 & \cellcolor[HTML]{FEFAF7}0.772 & \cellcolor[HTML]{FDE3D4}0.562 & \cellcolor[HTML]{FDE7DA}0.600 & \cellcolor[HTML]{FEF2EB}0.701 & \cellcolor[HTML]{FDE7DA}0.598 & \cellcolor[HTML]{FEEEE4}0.658 & \cellcolor[HTML]{FEF3ED}0.711 & \cellcolor[HTML]{FEF0E8}0.680 & \cellcolor[HTML]{FDE9DE}0.620 & \cellcolor[HTML]{FEF4EF}0.721 \\
\multirow{-3}{*}{Rank GLTR}      & So.                                                                                                                        & \cellcolor[HTML]{FEF0E9}0.684 & \cellcolor[HTML]{FEF7F3}0.748 & \cellcolor[HTML]{FEFDFC}0.802 & \cellcolor[HTML]{FEEEE5}0.662 & \cellcolor[HTML]{FEF0E8}0.680 & \cellcolor[HTML]{FEFEFD}0.808 & \cellcolor[HTML]{FEF1E9}0.686 & \cellcolor[HTML]{FEF3ED}0.709 & \cellcolor[HTML]{FEFCFB}0.795 & \cellcolor[HTML]{FDE6D9}0.592 & \cellcolor[HTML]{FEEEE5}0.660 & \cellcolor[HTML]{FEF4ED}0.713 & \cellcolor[HTML]{FDECE2}0.644 & \cellcolor[HTML]{FDE6D8}0.586 & \cellcolor[HTML]{FEF1EA}0.694 \\ \midrule
                                 & Hu.                                                                                                                        & \cellcolor[HTML]{EBF4F9}0.897 & \cellcolor[HTML]{EFF6FA}0.879 & \cellcolor[HTML]{EDF5FA}0.887 & \cellcolor[HTML]{F8FBFD}0.845 & \cellcolor[HTML]{FAFCFE}0.836 & \cellcolor[HTML]{F1F7FB}0.873 & \cellcolor[HTML]{F2F8FB}0.867 & \cellcolor[HTML]{EAF3F9}0.899 & \cellcolor[HTML]{EEF6FA}0.883 & \cellcolor[HTML]{FEFAF7}0.772 & \cellcolor[HTML]{FEF2EB}0.700 & \cellcolor[HTML]{FEFDFC}0.802 & \cellcolor[HTML]{FEFDFD}0.803 & \cellcolor[HTML]{FEEEE5}0.663 & \cellcolor[HTML]{FEFCFB}0.792 \\
                                 & ST.                                                                                                                        & \cellcolor[HTML]{EAF3F9}0.899 & \cellcolor[HTML]{EFF6FA}0.881 & \cellcolor[HTML]{E9F3F8}0.903 & \cellcolor[HTML]{F6FAFC}0.854 & \cellcolor[HTML]{FBFDFE}0.833 & \cellcolor[HTML]{EFF6FA}0.880 & \cellcolor[HTML]{F6FAFC}0.854 & \cellcolor[HTML]{E4F0F7}0.923 & \cellcolor[HTML]{E8F2F8}0.908 & \cellcolor[HTML]{FEF7F3}0.748 & \cellcolor[HTML]{FEF3ED}0.710 & \cellcolor[HTML]{FEF7F3}0.748 & \cellcolor[HTML]{FEF2EA}0.694 & \cellcolor[HTML]{FEF0E8}0.680 & \cellcolor[HTML]{FEEEE4}0.658 \\
\multirow{-3}{*}{Binoculars}     & So.                                                                                                                        & \cellcolor[HTML]{F0F7FA}0.877 & \cellcolor[HTML]{EEF6FA}0.883 & \cellcolor[HTML]{E7F2F8}0.911 & \cellcolor[HTML]{FEFEFE}0.813 & \cellcolor[HTML]{FDFEFF}0.824 & \cellcolor[HTML]{ECF5F9}0.890 & \cellcolor[HTML]{FBFDFE}0.834 & \cellcolor[HTML]{E4F0F7}0.924 & \cellcolor[HTML]{E6F1F7}0.916 & \cellcolor[HTML]{FEF6F2}0.738 & \cellcolor[HTML]{FDEBE1}0.639 & \cellcolor[HTML]{FEFDFC}0.800 & \cellcolor[HTML]{FEFDFC}0.801 & \cellcolor[HTML]{FEEFE6}0.670 & \cellcolor[HTML]{FEFCFB}0.792 \\ \midrule
                                 & Hu.                                                                                                                        & \cellcolor[HTML]{FFFFFF}0.817 & \cellcolor[HTML]{FEFDFC}0.799 & \cellcolor[HTML]{F2F8FB}0.868 & \cellcolor[HTML]{FEF9F5}0.759 & \cellcolor[HTML]{FEF9F5}0.761 & \cellcolor[HTML]{F9FCFD}0.840 & \cellcolor[HTML]{FEFDFC}0.801 & \cellcolor[HTML]{F9FCFD}0.842 & \cellcolor[HTML]{EBF4F9}0.895 & \cellcolor[HTML]{FEF4EE}0.718 & \cellcolor[HTML]{FEF1E9}0.688 & \cellcolor[HTML]{FEF6F2}0.740 & \cellcolor[HTML]{FEF3ED}0.713 & \cellcolor[HTML]{FEEFE7}0.672 & \cellcolor[HTML]{FEF8F5}0.758 \\
                                 & ST.                                                                                                                        & \cellcolor[HTML]{FEFFFF}0.819 & \cellcolor[HTML]{FFFFFF}0.817 & \cellcolor[HTML]{EFF6FA}0.878 & \cellcolor[HTML]{FEF9F6}0.765 & \cellcolor[HTML]{FEF9F5}0.760 & \cellcolor[HTML]{F8FBFD}0.843 & \cellcolor[HTML]{FEFDFC}0.801 & \cellcolor[HTML]{F9FCFD}0.842 & \cellcolor[HTML]{EBF4F9}0.897 & \cellcolor[HTML]{FEF5EF}0.723 & \cellcolor[HTML]{FEF1E9}0.688 & \cellcolor[HTML]{FEF8F4}0.754 & \cellcolor[HTML]{FEF2EC}0.702 & \cellcolor[HTML]{FEF0E7}0.677 & \cellcolor[HTML]{FEF1E9}0.689 \\
\multirow{-3}{*}{Fast-DetectGPT} & So.                                                                                                                        & \cellcolor[HTML]{FFFFFF}0.815 & \cellcolor[HTML]{FFFFFF}0.816 & \cellcolor[HTML]{EDF5FA}0.887 & \cellcolor[HTML]{FEFAF7}0.769 & \cellcolor[HTML]{FEF8F5}0.755 & \cellcolor[HTML]{F9FCFD}0.842 & \cellcolor[HTML]{FEFDFC}0.802 & \cellcolor[HTML]{FAFCFE}0.837 & \cellcolor[HTML]{EAF3F9}0.899 & \cellcolor[HTML]{FEF5EF}0.725 & \cellcolor[HTML]{FEF1E9}0.689 & \cellcolor[HTML]{FEF8F4}0.752 & \cellcolor[HTML]{FEF2EA}0.695 & \cellcolor[HTML]{FEEDE4}0.656 & \cellcolor[HTML]{FEF8F5}0.756 \\ \midrule
                                 & Hu.                                                                                                                        & \cellcolor[HTML]{D7E8F2}0.978 & \cellcolor[HTML]{F0F7FA}0.876 & \cellcolor[HTML]{E3EFF6}0.930 & \cellcolor[HTML]{D5E7F2}0.985 & \cellcolor[HTML]{F2F8FB}0.870 & \cellcolor[HTML]{E0EEF5}0.941 & \cellcolor[HTML]{D3E6F1}0.992 & \cellcolor[HTML]{EDF5F9}0.890 & \cellcolor[HTML]{DAEAF3}0.963 & \cellcolor[HTML]{D4E7F1}0.987 & \cellcolor[HTML]{E6F1F7}0.915 & \cellcolor[HTML]{DBEBF4}0.958 & \cellcolor[HTML]{D7E8F2}0.977 & \cellcolor[HTML]{FEFEFE}0.810 & \cellcolor[HTML]{E6F1F7}0.918 \\
                                 & ST.                                                                                                                        & \cellcolor[HTML]{F9FCFD}0.840 & \cellcolor[HTML]{D4E7F1}0.987 & \cellcolor[HTML]{D9EAF3}0.967 & \cellcolor[HTML]{E2EFF6}0.932 & \cellcolor[HTML]{D6E8F2}0.980 & \cellcolor[HTML]{DBEBF4}0.958 & \cellcolor[HTML]{DFEDF5}0.946 & \cellcolor[HTML]{D4E7F1}0.989 & \cellcolor[HTML]{D5E7F1}0.986 & \cellcolor[HTML]{E5F1F7}0.919 & \cellcolor[HTML]{D6E8F2}0.981 & \cellcolor[HTML]{D8E9F3}0.971 & \cellcolor[HTML]{DFEDF5}0.942 & \cellcolor[HTML]{D5E7F2}0.985 & \cellcolor[HTML]{DEEDF5}0.946 \\
\multirow{-3}{*}{DistillBert-F}  & So.                                                                                                                        & \cellcolor[HTML]{E1EEF5}0.938 & \cellcolor[HTML]{E0EEF5}0.940 & \cellcolor[HTML]{DCEBF4}0.957 & \cellcolor[HTML]{DAEAF3}0.963 & \cellcolor[HTML]{DCEBF4}0.957 & \cellcolor[HTML]{D7E9F2}0.975 & \cellcolor[HTML]{D8E9F3}0.972 & \cellcolor[HTML]{D7E9F2}0.976 & \cellcolor[HTML]{D4E7F1}0.987 & \cellcolor[HTML]{D5E8F2}0.982 & \cellcolor[HTML]{D6E8F2}0.978 & \cellcolor[HTML]{D5E8F2}0.982 & \cellcolor[HTML]{DAEAF3}0.964 & \cellcolor[HTML]{DFEDF5}0.943 & \cellcolor[HTML]{DBEBF4}0.960 \\ \midrule
                                 & Hu.                                                                                                                        & \cellcolor[HTML]{D2E6F1}0.995 & \cellcolor[HTML]{D5E7F2}0.984 & \cellcolor[HTML]{D2E6F1}0.995 & \cellcolor[HTML]{D2E6F1}0.997 & \cellcolor[HTML]{D3E6F1}0.993 & \cellcolor[HTML]{D2E6F1}0.994 & \cellcolor[HTML]{D3E6F1}0.994 & \cellcolor[HTML]{D4E7F1}0.987 & \cellcolor[HTML]{D2E6F1}0.996 & \cellcolor[HTML]{D2E6F1}0.994 & \cellcolor[HTML]{D3E6F1}0.991 & \cellcolor[HTML]{D2E6F1}0.996 & \cellcolor[HTML]{D4E7F1}0.986 & \cellcolor[HTML]{E8F2F8}0.907 & \cellcolor[HTML]{DBEBF4}0.959 \\
                                 & ST.                                                                                                                        & \cellcolor[HTML]{D3E6F1}0.992 & \cellcolor[HTML]{D4E7F1}0.987 & \cellcolor[HTML]{D2E6F1}0.995 & \cellcolor[HTML]{D2E6F1}0.997 & \cellcolor[HTML]{D3E6F1}0.993 & \cellcolor[HTML]{D3E6F1}0.993 & \cellcolor[HTML]{D4E7F1}0.989 & \cellcolor[HTML]{D3E6F1}0.994 & \cellcolor[HTML]{D1E5F0}0.998 & \cellcolor[HTML]{D4E7F1}0.988 & \cellcolor[HTML]{D4E7F1}0.987 & \cellcolor[HTML]{D3E6F1}0.993 & \cellcolor[HTML]{D7E9F2}0.975 & \cellcolor[HTML]{D4E7F1}0.986 & \cellcolor[HTML]{D8E9F3}0.973 \\
\multirow{-3}{*}{Roberta-F}      & So.                                                                                                                        & \cellcolor[HTML]{D3E6F1}0.993 & \cellcolor[HTML]{D5E7F2}0.985 & \cellcolor[HTML]{D2E6F1}0.995 & \cellcolor[HTML]{D1E5F0}0.998 & \cellcolor[HTML]{D2E6F1}0.995 & \cellcolor[HTML]{D2E6F1}0.996 & \cellcolor[HTML]{D4E7F1}0.988 & \cellcolor[HTML]{D4E7F1}0.988 & \cellcolor[HTML]{D2E6F1}0.997 & \cellcolor[HTML]{D3E6F1}0.993 & \cellcolor[HTML]{D2E6F1}0.994 & \cellcolor[HTML]{D2E6F1}0.995 & \cellcolor[HTML]{D6E8F2}0.980 & \cellcolor[HTML]{D4E7F1}0.988 & \cellcolor[HTML]{D5E8F2}0.983 \\\bottomrule
\end{tabular}
}
\end{table*}

\mypara{LLM Transfer}
Table \ref{tab:complete-llm-transfer} presents the full results of detector generalization across different LLM generations.
\begin{table*}[]
\centering
\caption{\textbf{Experiment Result for Transferring Across Different LLMs in Binary Classification.}
We train the model on data generated by one LLM and test the model on the data generated by another LLM.
ST. represents STEM, Hu. represents Humanity, and So. represents Social Science. 
The results are reported using F1 score and averaged across three domains.
The larger values with blue colors indicate better performance and lower values with red colors indicate worse performance.}
\label{tab:binary-transfer-LLM}
\setlength{\tabcolsep}{3pt}
\renewcommand{\arraystretch}{1.1} 
\small
\resizebox{\textwidth}{!}{
\begin{tabular}{c|c|ccccc|c|cccccc} \toprule
                                 & \begin{tabular}[c]{@{}c@{}}Target Topic $\rightarrow$ \\ Source Topic $\downarrow$\end{tabular} & Llama3.1                      & Mixtral                       & Moonshot                      & GPT-4omini                    & GPT-3.5                       &                                  &            & Llama3.1                      & Mixtral                       & Moonshot                      & GPT-4omini                    & GPT-3.5                       \\ \midrule
                                 & Llama3.1                                                                        & \cellcolor[HTML]{FFFFFF}0.770 & \cellcolor[HTML]{FEF9F6}0.735 & \cellcolor[HTML]{FEF9F6}0.734 & \cellcolor[HTML]{FDECE1}0.647 & \cellcolor[HTML]{FDE7DA}0.618 &                                  & Llama3.1   & \cellcolor[HTML]{F1F7FB}0.840 & \cellcolor[HTML]{FEFFFF}0.777 & \cellcolor[HTML]{F0F7FA}0.847 & \cellcolor[HTML]{FDDBC7}0.535 & \cellcolor[HTML]{FDE8DB}0.620 \\
                                 & Mixtral                                                                         & \cellcolor[HTML]{FEFFFF}0.775 & \cellcolor[HTML]{FEFAF7}0.740 & \cellcolor[HTML]{FEFBF9}0.748 & \cellcolor[HTML]{FEEDE4}0.658 & \cellcolor[HTML]{FDE9DD}0.627 &                                  & Mixtral    & \cellcolor[HTML]{F2F8FB}0.837 & \cellcolor[HTML]{FCFDFE}0.787 & \cellcolor[HTML]{F0F7FA}0.845 & \cellcolor[HTML]{FDE9DD}0.630 & \cellcolor[HTML]{FEF0E7}0.672 \\
                                 & Moonshot                                                                        & \cellcolor[HTML]{FDFEFF}0.781 & \cellcolor[HTML]{FEFBF9}0.745 & \cellcolor[HTML]{FEFDFC}0.759 & \cellcolor[HTML]{FEF0E8}0.675 & \cellcolor[HTML]{FDEADF}0.637 &                                  & Moonshot   & \cellcolor[HTML]{F1F7FB}0.842 & \cellcolor[HTML]{FDFEFF}0.783 & \cellcolor[HTML]{F0F7FA}0.847 & \cellcolor[HTML]{FDE0CE}0.568 & \cellcolor[HTML]{FDEBE0}0.642 \\
                                 & GPT-4omini                                                                      & \cellcolor[HTML]{FEFFFF}0.775 & \cellcolor[HTML]{FEFBF9}0.745 & \cellcolor[HTML]{FEFDFC}0.758 & \cellcolor[HTML]{FEF3ED}0.697 & \cellcolor[HTML]{FDECE1}0.647 &                                  & GPT-4omini & \cellcolor[HTML]{FEFCFB}0.755 & \cellcolor[HTML]{FEFCFB}0.755 & \cellcolor[HTML]{FEFEFE}0.765 & \cellcolor[HTML]{FEF7F3}0.719 & \cellcolor[HTML]{FEF6F1}0.713 \\
\multirow{-5}{*}{log-likelihood} & GPT-3.5                                                                         & \cellcolor[HTML]{FEF9F6}0.735 & \cellcolor[HTML]{FEF5EF}0.706 & \cellcolor[HTML]{FEF5F0}0.707 & \cellcolor[HTML]{FDE6D9}0.612 & \cellcolor[HTML]{FDE4D6}0.599 & \multirow{-5}{*}{Fast-DetectGPT} & GPT-3.5    & \cellcolor[HTML]{FEFEFE}0.768 & \cellcolor[HTML]{FEFEFD}0.764 & \cellcolor[HTML]{FFFFFF}0.772 & \cellcolor[HTML]{FEF6F1}0.714 & \cellcolor[HTML]{FEF6F2}0.715 \\ \midrule
                                 & Llama3.1                                                                        & \cellcolor[HTML]{FEFDFC}0.760 & \cellcolor[HTML]{FEF7F2}0.719 & \cellcolor[HTML]{FEF5F0}0.707 & \cellcolor[HTML]{FDE5D7}0.602 & \cellcolor[HTML]{FDE8DC}0.626 &                                  & Llama3.1   & \cellcolor[HTML]{D6E8F2}0.974 & \cellcolor[HTML]{DDECF4}0.941 & \cellcolor[HTML]{D4E7F1}0.983 & \cellcolor[HTML]{E5F1F7}0.899 & \cellcolor[HTML]{F2F8FB}0.837 \\
                                 & Mixtral                                                                         & \cellcolor[HTML]{FEFEFD}0.765 & \cellcolor[HTML]{FEF7F3}0.721 & \cellcolor[HTML]{FEF6F1}0.714 & \cellcolor[HTML]{FDE7D9}0.614 & \cellcolor[HTML]{FDEADF}0.636 &                                  & Mixtral    & \cellcolor[HTML]{D8E9F2}0.965 & \cellcolor[HTML]{D5E7F2}0.980 & \cellcolor[HTML]{D3E6F1}0.988 & \cellcolor[HTML]{ECF4F9}0.867 & \cellcolor[HTML]{F4F9FC}0.824 \\
                                 & Moonshot                                                                        & \cellcolor[HTML]{FEFDFD}0.762 & \cellcolor[HTML]{FEF7F3}0.723 & \cellcolor[HTML]{FEF5F0}0.710 & \cellcolor[HTML]{FDE7DA}0.615 & \cellcolor[HTML]{FDE9DE}0.632 &                                  & Moonshot   & \cellcolor[HTML]{E2EFF6}0.916 & \cellcolor[HTML]{E4F0F6}0.906 & \cellcolor[HTML]{D3E6F1}0.989 & \cellcolor[HTML]{E8F2F8}0.887 & \cellcolor[HTML]{FEF7F3}0.720 \\
                                 & GPT-4omini                                                                      & \cellcolor[HTML]{FEFFFF}0.778 & \cellcolor[HTML]{FEFBFA}0.749 & \cellcolor[HTML]{FEFCFA}0.752 & \cellcolor[HTML]{FEF2EB}0.688 & \cellcolor[HTML]{FEF1E9}0.680 &                                  & GPT-4omini & \cellcolor[HTML]{E5F1F7}0.898 & \cellcolor[HTML]{EEF6FA}0.853 & \cellcolor[HTML]{D6E8F2}0.975 & \cellcolor[HTML]{D4E7F1}0.983 & \cellcolor[HTML]{FEF1E9}0.680 \\
\multirow{-5}{*}{Rank   GLTR}    & GPT-3.5                                                                         & \cellcolor[HTML]{FFFFFF}0.771 & \cellcolor[HTML]{FEFAF7}0.739 & \cellcolor[HTML]{FEFAF7}0.740 & \cellcolor[HTML]{FDECE2}0.650 & \cellcolor[HTML]{FEEEE5}0.664 & \multirow{-5}{*}{DistillBert-F}  & GPT-3.5    & \cellcolor[HTML]{FDE5D7}0.602 & \cellcolor[HTML]{FDE7DA}0.616 & \cellcolor[HTML]{FEF0E8}0.673 & \cellcolor[HTML]{FDDDCB}0.552 & \cellcolor[HTML]{D6E8F2}0.974 \\ \midrule
                                 & Llama3.1                                                                        & \cellcolor[HTML]{E6F1F7}0.896 & \cellcolor[HTML]{EEF5FA}0.858 & \cellcolor[HTML]{E5F0F7}0.901 & \cellcolor[HTML]{FEF3ED}0.694 & \cellcolor[HTML]{FEFAF8}0.743 &                                  & Llama3.1   & \cellcolor[HTML]{D2E6F1}0.992 & \cellcolor[HTML]{D2E6F1}0.991 & \cellcolor[HTML]{D2E6F1}0.992 & \cellcolor[HTML]{D3E6F1}0.989 & \cellcolor[HTML]{E7F2F7}0.891 \\
                                 & Mixtral                                                                         & \cellcolor[HTML]{E6F1F7}0.893 & \cellcolor[HTML]{EEF6FA}0.856 & \cellcolor[HTML]{E5F0F7}0.900 & \cellcolor[HTML]{FEF6F2}0.717 & \cellcolor[HTML]{FEFBF9}0.746 &                                  & Mixtral    & \cellcolor[HTML]{D4E7F1}0.982 & \cellcolor[HTML]{D1E5F0}0.995 & \cellcolor[HTML]{D2E6F1}0.991 & \cellcolor[HTML]{D3E6F1}0.990 & \cellcolor[HTML]{E6F1F7}0.892 \\
                                 & Moonshot                                                                        & \cellcolor[HTML]{E6F1F7}0.894 & \cellcolor[HTML]{EEF5FA}0.856 & \cellcolor[HTML]{E4F0F7}0.902 & \cellcolor[HTML]{FEF5F0}0.707 & \cellcolor[HTML]{FEFAF8}0.742 &                                  & Moonshot   & \cellcolor[HTML]{D3E6F1}0.988 & \cellcolor[HTML]{D3E6F1}0.989 & \cellcolor[HTML]{D2E6F1}0.995 & \cellcolor[HTML]{D4E7F1}0.985 & \cellcolor[HTML]{E8F2F8}0.884 \\
                                 & GPT-4omini                                                                      & \cellcolor[HTML]{F3F8FB}0.832 & \cellcolor[HTML]{F5FAFC}0.819 & \cellcolor[HTML]{F3F8FB}0.832 & \cellcolor[HTML]{FEFDFC}0.761 & \cellcolor[HTML]{FEFDFB}0.756 &                                  & GPT-4omini & \cellcolor[HTML]{D3E7F1}0.986 & \cellcolor[HTML]{D2E6F1}0.994 & \cellcolor[HTML]{D2E6F1}0.991 & \cellcolor[HTML]{D2E6F1}0.992 & \cellcolor[HTML]{E6F1F7}0.897 \\
\multirow{-5}{*}{Binoculars}     & GPT-3.5                                                                         & \cellcolor[HTML]{F6FAFC}0.818 & \cellcolor[HTML]{F9FCFD}0.802 & \cellcolor[HTML]{F8FBFD}0.806 & \cellcolor[HTML]{FEF8F4}0.725 & \cellcolor[HTML]{FEFDFC}0.758 & \multirow{-5}{*}{Roberta-F}      & GPT-3.5    & \cellcolor[HTML]{E3EFF6}0.910 & \cellcolor[HTML]{DEECF4}0.935 & \cellcolor[HTML]{DDECF4}0.938 & \cellcolor[HTML]{E9F3F8}0.879 & \cellcolor[HTML]{D4E7F1}0.985 \\\bottomrule
\end{tabular}
}
\end{table*}

\begin{table*}[]
\centering
\caption{Full Results of Zero-shot Detectors using Different Classifier in Model Attribution}
\label{tab:zeroshot-attribution}
\resizebox{\textwidth}{!}{
\begin{tabular}{c|ccc|c|ccc}\toprule
\textbf{Logistic   Regression} & STEM                           & Humanities                     & Social\_sci.                   & \textbf{SVM}         & STEM                           & Humanities                     & Social\_sci.                   \\ \midrule
LL (Llama-2)                   & \cellcolor[HTML]{FEF6F1}0.1661 & \cellcolor[HTML]{FFFFFF}0.1806 & \cellcolor[HTML]{F0F7FB}0.2151 & LL (Llama-2)         & \cellcolor[HTML]{FEF0E8}0.1519 & \cellcolor[HTML]{FEFAF7}0.1736 & \cellcolor[HTML]{F0F7FB}0.2225 \\
Rank (Llama-2)                 & \cellcolor[HTML]{FDE7D9}0.1392 & \cellcolor[HTML]{FDECE1}0.1481 & \cellcolor[HTML]{FEF4EE}0.1624 & Rank (Llama-2)       & \cellcolor[HTML]{FDDBC7}0.1021 & \cellcolor[HTML]{FEF8F5}0.1702 & \cellcolor[HTML]{FAFDFE}0.1978 \\
LRR (Llama-2)             & \cellcolor[HTML]{FEF2EA}0.1581 & \cellcolor[HTML]{FDDBC7}0.1182 & \cellcolor[HTML]{FCFDFE}0.1894 & LRR (Llama-2)   & \cellcolor[HTML]{FEEFE7}0.1499 & \cellcolor[HTML]{FDE2D2}0.1198 & \cellcolor[HTML]{FFFFFF}0.1852 \\
Rank-GLTR (Llama-2)            & \cellcolor[HTML]{FDE8DB}0.1412 & \cellcolor[HTML]{FDE3D4}0.1334 & \cellcolor[HTML]{FBFDFE}0.1912 & Rank-GLTR (Llama-2)  & \cellcolor[HTML]{FDDCC8}0.105  & \cellcolor[HTML]{FDDECC}0.1095 & \cellcolor[HTML]{FDFEFF}0.1899 \\
Entropy (Llama-2)              & \cellcolor[HTML]{FAFCFE}0.1941 & \cellcolor[HTML]{FDE8DC}0.1422 & \cellcolor[HTML]{EBF4F9}0.2282 & Entropy (Llama-2)    & \cellcolor[HTML]{F6FAFC}0.2089 & \cellcolor[HTML]{FDE7D9}0.1299 & \cellcolor[HTML]{EAF4F9}0.237  \\
Fast-DetectGPT                 & \cellcolor[HTML]{F1F7FB}0.2137 & \cellcolor[HTML]{ECF4F9}0.2251 & \cellcolor[HTML]{D1E5F0}0.2849 & Fast-DetectGPT       & \cellcolor[HTML]{FFFFFF}0.1848 & \cellcolor[HTML]{EEF5FA}0.2289 & \cellcolor[HTML]{D1E5F0}0.299  \\
\bottomrule
\end{tabular}}
\end{table*}

\begin{table*}[]
\centering
\caption{Transferring Results Across Different Domains}
\label{tab:complete-domain-transfer}
\resizebox{\textwidth}{!}{\begin{tabular}{c|c|c|ccccccccccccccc}\toprule
                                             &                              &                    & \multicolumn{3}{c|}{LLama3}                                                                                        & \multicolumn{3}{c|}{Mixtral}                                                                                       & \multicolumn{3}{c|}{Moonshot}                                                                                      & \multicolumn{3}{c|}{GPT-4omini}                                                                                    & \multicolumn{3}{c}{GPT-3.5}                                                                   \\ \cmidrule{4-18} 
\multirow{-2}{*}{\textbf{Domain   transfer}} & \multirow{-2}{*}{Classifier} & \multirow{-2}{*}{} & Hu.                           & ST.                           & \multicolumn{1}{c|}{So.}                           & Hu.                           & ST.                           & \multicolumn{1}{c|}{So.}                           & Hu.                           & ST.                           & \multicolumn{1}{c|}{So.}                           & Hu.                           & ST.                           & \multicolumn{1}{c|}{So.}                           & Hu.                           & ST.                           & So.                           \\
\bottomrule
\end{tabular}}
\end{table*}

\begin{table*}[]
\centering
\caption{Full LLM Transfer Results}
\label{tab:complete-llm-transfer}
\resizebox{\textwidth}{!}{\begin{tabular}{c|c|c|ccccc|ccccc|ccccc}\toprule                                  &                              &                    & \multicolumn{5}{c|}{Humanities}                                                                                                                               & \multicolumn{5}{c|}{STEM}                                                                                                                                     & \multicolumn{5}{c}{Social Sciences}                                                                                                                           \\ \cmidrule{4-18} 
\multirow{-2}{*}{\textbf{}}       & \multirow{-2}{*}{Classifier} & \multirow{-2}{*}{} & Llama3                        & Mixtral                       & Moonshot                      & GPT-4omini                    & GPT-3.5                       & Llama3                        & Mixtral                       & Moonshot                      & GPT-4omini                    & GPT-3.5                       & Llama3                        & Mixtral                       & Moonshot                      & GPT-4omini                    & GPT-3.5                       \\ \midrule
                                  &                              & Llama3             & \cellcolor[HTML]{E6F1F7}0.876 & \cellcolor[HTML]{F1F7FB}0.811 & \cellcolor[HTML]{EDF5F9}0.832 & \cellcolor[HTML]{FEFBF9}0.645 & \cellcolor[HTML]{F5FAFC}0.782 & \cellcolor[HTML]{E5F1F7}0.880 & \cellcolor[HTML]{F0F7FA}0.813 & \cellcolor[HTML]{DFEDF5}0.918 & \cellcolor[HTML]{FEF5F0}0.540 & \cellcolor[HTML]{FEF9F6}0.615 & \cellcolor[HTML]{E2EFF6}0.898 & \cellcolor[HTML]{E7F2F8}0.868 & \cellcolor[HTML]{DFEDF5}0.913 & \cellcolor[HTML]{FEFEFE}0.711 & \cellcolor[HTML]{F9FCFD}0.759 \\
\bottomrule
\end{tabular}}
\end{table*}

\begin{figure}[htbp]
    \centering
    \includegraphics[width=\linewidth]{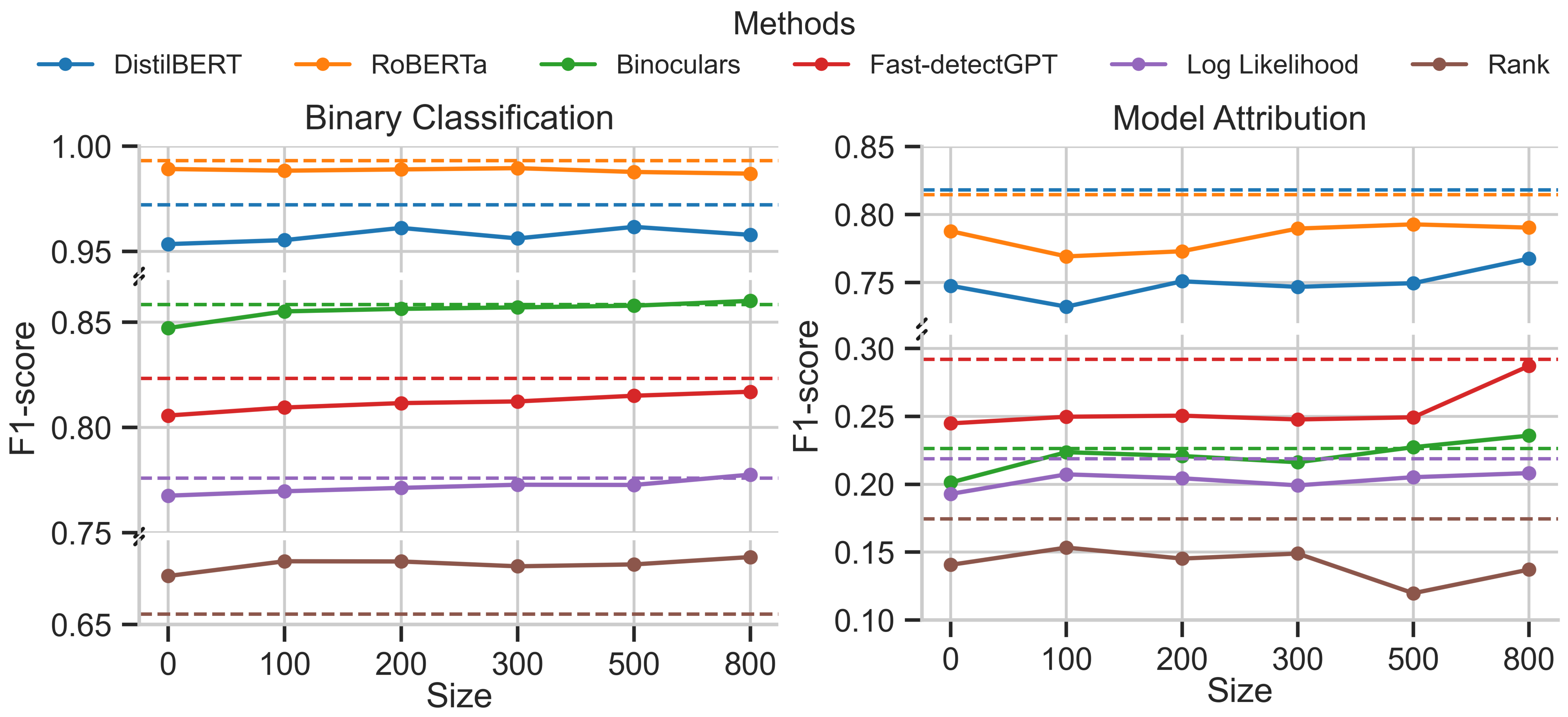}
    \caption{Transfer to Social Sciences Domain}
    \label{fig:mitigate-social}
\end{figure}

\begin{figure}[t!]
    \centering
    \includegraphics[width=\linewidth]{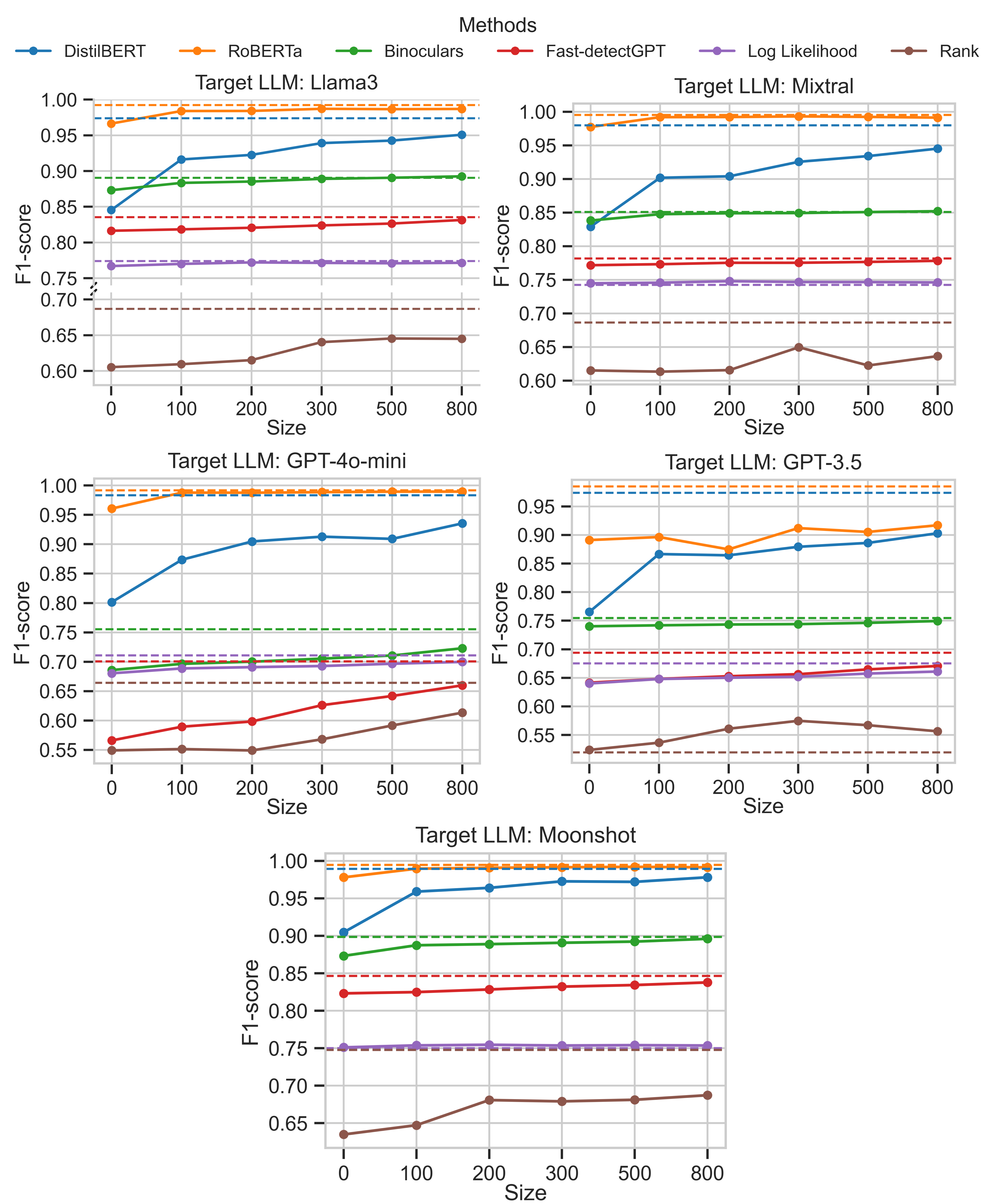}
    \caption{Transfer to Different Target LLMs}
    \label{fig:mitigate-llm}
\end{figure}

\begin{figure}[t!]
    \centering
    \includegraphics[width=\linewidth]{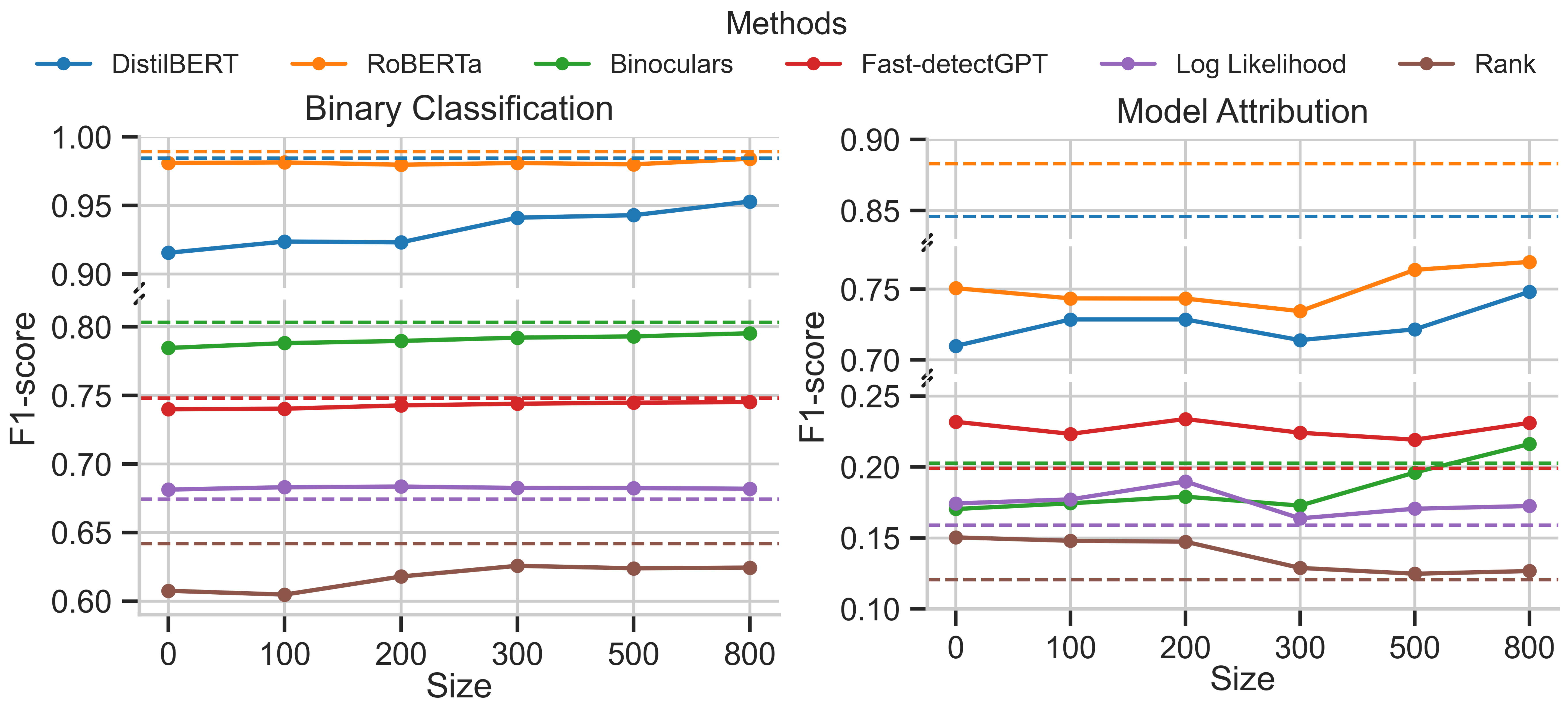}
    \caption{Transfer to STEM Domain}
    \label{fig:mitigate-stem}
\end{figure}

\begin{figure}[t!]
    \centering
    \includegraphics[width=\linewidth]{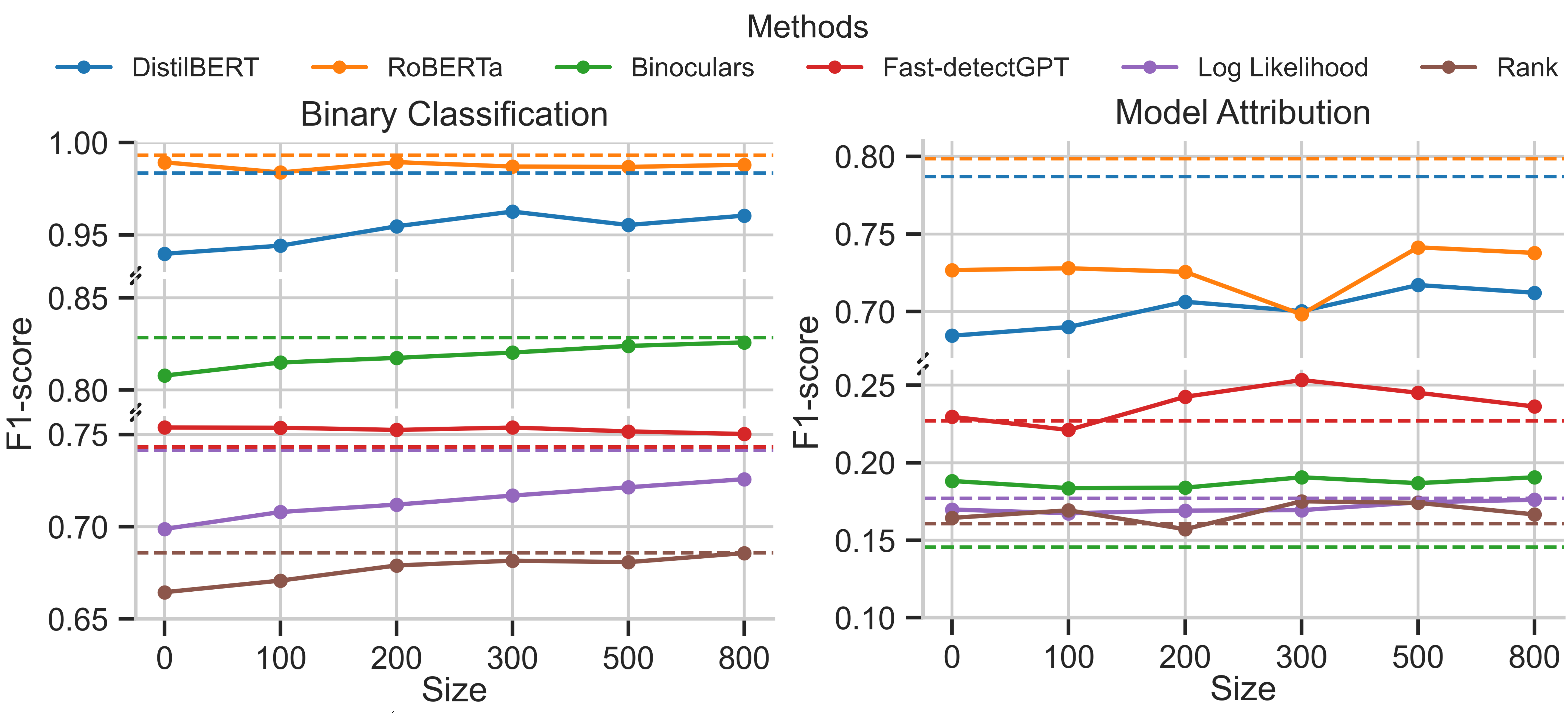}
    \caption{Transfer to Humanities Domain}
    \label{fig:mitigate-human}
\end{figure}

\mypara{Full Mitigation Result}
Figure~\ref{fig:mitigate-social}, ~\ref{fig:mitigate-human}, and~\ref{fig:mitigate-stem} illustrate the mitigation effects when transferring across domain topics.
Similarly, Figure~\ref{fig:mitigate-llm} shows the effects when transferring across different LLMs.

\section{Class Incremental Techniques}\label{sec-app:CILtechniques}
\mypara{LwF~\cite{li2017learning}}
LwF is a regularization-based method that relieves CF by distilling the logits from the previous model into the new one. 
The key idea is to preserve the logits of the previous model on old tasks during new tasks by adding a regularization term to the loss function:
\begin{equation}
    \mathcal{L} = \mathcal{L}_{\text{new}}(x, y) + \lambda \mathcal{L}_{\text{distill}}(x),
\end{equation}
where $\mathcal{L}_{\text{new}}$ is the classification loss for new tasks, $\mathcal{L}_{\text{distill}}$ is the distillation loss to keep the results for old tasks, and $\lambda$ balances the two terms. 
This encourages the model to learn new knowledge while retaining old information.

\mypara{iCaRL~\cite{rebuffi2017icarl}}
iCaRL introduces a rehearsal-based approach to CIL. 
The method addresses forgetting by memory replaying.
It maintains a fixed memory buffer to store a subset of examples from previous tasks, which are replayed during training to retain knowledge.
To balance old and new knowledge, iCaRL uses a knowledge distillation loss similar to LwF.
The exemplars allow iCaRL to replay old class information effectively, thus mitigating forgetting.

\mypara{BiC~\cite{wu2019large}}
BiC focuses on addressing the bias towards new classes that arise in CIL. 
When new classes are added, the model tends to favor these classes due to their dominance in the learning process. 
BiC first trains the model on the data (n old classes and m new classes) without any correction.
Then, it adds a bias correction layer that adjusts the logits of new classes:
    \begin{equation}
        q_k =
        \begin{cases}
            o_k, & \text{if } 1 \leq k \leq n, \\
            \alpha o_k + \beta, & \text{if } n+1 \leq k \leq n+m,
        \end{cases}
    \end{equation}
where $q_k$ is the new logit for class $k$, $o_k$ is the original logit, and $\alpha$, $\beta$ are one-dimensional learnable parameters. 
The bias correction layer ensures that the model's predictions remain balanced across old and new classes.

\mypara{Combine}
We integrate the knowledge distillation loss, memory replayer, and logit calibration techniques to produce a combined method to see if the performance can be different.

\begin{table*}[t]
\centering
\caption{Results for Introducing Two New LLMs in the Update Stage.}
\label{tab:cil-42}
\resizebox{\textwidth}{!}{\begin{tabular}{c|c|ccccc|ccccc} \toprule
 &  & \multicolumn{5}{c|}{DistilBert} & \multicolumn{5}{c}{RoBERTa} \\ \cmidrule{3-12} 
\multirow{-2}{*}{Domain} & \multirow{-2}{*}{\textbf{Last Model}} & Normal & LwF & iCaRL & \multicolumn{1}{c|}{BiC} & Attribution & Normal & LwF & iCaRL & \multicolumn{1}{c|}{BiC} & Attribution \\ \midrule
Social Science &  & \cellcolor[HTML]{FEF2EB}0.4529 & \cellcolor[HTML]{F6FAFC}0.4791 & \cellcolor[HTML]{F6FAFC}0.4788 & \multicolumn{1}{c|}{\cellcolor[HTML]{F6FAFC}0.4792} & 0.8174 & \cellcolor[HTML]{FDEADE}0.4407 & \cellcolor[HTML]{FDDBC7}0.4187 & \cellcolor[HTML]{EAF3F8}0.4891 & \multicolumn{1}{c|}{\cellcolor[HTML]{EAF3F8}0.4891} & 0.8177 \\
STEM &  & \cellcolor[HTML]{FEF3EC}0.4540 & \cellcolor[HTML]{FFFFFF}0.4712 & \cellcolor[HTML]{FEFEFD}0.4698 & \multicolumn{1}{c|}{\cellcolor[HTML]{FEFEFE}0.4703} & 0.8444 & \cellcolor[HTML]{EEF6FA}0.4857 & \cellcolor[HTML]{D1E5F0}0.5098 & \cellcolor[HTML]{DDECF4}0.5000 & \multicolumn{1}{c|}{\cellcolor[HTML]{DCECF4}0.5005} & 0.8881 \\
Humanity & \multirow{-3}{*}{\begin{tabular}[c]{@{}c@{}}GPT-4omini\\ +\\ Llama3\end{tabular}} & \cellcolor[HTML]{FDE3D4}0.4313 & \cellcolor[HTML]{FDE4D6}0.4327 & \cellcolor[HTML]{FEFBF9}0.4657 & \multicolumn{1}{c|}{\cellcolor[HTML]{FEFBF9}0.4657} & 0.7835 & \cellcolor[HTML]{FEF3EC}0.4538 & \cellcolor[HTML]{FEF2EA}0.4521 & \cellcolor[HTML]{F2F8FB}0.4821 & \multicolumn{1}{c|}{\cellcolor[HTML]{F1F7FB}0.4833} & 0.8151 \\\bottomrule
\end{tabular}}
\end{table*}

\section{Few-shot Experiments}\label{sec-app:few}

\begin{figure}[t]
    \centering
    \includegraphics[width=1\linewidth]{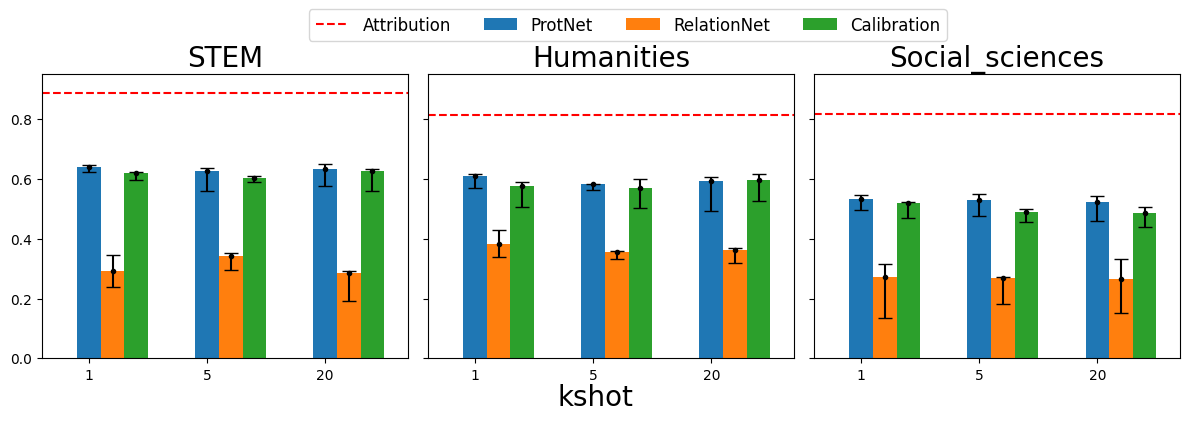}
    \caption{The number of newly introduced classes is two.}
    \label{fig:few_roberta_32}
\end{figure}

\begin{figure}[t]
    \centering
    \includegraphics[width=1\linewidth]{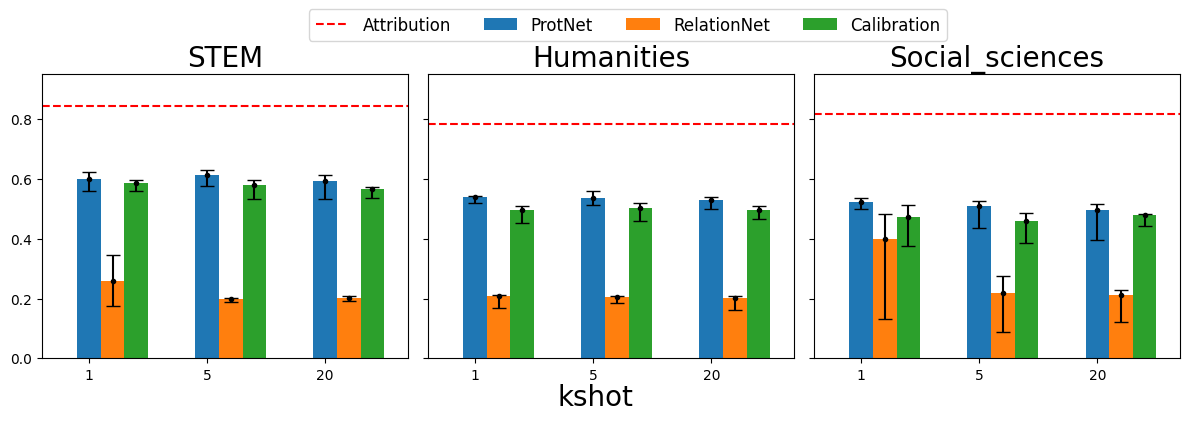}
    \caption{The number of newly introduced classes is two.}
    \label{fig:few_disti_32}
\end{figure}

\begin{figure*}[htbp]
    \centering
    \includegraphics[width=1\textwidth]{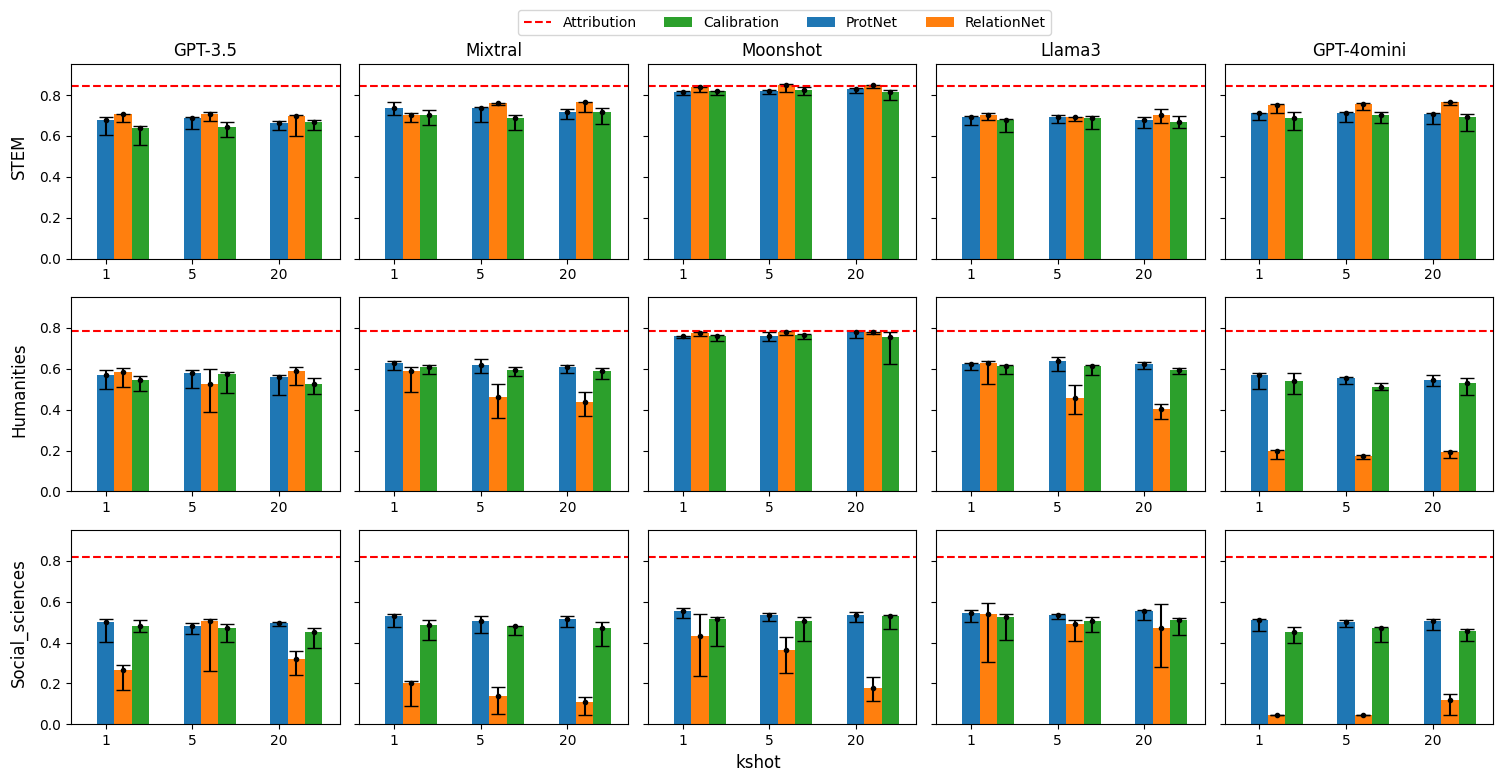}
    \caption{The number of newly introduced class is one.}
    \label{fig:few_distil}
\end{figure*}

\begin{figure*}[t]
    \centering
    \includegraphics[width=1\textwidth]{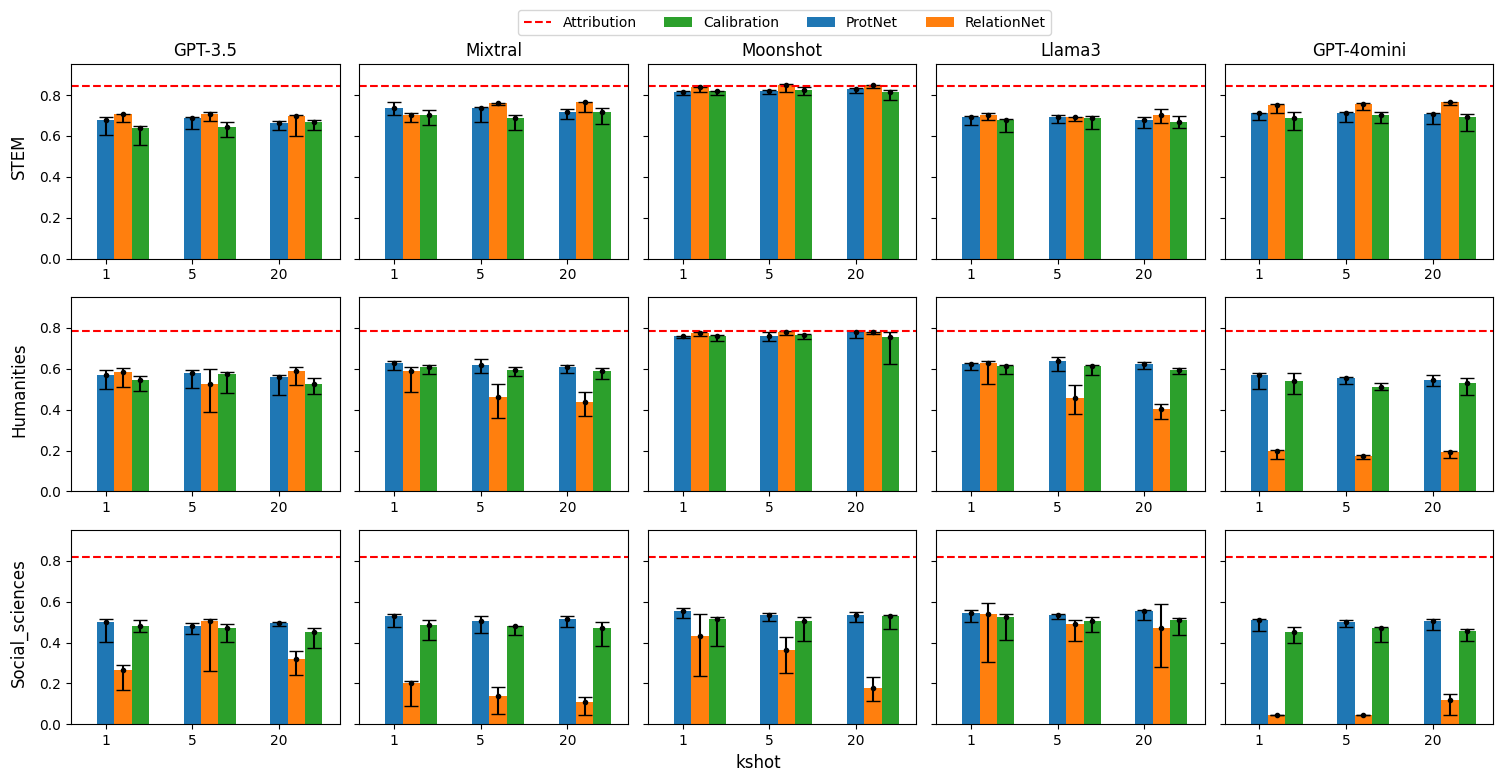}
    \caption{The number of newly introduced class is one.}
    \label{fig:few_roberta}
\end{figure*}

For DistilBert, the results are in Figure~\ref{fig:few_distil} and Figure~\ref{fig:few_disti_32}.
For Roberta, the results are in Figure~\ref{fig:few_roberta} and Figure~\ref{fig:few_roberta_32}.

\section{Many-shot Experiments}\label{sec-app:cil}

Table~\ref{tab-app:cil-deberta} shows the performance of DeBerta.
\Cref{tab:cil-42} shows the performance of introducing two LLMs in the update stage.
The results drop rapidly as the number of new LLMs increases.


\begin{table}[htbp]
\centering
\caption{Result for DeBerta in many-shot settings}
\label{tab-app:cil-deberta}
\begin{tabular}{lllll}
DeBerta & \multicolumn{1}{c}{Normal} & \multicolumn{1}{c}{LwF} & \multicolumn{1}{c}{iCaRL} & \multicolumn{1}{c}{BiC} \\ \hline
Social Science & 0.4839 & 0.4786 & 0.4996 & 0.4978 \\
STEM & 0.5926 & 0.5807 & 0.6342 & 0.6358 \\
Humanity & 0.4779 & 0.4903 & 0.5093 & 0.5049
\end{tabular}
\end{table}



\end{document}